\documentclass[11pt]{article}
\usepackage{amsmath}
\usepackage{amsthm}
\usepackage{amssymb}
\usepackage{graphicx}
\usepackage{adjustbox}
\usepackage{mathtools}
\usepackage{bbm}
\usepackage[margin=2.5cm]{geometry}
\usepackage{setspace}
\usepackage{epstopdf}
\usepackage{subfig}
\usepackage{empheq}
\usepackage{algorithm}
\usepackage{algorithmicx}
\usepackage{algpseudocode}
\usepackage{multirow}
\usepackage{url}
\usepackage[title]{appendix}
\usepackage{enumitem}
\usepackage[svgnames]{xcolor}
\usepackage{diagbox}
\usepackage{tabu}
\usepackage{nicematrix}
\usepackage{mathrsfs}
\usepackage{amsbsy}

\makeatletter

\theoremstyle{plain}
\newtheorem{thm}{\protect\theoremname}
  \theoremstyle{definition}
  \newtheorem{defn}[thm]{\protect\definitionname}
  
  \theoremstyle{plain}
  \newtheorem{lem}[thm]{\protect\lemmaname}
  \newtheorem{prop}[thm]{\protect\propname}
  \newtheorem{cor}[thm]{\protect\corollaryname}
  
  \newtheorem{assump}{Assumption}
  
  \providecommand{\propname}{Proposition}
  \providecommand{\condname}{Condition}

  \providecommand{\keywords}[1]
{
  \small	
  \textbf{\textit{Keywords---}} #1
}

\makeatother

  \providecommand{\definitionname}{Definition}
  \providecommand{\lemmaname}{Lemma}
  \providecommand{\corollaryname}{Corollary}
\providecommand{\theoremname}{Theorem}

\usepackage{endnotes}
\usepackage{color}
\usepackage{xparse}

\newtheorem{myremark}{Remark}

\title{Entropic Optimal Transport Eigenmaps for Nonlinear Alignment and Joint Embedding of High-Dimensional Datasets}

\author{ Boris Landa${^{1,3,*}}$~~~~Yuval Kluger${^{3,4,5}}$~~~~Rong Ma${^{2,\dagger}}$\\
\small{${^1}$Department of Electrical and Computer Engineering, Yale University}\\
\small{${^2}$Department of Biostatistics, Harvard University}\\
\small{${^3}$Program in Applied Mathematics, Yale University}\\
\small{${^4}$Interdepartmental Program in Computational Biology and Bioinformatics, Yale University}\\
\small{${^5}$Department of Pathology, Yale University School of Medicine}\\
\small{${^*}$Corresponding author. Email: boris.landa@yale.edu}\\
\small{${^\dagger}$Corresponding author. Email: rongma@hsph.harvard.edu}
}

\begin{document}

\maketitle

\begin{abstract}
Embedding high-dimensional data into a low-dimensional space is an indispensable component of data analysis. In numerous applications, it is necessary to align and jointly embed multiple datasets from different studies or experimental conditions. Such datasets may share underlying structures of interest but exhibit individual distortions, resulting in misaligned embeddings using traditional techniques. In this work, we propose \textit{Entropic Optimal Transport (EOT) eigenmaps}, a principled approach for aligning and jointly embedding a pair of datasets with theoretical guarantees. Our approach leverages the leading singular vectors of the EOT plan matrix between two datasets to extract their shared underlying structure and align them in a common embedding space. We interpret our approach as an inter-data variant of the classical Laplacian eigenmaps and diffusion maps embeddings, showing that it enjoys many favorable analogous properties. {We analyze a generative model in which two observed high-dimensional datasets share latent variables supported on a common low-dimensional manifold, while each dataset is subject to translation, geometric distortion, orthogonal nuisance structure, and noise. In a large-sample, high-dimensional regime, we prove that the EOT plan concentrates around a population kernel on an effective manifold determined by the geometric mean of the distortions, with invariance to translations, orthogonal nuisance structure, and noise.} Subsequently, we relate our embedding to eigenfunctions of population-level operators encoding the density and geometry of the shared manifold. Finally, we showcase the performance of our approach for data integration and embedding through simulations and analyses of real-world biological data, demonstrating its advantages over alternative methods in challenging scenarios.
\end{abstract}


\section{Introduction}
The challenge of effectively analyzing high-dimensional data is prevalent across many scientific disciplines. Techniques for embedding high-dimensional data into lower-dimensional spaces are commonly utilized to simplify complex data and aid in various data analytic tasks including clustering, visualization, and manifold learning. Traditional embedding and dimensionality reduction methods primarily focus on extracting low-dimensional structures from a single dataset. Specifically, given a dataset $\mathcal{X} = \{x_1,\ldots,x_m\} \subset \mathbb{R}^p$, their goal is to embed $\mathcal{X}$ into $\mathbb{R}^q$, producing $\widetilde{\mathcal{X}} = \{\widetilde{x}_1,\ldots,\widetilde{x}_m\} \subset \mathbb{R}^q$, typically with $q \ll p$. To capture nonlinear structures in the data, popular techniques such as Laplacian eigenmaps~\cite{belkin2003laplacian}, diffusion maps~\cite{coifman2006diffusionMaps}, t-distributed Stochastic Neighbor Embedding (tSNE)~\cite{maaten2008visualizing}, and UMAP~\cite{becht2019dimensionality}, first construct a similarity graph over the data points in $\mathbb{R}^p$, represented by an affinity matrix $W\in\mathbb{R}^{m\times m}$. Then, the embedded data points are arranged in $\mathbb{R}^q$ according to the entries of $W$ or its eigen-decomposition, preserving key structural characteristics of the data. 

In many modern applications such as genomics \cite{tran2020benchmark}, precision medicine \cite{sheller2020federated} and business analytics \cite{dayal2009data}, there is a need to integrate and jointly analyze multiple datasets. Specifically, in single-cell omics research, to understand a biological process of interest, it is common that multiple high-dimensional datasets are generated using the same type of omics assays but based on different samples or experiments~\cite{stuart2019integrative,argelaguet2021computational,luecken2022benchmarking}. Such datasets may share common low-dimensional structures that characterize the underlying biological process. However, due to variations in biological samples or experimental conditions, each dataset may exhibit unique deformations, corruptions, and nuisance structures. These phenomena are commonly known as batch effects in genomics and related fields~\cite{haghverdi2018batch,tran2020benchmark}. To effectively analyze such datasets, it is desirable to embed them jointly into a lower-dimensional space while preserving their common underlying structures and filtering out individual distortions.

Here, we consider a setup with two datasets in $\mathbb{R}^p$, $\mathcal{X} = \{x_1,\ldots,x_m\} \subset \mathbb{R}^p$ and $\mathcal{Y} = \{y_1,\ldots,y_n\} \subset \mathbb{R}^p$, where the goal is to jointly embed them into $\mathbb{R}^q$ as $\widetilde{\mathcal{X}} = \{\widetilde{x}_1,\ldots,\widetilde{x}_m\} \subset \mathbb{R}^{q}$ and $\widetilde{\mathcal{Y}} = \{\widetilde{y}_1,\ldots,\widetilde{y}_n\} \subset \mathbb{R}^{q}$, respectively. A naive approach is to concatenate the datasets, i.e., to form $\{\mathcal{X},\mathcal{Y}\}$, and then apply standard off-the-shelf embedding techniques. However, if dataset-specific distortions are present, traditional methods \cite{pham2018large,shen2022scalability,stuart2019comprehensive,ding2023learning} designed for single datasets are typically suboptimal in extracting or preserving common structures based on the concatenated data, resulting in embeddings where $\widetilde{\mathcal{X}}$ and $\widetilde{\mathcal{Y}}$ are misaligned \cite{ma2024principled,maan2024characterizing}; see also Figure~\ref{simu.fig} in Section~\ref{simu.sec} for numerical evidence. To address this issue, numerous approaches adopt a multi-step process in which the datasets are explicitly aligned, e.g., using affine transformations, either before or after the embedding~\cite{stuart2019comprehensive,hie2019efficient,argelaguet2020mofa+,liu2020jointly}. Other common approaches rely on nonlinear functions such as neural networks to generate embeddings where the datasets are optimally aligned according to a prescribed loss function~\cite{li2020deep,amodio2018magan,lopez2018deep,cao2022unified,demetci2022scot}. 
However, most existing techniques were designed for specialized applications and are difficult to interpret in more general settings \cite{chari2023specious,ma2024principled,antonsson2024batch}. Moreover, their analytical properties are often not well understood, especially under challenging conditions involving data-specific distortions or imbalances in signal magnitude and sample size. 

\begin{figure} 
  \centering
    \includegraphics[width=1\textwidth]{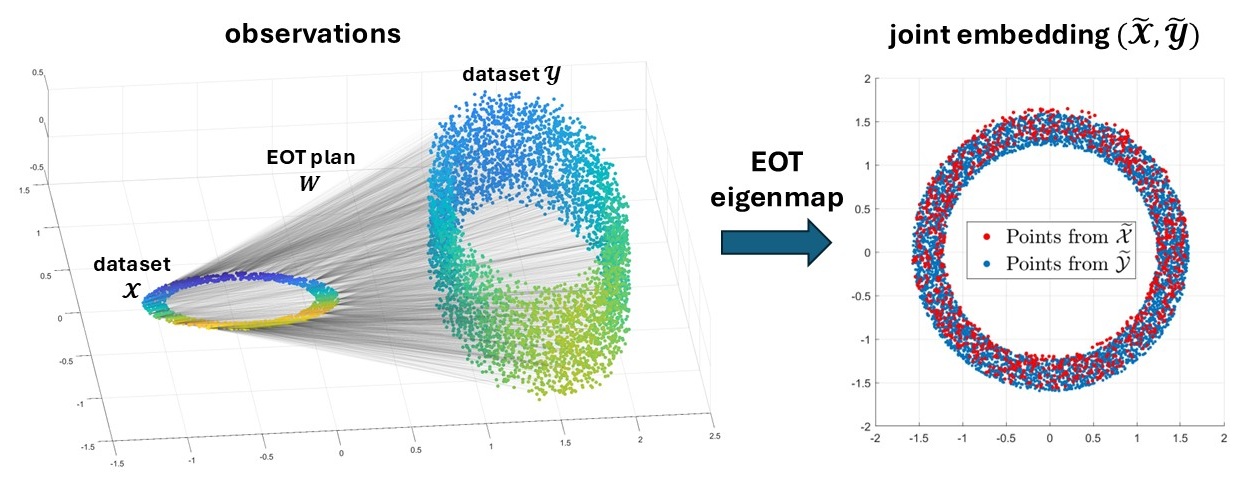} 
    \caption
    {\footnotesize 
    {Datasets $\mathcal{X}$ and $\mathcal{Y}$ in $\mathbb{R}^3$ (left) and their joint embedding $(\widetilde{\mathcal{X}},\widetilde{\mathcal{Y}})$ in $\mathbb{R}^2$ (right) obtained via EOT eigenmaps. The dataset $\mathcal{X}$ consists of $m=1{,}000$ points sampled from a base annulus in the $xy$-plane and then compressed by a factor of $2$ along the $y$-axis. The dataset $\mathcal{Y}$ consists of $n=5{,}000$ points sampled from the same base annulus, then shifted and stretched by a factor of $2$ along the $y$-axis, with additional random variation in the $z$-direction. The transport plan $W\in\mathbb{R}^{m\times n}$ encodes cross-dataset affinities between points in $\mathcal{X}$ and $\mathcal{Y}$. Our method embeds $\mathcal{X}$ (respectively $\mathcal{Y}$) into $\mathbb{R}^2$ using the second and third left (respectively right) singular vectors of $W$, with appropriate scaling (see eq.~\eqref{eq: embedding formula} in Section~\ref{sec: the proposed method}, with $q=2$ and $t=0$). In the left panel, points in $\mathcal{X}$ and $\mathcal{Y}$ are colored by the corresponding second singular vector entries, illustrating the learned correspondence across datasets. The resulting joint embedding recovers the shared annulus in the $xy$-plane and aligns $\mathcal{X}$ and $\mathcal{Y}$ despite the dataset-specific translation, deformations, and out-of-plane variation.}
    } 
    \label{fig: intro illustration}
    \end{figure} 

\subsection{Our approach and contributions}
We propose a principled approach for aligning high-dimensional datasets $\mathcal{X}$ and $\mathcal{Y}$ in $\mathbb{R}^p$ and embedding them jointly into $\mathbb{R}^q$. Our approach, termed \textit{Entropic Optimal Transport eigenmaps}, is  interpretable and  amenable to theoretical analysis under challenging conditions frequently encountered in real-world applications. 

Our technique relies on Entropic Optimal Transport (EOT) between distributions~\cite{cuturi2013sinkhorn,peyre2019computational}---a regularized variant of classical Optimal Transport (OT)~\cite{villani2009optimal}. In particular, we utilize the entropic transport plan between the datasets $\mathcal{X}$ and $\mathcal{Y}$ in $\mathbb{R}^p$, represented by a nonnegative matrix $W\in\mathbb{R}^{m\times n}$. This matrix describes a coupling between the datasets $\mathcal{X}$ and $\mathcal{Y}$ and encodes the inter-data pairwise affinities between them. We use the leading nontrivial singular vectors of $W$ after a suitable rescaling to embed $\mathcal{X}$ and $\mathcal{Y}$ into $\mathbb{R}^q$; see Section~\ref{sec: the proposed method} for a description of EOT and our method (Algorithm~\ref{alg:EOT eigenmaps}). Figure~\ref{fig: intro illustration} exemplifies our setup and proposed approach in a simple simulation.

{First, in Sections~\ref{sec: inter-data Laplacian eigenmaps} and~\ref{sec: inter-data diffusion maps}, we derive basic spectral properties of our approach and present several useful interpretations that establish its connections to Laplacian eigenmaps~\cite{belkin2003laplacian} and diffusion maps~\cite{coifman2006diffusionMaps}, showing that our approach enjoys many similar properties.
In particular, we explain how the embedded datasets---using the singular vectors of the EOT plan $W$---reflect the inter-dataset affinities encoded by $W$ in a geometrically meaningful way, analogously to the embedding of a single dataset via the eigenvectors of the graph Laplacian.}

{Then, in Section~\ref{sec: analysis in the latent manifold model}, we analyze a data-generative model in which the datasets share a common low-dimensional manifold $\mathcal{M}$ governed by latent variables of interest, but each dataset individually exhibits distinct geometric distortion, translation, orthogonal nuisance structures, and sub-Gaussian noise. The geometric distortions are described by commuting positive-definite matrices applied to the latent space coordinates. The noise in our setting can be strong, dependent, and heteroskedastic in a high ambient dimension; see Section~\ref{sec: model and assumptions} for a detailed description of our model and assumptions.}

\begin{figure} 
  \centering
    \includegraphics[width=1\textwidth]{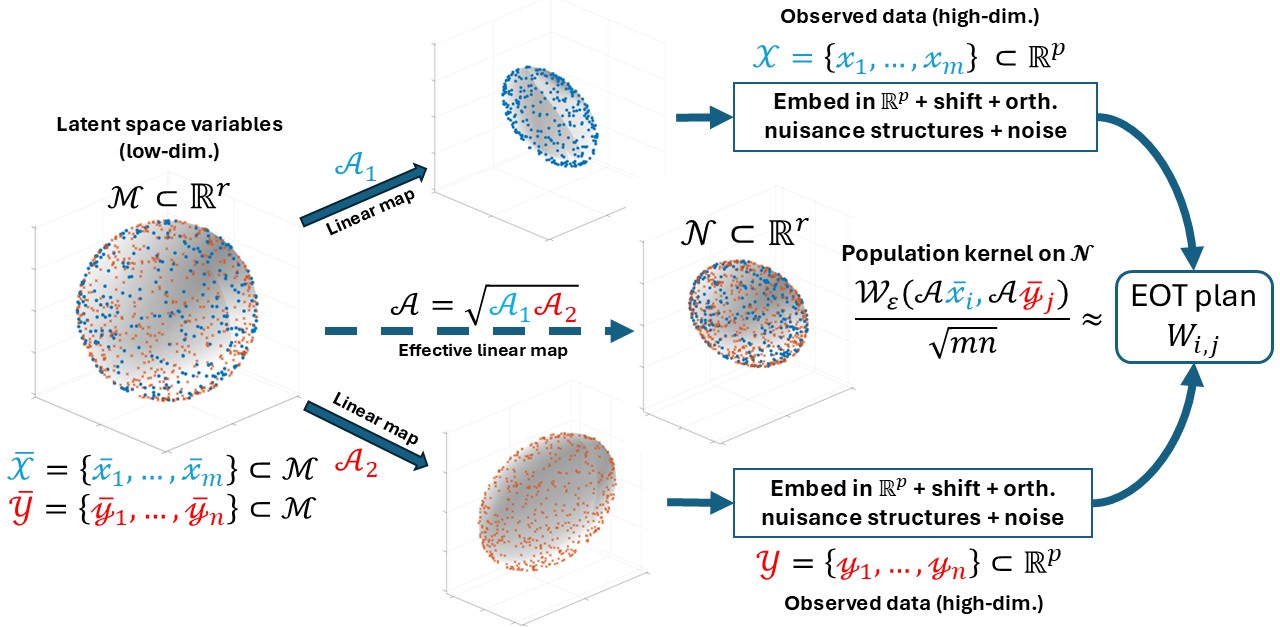} 
    \caption
     {\footnotesize 
Latent datasets $\overline{\mathcal{X}}=\{\bar x_1,\ldots,\bar x_m\}$ and
$\overline{\mathcal{Y}}=\{\bar y_1,\ldots,\bar y_n\}$ are sampled from a shared low-dimensional manifold
$\mathcal{M}\subset \mathbb{R}^r$.
They are distorted by commuting positive-definite linear maps $\mathcal{A}_1$ and $\mathcal{A}_2$, and then embedded into
$\mathbb{R}^p$ with individual translation, orthogonal nuisance structures, and noise, producing the observed datasets
$\mathcal{X}=\{x_i\}_{i=1}^m\subset \mathbb{R}^p$ and $\mathcal{Y}=\{y_j\}_{j=1}^n\subset \mathbb{R}^p$;
see~\eqref{eq: shared manifold observation model}.
Theorem~\ref{thm: concentration of transport plan} shows that the EOT plan $W=(W_{ij})$ computed between
$\mathcal{X}$ and $\mathcal{Y}$ admits a population description: in the large-sample, high-dimensional regime,
a suitably rescaled version of $W_{ij}$ is well-approximated by a population kernel (see~\eqref{eq: W integral def})
evaluated at the coordinates $\mathcal{A}\bar x_i$ and $\mathcal{A}\bar y_j$ on the effective manifold
$\mathcal{N}:=\mathcal{A}(\mathcal{M})$, where $\mathcal{A}=(\mathcal{A}_1\mathcal{A}_2)^{1/2}$.
Equivalently, the EOT plan concentrates around a kernel on $\mathcal{N}$, revealing the shared geometry induced by the
geometric mean of the two distortions, regardless of the translations, orthogonal nuisance structures, and noise.
    } 
    \label{fig: model and results illustration}
    \end{figure}

{Our main theoretical result (Theorem~\ref{thm: concentration of transport plan}) shows that for a fixed entropic regularization, large sample sizes $m$ and $n$, and high dimension $p$, the EOT plan $W$ concentrates around a doubly stochastic Gaussian kernel (see~\eqref{eq: W integral def} and~\eqref{eq:integral scaling eq with density}) evaluated at the locations of the shared latent variables mapped from $\mathcal{M}$ to a new manifold $\mathcal{N}$. This new manifold merges the geometric distortions across the datasets and represents the effective joint structure captured by the EOT plan; see Section~\ref{sec: concentration of EOT plan}. In particular, the distance metric changes from the standard Euclidean distance to a Mahalanobis distance that depends on the geometric mean of the individual (dataset-specific) distortions. This population form is invariant to the translations, orthogonal nuisance structures, and noise in the model, and only changes the geometry of the shared latent manifold. Our result is stated as a probabilistic bound that provides convergence rates in terms of $m$, $n$, $p$, and the sub-Gaussian norm of the noise. Overall, our result explains how the EOT plan reflects the geometry of two deformed datasets that share a common low-dimensional latent structure, and provides robustness guarantees under translations, orthogonal nuisance structures, and high-dimensional sub-Gaussian noise. Figure~\ref{fig: model and results illustration} schematically illustrates our generative data model and main result.}

{Building on this result, in Section~\ref{sec: population interpretation of embedding under latent manifold model} we relate our embedding to a population-level integral operator that describes a random walk between two copies of the new manifold $\mathcal{N}$. The eigenfunctions of this operator provide a population interpretation to our embedding, explaining how it integrates the two datasets according to their intrinsic latent variables. 
Additionally, in Section~\ref{sec: behavior under small bandwidth}, we consider the case of diminishing entropic regularization for sufficiently large sample sizes and high dimension. In this case, we show that the EOT plan approximates the Gaussian kernel normalized symmetrically by the square-root density on the effective manifold $\mathcal{N}$ (see~\eqref{eq: W_tilde_hat kernel def}). This is the kernel used by the so-called symmetric normalized graph Laplacian, which is commonly utilized for spectral clustering and manifold learning~\cite{von2007tutorial,hein2007graph,hoffmann2022spectral,trillos2021geometric}. Subsequently, we relate our embedding in this case to the weighted manifold Laplacian on $\mathcal{N}$ and its eigenfunctions~\cite{grigor2006heat}.}

Lastly, in Sections~\ref{simu.sec} and~\ref{sec: applications to single-cell data integration}, we demonstrate the practical advantages of our approach on simulated and real biological data. For simulated data, our methods outperform alternatives in manifold alignment with noisy observations and joint clustering of datasets with shared structures. In real data, they provide better and more reliable alignment of diverse single-cell omics datasets. The R codes of our proposed methods and those used to generate our numerical results are accessible at our GitHub repository \url{https://github.com/rongstat/EOT-eigenmap}. 

\subsection{Related work} \label{sec: related work}
$\vphantom{a}$

\textbf{Spectral methods for nonlinear data analysis.} Many popular methods for nonlinear data analysis utilize the eigen-decomposition of operators constructed from data, such as affinity matrices and graph Laplacians, for low-dimensional embedding and clustering; see, e.g.,~\cite{hein2005graphs,ng2002spectral,belkin2003laplacian,coifman2006diffusionMaps,zelnik2004self}. The theoretical analysis supporting these methods involves establishing the convergence of the discrete (finite-sample) operator in a suitable sense to a corresponding population analog---typically an integral or differential operator~\cite{singer2006graph,hein2007graph,berry2016variable,cheng2022eigen,trillos2020error,trillos2021geometric,calder2022improved,cheng2022convergence}. A common underlying assumption is the so-called latent manifold model, where the data points are sampled independently from a low-dimensional Riemannian manifold embedded in the ambient space (or a union of several such manifolds), possibly under noise~\cite{el2010information,landa2021doubly,landa2023robust}. 
{Existing methods and theoretical results primarily focus either on the case of a single cohesive dataset or the case of multi-modal data, where the datasets correspond to different views (or modalities) of the same set of samples~\cite{talmon2019latent,lindenbaum2020multi,lederman2018learning}. This setting assumes the same number of samples in each modality and access to a registration between corresponding points.}

{In contrast,  we consider the setting of two unregistered datasets with arbitrary sample sizes. Our results are  related to those of classical Laplacian-based techniques, but are adapted to the case of rectangular cross-dataset affinity matrices, especially those arising from EOT. In particular, we establish the convergence of the EOT plan and related finite-sample operators to population analogs in a high-dimensional scenario involving substantial dataset-specific deformations and sub-Gaussian heteroskedastic noise. These are challenging conditions under which existing Laplacian-based techniques and theories do not apply. Nonetheless, we show that our approach is capable of recovering a shared manifold structure from the corrupted datasets.} 
Note that if the two datasets in our setup are identical, i.e., $\mathcal{X} = \mathcal{Y}$, then our approach reduces to the case of the doubly stochastic Gaussian kernel computed from a single dataset (see Section~\ref{sec: approach and properties}), for which specialized results have been established~\cite{marshall2019manifold,landa2021doubly,wormell2021spectral,landa2023robust}.

\textbf{Entropic Optimal Transport.}
Optimal Transport (OT) characterizes differences between distributions by finding a minimum-cost coupling.
Entropic OT (EOT) approximates OT by adding an entropic regularization term (see \eqref{eq: optimal transport optimization problem} in Section~\ref{sec: the proposed method}),
which yields a unique plan that can be computed efficiently via the Sinkhorn--Knopp algorithm~\cite{cuturi2013sinkhorn,sinkhorn1967concerning}
and has numerous applications~\cite{peyre2019computational}.
In our setup, the EOT plan between $\mathcal{X}$ and $\mathcal{Y}$ (viewed as empirical measures) produces $W\in\mathbb{R}^{m\times n}$; for a non-vanishing regularization level, $W$ acts as a smooth and numerically stable cross-dataset affinity.

The analytical properties of EOT have been extensively studied, including convergence of the entropic cost/plan to their OT counterparts as the regularization vanishes~\cite{nutz2022entropic,carlier2017convergence},
and convergence from finite samples to population analogs as sample sizes grow~\cite{mena2019statistical,goldfeld2024limit,rigollet2025sample,chewi2025statisticalOT}.
These results are typically formulated at the level of the ambient sampling distributions.
In our data-integration setting, however, dataset-specific deformations and corruptions can alter the ambient measures substantially, so the continuum entropic plan couples potentially very different distributions and need not transparently reflect the shared latent geometry.
Thus, for our setup, generic convergence statements alone do not explain how the common latent manifold structure is encoded in $W$ relative to dataset-specific deformations and corruptions.

EOT has also been used to estimate transfer operators of dynamical systems from data~\cite{junge2024entropic,beier2025transfer,koltai2021transfer}.
In particular, Koltai et al.~\cite{koltai2021transfer} use the {unbalanced} EOT plan to register two measures obtained by sampling a dynamical system at two time points. Their goal is to approximate a transfer (Perron--Frobenius) operator, motivated by a Brownian-motion prior, and they use the leading singular-vector pair of the resulting operator to identify coherent sets via entrywise thresholding. Our approach is related to~\cite{koltai2021transfer} at the level of using singular vectors of an entropic-OT-derived operator that induces a registration, but differs in goal, analysis, and algorithmic aspects. Specifically, we study the {balanced} EOT plan as a mechanism for jointly embedding two high-dimensional point clouds that are assumed to share a common low-dimensional geometric structure. We use {multiple} singular-vector pairs to define this embedding, and our theory is developed under a latent-manifold model with dataset-specific deformations and corruptions. Importantly, in our framework, the EOT plan consolidates these effects in a way that enables a population-level characterization of the recovered geometry---features that are not the focus of~\cite{koltai2021transfer}.

Lastly, we note that the entropic regularization in our setup serves a role beyond computational/analytical convenience: it enables a stable, dense coupling that preserves local geometric information needed for a meaningful spectral embedding.
In contrast, unregularized OT typically yields highly sparse couplings that may discard geometric structure, akin to using a 1-NN graph for embedding a manifold.

\textbf{Manifold alignment.} Our problem setup is also closely related to manifold alignment~\cite{wang2011manifold,ham2003learning}, where the goal is to align and jointly embed two datasets from possibly different feature spaces. Existing methods include semi-supervised algorithms that use (partially) labeled data for alignment~\cite{wang2008manifold,tuia2016kernel,duque2023diffusion,correa2023manifold} and unsupervised algorithms~\cite{cui2014generalized,wang2009manifold,lafon2006data,wang2009general,stanley2020harmonic} that rely on the local geometric structure of individual datasets to align them in the embedding space. {We note that several unsupervised approaches also incorporated OT and EOT cost functions into their proposed optimization problems for manifold alignment~\cite{chen2023unsupervised,grave2019unsupervised}. }

While existing unsupervised methods for manifold alignment can accommodate general setups---where datasets belong to different feature spaces (and thus the inter-data pairwise similarities cannot be directly evaluated)---they often lack population-level interpretations and theoretical guarantees. Moreover, relying on the local geometric structure of individual datasets can be prohibitive, particularly when the datasets are subject to individual nuisance structures or high-dimensional noise.
Here, we focus on the setting where the datasets are given in the same feature space with no labeled data, where our approach provides rigorous theoretical guarantees and population interpretations in challenging high-dimensional scenarios involving deformations, orthogonal nuisance structures, and noise.
    
\textbf{Landmark diffusion.} A class of closely related landmark-based embedding methods~\cite{pham2018large,shen2022scalability} were developed to improve the scalability of spectral embedding techniques such as diffusion maps and Laplacian eigenmaps for a single dataset. The key idea is to select a small set of \emph{landmarks} and construct a bipartite random walk between the full sample and these landmarks, yielding a (rectangular) affinity matrix that connects data points to landmarks. The embedding is then obtained from the singular vectors of this matrix, which serves as a computationally efficient surrogate for the diffusion maps embedding that uses all pairwise affinities. Existing analyses assume that both the data and the landmarks are sampled from the same distribution, typically supported on a (possibly noisy) low-dimensional manifold, and relate the landmark-based construction to the corresponding population diffusion operator.

Although these methods are designed for a single dataset, they can be adapted to our two-dataset setting by taking one of the datasets as the landmark set. At this level, both approaches can be viewed as instances of bipartite-graph spectral embedding. The key difference lies in the choice of cross-dataset weights and, consequently, in the performance and theory. Landmark-based approaches rely on traditional normalizations of the Gaussian kernel (e.g., row/column  stochastic and their square-root variants) that can be sensitive to dataset-specific deformations and corruptions. For instance, in the setup of Figure~\ref{fig: intro illustration}, the shift causes the points in the right part of $\mathcal{X}$ to be closer (in ambient distance) to the points in the left part of $\mathcal{Y}$, so a naive distance-based kernel may produce systematically incorrect correspondences. In contrast, we utilize the EOT plan, which is an implicit bi-diagonal normalization of the Gaussian kernel under marginal constraints. This choice is important, both theoretically and empirically, for aligning the two datasets in our setting and recovering the shared latent geometry under deformations and corruptions. In particular, the behavior of the scaling factors $\alpha$ and $\beta$ play a key role in proving our theoretical results in Section~\ref{sec: analysis in the latent manifold model}. We demonstrate these differences empirically in Sections~\ref{simu.sec} and~\ref{sec: applications to single-cell data integration}, where we compare against landmark-based baselines and highlight the advantages of EOT eigenmaps on both synthetic and real data.

\section{The method, spectral properties, and interpretations} \label{sec: approach and properties}
\subsection{EOT and the proposed embedding} \label{sec: the proposed method}
Let $\mathcal{X} = \{x_1,\ldots,x_m\} \subset \mathbb{R}^p$ and \sloppy $\mathcal{Y} = \{y_1,\ldots,y_n\} \subset \mathbb{R}^p$ be two given datasets, assuming without loss of generality that $m\leq n$ (otherwise, we can always interchange the roles of $\mathcal{X}$ and $\mathcal{Y}$). We denote the affinity between $x_i$ and $y_j$ as $W_{i,j}$, where $W\in\mathbb{R}^{m\times n}$ is the entropic optimal transport (EOT) plan between $\mathcal{X}$ and $\mathcal{Y}$ using the Euclidean distance~\cite{peyre2019computational}. Specifically,
\begin{equation}
    W = \underset{W^{'}\in\mathcal{B}_{m,n}}{\operatorname{argmin}} \left\{ \sum_{i=1}^m \sum_{j=1}^n \Vert x_i - y_j \Vert_2^2 W^{'}_{i,j} +\varepsilon\sum_{i=1}^m \sum_{j=1}^n W^{'}_{{i,j}} \log W^{'}_{i,j} \right\}, \label{eq: optimal transport optimization problem}
\end{equation}
where $\varepsilon \geq 0$ is the regularization parameter and $\mathcal{B}_{m,n}$ is the set of feasible transport plans, 
\begin{equation}
    \mathcal{B}_{m,n} = \left\{ A\in\mathbb{R}^{m\times n}: A_{i,j} \geq 0, \;\; \sum_{k=1}^n A_{i,k} = \sqrt{\frac{n}{m}}, \;\; \sum_{\ell=1}^m A_{\ell,j} = \sqrt{\frac{m}{n}}, \;\; i\in [m], \;\; j\in [n] \right\}. \label{eq: B def}
\end{equation}
We remark that the definition of \( \mathcal{B}_{m,n} \) above adopts a specific normalization in which the entries of the transport plan sum to \( \sqrt{mn} \), rather than $1$ as is common in other works. {This choice ensures that the largest singular value of the resulting plan \( W \) is $1$ (see the discussion below~\eqref{eq: SVD of W}), which is convenient for the presentation and analysis of our proposed approach.}

\begin{algorithm}
\caption{EOT eigenmaps}\label{alg:EOT eigenmaps}
\begin{algorithmic}[1]
\Statex{\textbf{Input:} Datasets $\mathcal{X} = \{x_1,\ldots,x_m\} \subset \mathbb{R}^p$ and $\mathcal{Y} = \{y_1,\ldots,y_n\} \subset \mathbb{R}^p$ with $m\leq n$, embedding dimension $q\leq m-1$, EOT regularization parameter $\varepsilon>0$, and $t\geq 0$.}
\State \label{alg: step 1}Compute the EOT plan $W\in\mathbb{R}^{m\times n}$ defined in~\eqref{eq: optimal transport optimization problem}, e.g., by scaling the rows and columns of the Gaussian kernel matrix $K$ from~\eqref{eq: W and K def} using the Sinkhorn-Knopp algorithm~\cite{sinkhorn1967concerning} such that each row sums to $\sqrt{n/m}$ and each column sums to $\sqrt{m/n}$.
\State \label{alg: step 2}Compute the leading $q$ nontrivial left singular vectors, right singular vectors, and singular values of $W$, denoted by $\{\mathbf{u}_2,\ldots,\mathbf{u}_{q+1}\}$, $\{\mathbf{v}_2,\ldots,\mathbf{v}_{q+1}\}$, and $\{s_2,\ldots,s_{q+1}\}$, respectively. 
\State \label{alg: step 3}Form the embedded datasets $\widetilde{\mathcal{X}} = \{\widetilde{x}_1,\ldots,\widetilde{x}_m\} \subset \mathbb{R}^{q}$ and $\widetilde{\mathcal{Y}} = \{\widetilde{y}_1,\ldots,\widetilde{y}_n\} \subset \mathbb{R}^{q}$ according to~\eqref{eq: embedding formula}.
\end{algorithmic}
\end{algorithm}

When $\varepsilon=0$, the matrix $W$ describes the exact optimal plan to redistribute mass from a uniform distribution on $\mathcal{X}$ to a uniform distribution on $\mathcal{Y}$ or vice versa, minimizing the total transport cost between the source and target points as measured by the squared Euclidean distance. In this context, $W_{i,j}/\sqrt{mn}$ is the relative mass transported from $x_i$ to $y_j$ (or vice versa). The resulting transport plan is generally sparse and may not be unique. {When $\varepsilon>0$, the entropic regularization term $\varepsilon\sum_{i=1}^m \sum_{j=1}^n W^{'}_{i,j} \log W^{'}_{i,j}$ promotes smoothness in the transport plan and forces the redistribution of mass to be less concentrated and more spread out. In this case, the regularized transport cost minimized in~\eqref{eq: optimal transport optimization problem} is strictly convex, and the optimal transport plan $W$ is unique. Importantly, the solution can be expressed explicitly by diagonally scaling the rows and columns of a cross-dataset Gaussian kernel matrix with bandwidth parameter $\varepsilon$~\cite{cuturi2013sinkhorn,peyre2019computational}. }Specifically,
\begin{equation}
    W_{i,j} = \alpha_i K_{i,j} \beta_j, \qquad \qquad K_{i,j} = \exp\left\{ - \frac{\Vert x_i - y_j \Vert^2}{\varepsilon} \right\}, \label{eq: W and K def} 
\end{equation}
where $\mathbf{\alpha} = [\alpha_1,\ldots,\alpha_m]$ and $\mathbf{\beta} = [\beta_1,\ldots,\beta_n]$ are chosen such that $W\in \mathcal{B}_{m,n}$, i.e., such that each row of $W$ sums to $\sqrt{n/m}$ and each column of $W$ sums to $\sqrt{m/n}$. 
We note that the vectors $\alpha$ and $\beta$ are directly related to the so-called dual potentials of the entropic OT problem~\cite{peyre2019computational} (which are typically defined as $\mathbf{f}_i = \varepsilon \log \alpha_i$ and $\mathbf{g}_j = \varepsilon \log \beta_j$). These vectors $\alpha$ and $\beta$ always exist and are unique up to a trivial scalar ambiguity. Moreover, they can be computed efficiently by the classical Sinkhorn-Knopp algorithm~\cite{sinkhorn1964relationship,sinkhorn1967concerning}; see also~\cite{allen2017much,lin2022efficiency} for more recent algorithmic developments. From this point on, we assume that $\varepsilon>0$.

We denote the (compact) singular value decomposition (SVD) of $W$ as
\begin{equation}
    W = USV^T, \qquad U = [\mathbf{u}_1,\ldots,\mathbf{u}_m], \qquad V = [\mathbf{v}_1,\ldots,\mathbf{v}_m], \qquad S = \operatorname{diag}\{s_1,\ldots,s_m\}, \label{eq: SVD of W}
\end{equation}
where $\{\mathbf{u}_i\}_{i=1}^m \in\mathbb{R}^m$, $\{\mathbf{v}_i\}_{i=1}^m \in\mathbb{R}^n$, and $\{s_i\}_{i=1}^m$ are the left singular vectors, right singular vectors, and singular values (sorted in descending order) of $W$, respectively. Since $W$ is entrywise positive and in $\mathcal{B}_{m,n}$, the first pair of singular vectors of $W$ is trivial~\cite{horn1994topics}: $\mathbf{u}_1 =  \mathbf{1}_m / \sqrt{m}$, $\mathbf{v}_1 =  \mathbf{1}_n/\sqrt{n}$, and $1 = s_1 > s_2$, where $\mathbf{1}_m$ ($\mathbf{1}_n$) is a vector of $m$ ($n$) ones.

We propose to embed $\mathcal{X}$ and $\mathcal{Y}$ into $\mathbb{R}^q$ using the leading $q$ nontrivial left and right singular vectors of $W$, respectively, up to a suitable scaling. Specifically, given a desired embedding dimension $q \leq m-1$ and a parameter $t\geq 0$, the embedded datasets $\widetilde{\mathcal{X}} = \{\widetilde{x}_1,\ldots,\widetilde{x}_m\} \subset \mathbb{R}^{q}$ and $\widetilde{\mathcal{Y}} = \{\widetilde{y}_1,\ldots,\widetilde{y}_n\} \subset \mathbb{R}^{q}$ are defined by
    \begin{equation}
        \widetilde{x}_i = \sqrt{m}
        \begin{bmatrix}
            s_2^t\mathbf{u}_2[i]\\ \vdots\\ s_{q+1}^t\mathbf{u}_{q+1}[i]
        \end{bmatrix},
        \qquad\qquad
        \widetilde{y}_j = \sqrt{n}
        \begin{bmatrix}
            s_2^t\mathbf{v}_2[j]\\ \vdots\\ s_{q+1}^t\mathbf{v}_{q+1}[j]
        \end{bmatrix}, \label{eq: embedding formula}
    \end{equation}
for $i=1,\ldots,m$ and $j=1,\ldots,n$, where $u_k[i]$  ($v_k[j]$) denotes the $i$'th ($j$'th) coordinate of $u_k$ ($v_k$). The proposed method, termed \textit{EOT eigenmaps}, is summarized in Algorithm~\ref{alg:EOT eigenmaps}. The parameter $t$ controls the variance of the embedding coordinates;  for $t=0$ all coordinates have an equal variance of $1$ across each dataset, while for $t>0$ coordinates that correspond to smaller singular values of $W$ are attenuated, exhibiting a smaller variance that decays more rapidly as $t$ increases.

Next, we derive basic spectral properties for our approach and present its interpretation in two special cases: \( t = 0 \) and integers \( t > 0 \). This enables us to provide insight into how our embedding reflects inter-data relationships and connects to classical techniques, particularly Laplacian eigenmaps~\cite{belkin2003laplacian} and diffusion maps~\cite{coifman2006diffusionMaps}, which have been extensively studied for embedding a single dataset. In particular, the parameters \( (\varepsilon, q, t) \) in our approach are analogous to the kernel bandwidth, embedding dimension, and diffusion time in diffusion maps; see Section~\ref{sec: inter-data diffusion maps}. Additional discussion on how to choose these parameters appears in Appendix~\ref{supp.simu.sec}.
{We note that the results below in Sections~\ref{sec: inter-data Laplacian eigenmaps} and~\ref{sec: inter-data diffusion maps} hold for general matrices in the class $\mathcal{B}_{m,n}$ in~\eqref{eq: B def}, not just for EOT plans. In Section~\ref{sec: analysis in the latent manifold model}, we analyze the EOT plan from~\eqref{eq: optimal transport optimization problem}  under a generative model with structured distortions, and show that \( W \) accurately captures the underlying inter-data affinities, making it well-suited for our spectral embedding framework.}

 \subsection{The case of $t=0$: inter-data Laplacian eigenmaps} \label{sec: inter-data Laplacian eigenmaps}
For the case of $t=0$, we interpret our approach as an inter-data variant of Laplacian eigenmaps~\cite{belkin2003laplacian}. 
We show that our proposed embedding solves an optimal data alignment problem closely related to the one solved by Laplacian eigenmaps for a single dataset. We then establish that our embedding can be obtained from the eigenvectors of a certain \textit{inter-data graph Laplacian}. We characterize the properties of this graph Laplacian to obtain useful interpretations of our approach. 

We begin by showing that our proposed method aligns two datasets optimally in the embedding space according to the pairwise affinities encoded by $W$. In particular, the low-dimensional embedded datasets $\widetilde{\mathcal{X}}$ and $\widetilde{\mathcal{Y}}$ achieve the minimal transport cost in $ \mathbb{R}^q$ under the transport plan $W$ from~\eqref{eq: W and K def} and the constraints that the coordinates of each embedded dataset have zero means, unit variances, and are pairwise uncorrelated (orthogonal). Concretely, consider the objective function
\begin{equation}
J(\mathcal{X}^{'},\mathcal{Y}^{'}) = \sum_{i=1}^m \sum_{j=1}^n \Vert {x}_i^{'} - {y}_j^{'} \Vert_2^2 W_{i,j}, \label{eq: cost function for embedding}
\end{equation}
for $\mathcal{X}^{'} = \{x_1^{'},\ldots,x_m^{'}\}\subset \mathbb{R}^q$ and $\mathcal{Y}^{'} = \{y_1^{'},\ldots,y_n^{'}\}\subset \mathbb{R}^q$, with $q\leq m-1$, and the constraints
\begin{equation}
    \frac{1}{m}\sum_{i=1}^m x_i^{'}[k] = 0, \qquad \frac{1}{n}\sum_{j=1}^n y_j^{'}[k] = 0, \qquad
    \frac{1}{m}\sum_{i=1}^m x_i^{'}[k] x_i^{'}[\ell]  = \delta_{k,\ell},  \qquad  \frac{1}{n}\sum_{j=1}^n y_j^{'}[k] y_j^{'}[\ell] = \delta_{k,\ell}, \label{eq: constraints for embedding}
\end{equation}
for all $k,\ell=1,\ldots,q$, where $\delta_{k,\ell}$ is the Kronecker delta ($\delta_{k,\ell} = 1$ if $k=\ell$ and $\delta_{k,\ell}=0$ otherwise). Note that the objective function in~\eqref{eq: cost function for embedding} has the same form as the one being minimized in~\eqref{eq: optimal transport optimization problem} to obtain the EOT plan but without the entropic regularization term---this term does not influence the minimization over $\mathcal{X}^{'}$ and $\mathcal{Y}^{'}$ since in~\eqref{eq: cost function for embedding} $W$ is given. In particular, if the two datasets are identical, i.e., $\mathcal{X} = \mathcal{Y}$, then the cost function~\eqref{eq: cost function for embedding} is the same as the one minimized in Laplacian eigenmaps~\cite{belkin2003laplacian}, except that the doubly stochastic affinity matrix $W$~\cite{zass2007doubly,marshall2019manifold,landa2021doubly} here replaces the traditional (un-normalized) Gaussian kernel matrix used in Laplacian eigenmaps.
The following proposition, whose proof can be found in Appendix~\ref{appendix: proof of proposition on embedding formula for two datasets}, establishes the optimality of~\eqref{eq: embedding formula} with $t=0$ for aligning the embedded datasets with respect to $W$.
\begin{prop} \label{prop: embedding formula}
    Under the constraints~\eqref{eq: constraints for embedding}, the function $J(\mathcal{X}^{'},\mathcal{Y}^{'})$ from~\eqref{eq: cost function for embedding} is minimized by $(\widetilde{\mathcal{X}},\widetilde{\mathcal{Y}})$ from~\eqref{eq: embedding formula} with $t=0$.
\end{prop}
We note that this minimizer is not unique and may be subject to rotation and reflection. For instance, replacing $(\mathbf{u}_k,\mathbf{v}_k)$ by $(-\mathbf{u}_k,-\mathbf{v}_k)$ will leave the transport cost unchanged. 

Next, we establish a connection between our embedding for $t=0$ and a certain graph Laplacian-type matrix.
Define the matrices \sloppy $\hat{W},L,D,\widetilde{L}\in\mathbb{R}^{(m+n)\times(m+n)}$ according to
\begin{equation}
    \hat{W} = 
    \begin{bmatrix}
        \mathbf{0}_{m\times m} & W \\
        W^T & \mathbf{0}_{n\times n}
    \end{bmatrix},
    \qquad L = I_{m+n} - \hat{W}, \qquad D = 
     \begin{bmatrix}
        \sqrt{m} I_{m} & \mathbf{0}_{m\times n} \\
        \mathbf{0}_{n\times m} & \sqrt{n} I_{n} 
    \end{bmatrix}, 
    \qquad
    \widetilde{L} = D L D^{-1}, 
    \label{eq: W_hat and L def}
\end{equation}
where $I_{n}$ is the $n\times n$ identity matrix.
The matrix $\hat{W}$ is the adjacency matrix of an undirected bipartite graph $\mathcal{G}$ whose nodes are the points of $\mathcal{X}$ and $\mathcal{Y}$ and the edge weights are given by the EOT plan $W$. Let $\mathbf{f}\in\mathbb{R}^{m+n}$ be a column vector describing a function over the (nodes of the) graph $\mathcal{G}$ and denote $\mathbf{f} = [\mathbf{g}^T, \mathbf{h}^T]^T$, where $\mathbf{g}\in\mathbb{R}^m$ and $\mathbf{h}\in\mathbb{R}^n$ are column vectors interpreted as functions over the graph nodes corresponding to $\mathcal{X}$ and $\mathcal{Y}$, respectively. 
The following proposition, whose proof can be found in Appendix~\ref{appendix: proof of spectrum of L}, provides basic spectral properties of $L$ and $\widetilde{L}$. 
\begin{prop} \label{prop: spectrum of L}
The following holds: 
\begin{enumerate}
    \item $L$ is symmetric and positive semidefinite (PSD) with eigenvalues (sorted in ascending order) \sloppy $\{\lambda_1,\ldots,\lambda_{m+n}\} = \{ 0, 1-{s_2},\ldots,1-{s_m},1,\ldots,1,1+{s_m},\ldots,1+{s_2},2\}$ and corresponding orthonormal eigenvectors \sloppy $\phi_1,\ldots,\phi_{m+n}\in\mathbb{R}^{m+n}$ given in Appendix~\ref{appendix: proof of spectrum of L}. Moreover, $L$ admits the quadratic form
\begin{equation}
    \mathbf{f}^T L \mathbf{f} = \frac{1}{\sqrt{mn}} \sum_{i=1}^m \sum_{j=1}^n \left( \sqrt{m} \mathbf{g}[i] - \sqrt{n} \mathbf{h}[j] \right)^2 W_{i,j}. \label{eq: quadratic form for L}
\end{equation}

\item $\widetilde{L}$ is eigen-decomposable with the same eigenvalues $\{\lambda_k\}$ of $L$ and corresponding eigenvectors $\psi_k = D \phi_k / \Vert D \phi_k\Vert_2$ for $k=1,\ldots,m+n$. Specifically, $\psi_1 = \mathbf{1}_{m+n}/\sqrt{m+n}$, and for $k = 2, \ldots,q +1 \leq m$,
\begin{equation}
     \psi_k = 
     \frac{1}{\sqrt{m+n} }
     \begin{bmatrix}
        \sqrt{m} \mathbf{u}_{k} \\ \sqrt{n} \mathbf{v}_{k} 
    \end{bmatrix} 
    = \frac{1}{\sqrt{m+n} }
    \begin{bmatrix}
        \widetilde{x}_1[k-1] \\
        \vdots \\
        \widetilde{x}_m[k-1] \\
        \widetilde{y}_1[k-1] \\
        \vdots \\
        \widetilde{y}_n[k-1]
    \end{bmatrix}. \label{eq: embedding interpretation with eigenvectors of P}
\end{equation}
\end{enumerate}
\end{prop}

For a given function $\mathbf{f} = [\mathbf{g}^T,\mathbf{h}^T]^T$ on the graph $\mathcal{G}$, the quadratic form in~\eqref{eq: quadratic form for L} quantifies the similarity between $\mathbf{g}$ and $\mathbf{h}$ with respect to the graph edge weights $W$ after rescaling $\mathbf{g}$ and $\mathbf{h}$ by $\sqrt{m}$ and $\sqrt{n}$, respectively. This rescaling is natural for comparing $\mathbf{g}$ and $\mathbf{h}$ if $\Vert \mathbf{g} \Vert_2 = \Vert \mathbf{h} \Vert_2$, since in this case we have $\Vert \sqrt{m} \mathbf{g}\Vert_2^2/m = \Vert \sqrt{n} \mathbf{h} \Vert_2^2/n$, making the individual entries of $\sqrt{m} \mathbf{g}$ and $\sqrt{n} \mathbf{h}$ comparable with the same average magnitude, irrespective of the dataset sizes. 
If the sizes of the datasets are the same, i.e., $m = n$, then $L$ becomes the standard graph Laplacian for the graph $\mathcal{G}$. Otherwise, $L$ can be interpreted as an inter-data graph Laplacian for imbalanced datasets. 

The functions over the graph $\mathcal{G}$ that minimize the quadratic form~\eqref{eq: quadratic form for L} are given by the eigenvectors of $L$ with the smallest eigenvalues. Specifically,
the vectors $\phi_k$ with small indices $k$ constitute orthonormal functions over the graph $\mathcal{G}$, each of the form $[\mathbf{g}^T,\mathbf{h}^T]^T$, such that $\sqrt{m}\mathbf{g}$ and $\sqrt{n}\mathbf{h}$ are similar with respect to the graph weights $W$. These vectors $\sqrt{m}\mathbf{g}$ and $\sqrt{n}\mathbf{h}$ are given precisely by the first $m$ coordinates and the last $n$ coordinates, respectively, of the eigenvectors $\psi_k$ of $\widetilde{L}$ (since $\psi_k$ is proportional to $D \phi_k$).  
Indeed, observe that
\begin{align}
    \frac{m+n}{\sqrt{mn}} \sum_{i=1}^m \sum_{j=1}^n \left( \psi_k[i] - \psi_k[m+j] \right)^2 W_{i,j} 
    &= \frac{1}{\sqrt{mn}} \sum_{i=1}^m \sum_{j=1}^n \left( \sqrt{m}\mathbf{u}_k[i] - \sqrt{n}\mathbf{v}_k[j] \right)^2 W_{i,j} \nonumber \\
    &= {\phi_k}^T L {\phi_k} = \lambda_k. \label{eq: quadratic form for eigenvectors of P}
\end{align}
Therefore, for smaller eigenvalues $\lambda_k$, the two parts of $\psi_k$ (corresponding to $\mathcal{X}$ and $\mathcal{Y}$) are more similar to each other. 

We conclude that the leading nontrivial eigenvectors of $\widetilde{L}$ are the non-constant orthonormal functions over the bipartite graph $\mathcal{G}$ that are the most similar across the two parts of the graph. This property is exemplified in the left side of Figure~\ref{fig: intro illustration}, where the colors represent the values of the leading nontrivial eigenvector of $\widetilde{L}$ (overlayed over all points of the two datasets). 
The leading nontrivial eigenvectors of $\widetilde{L}$ constitute the coordinates of the embedded data $\{\widetilde{\mathcal{X}},\widetilde{\mathcal{Y}}\}$ from~\eqref{eq: embedding formula} for $t=0$ up to the factor $\sqrt{m+n}$.

\subsection{The case of an integer $t>0$: inter-data diffusion maps} \label{sec: inter-data diffusion maps}
We now consider the case of an integer $t>0$ in the embedding~\eqref{eq: embedding formula}. In this case, we interpret our approach as an inter-data variant of diffusion maps~\cite{coifman2006diffusionMaps}.  In this section, we relate our embedding to a random walk on the bipartite graph $\mathcal{G}$ and establish that our embedding preserves a suitable diffusion distance derived from this random walk, where $t$ is the number of steps taken.

We begin by defining the matrix $P\in \mathbb{R}^{(m+n)\times(m+n)}$ as
\begin{equation}
    P = I_{m+n} - \widetilde{L} =  D \hat{W} D^{-1} = 
    \begin{bmatrix}
        \mathbf{0}_{m\times m} & \sqrt{\frac{m}{n}} W \\
        \sqrt{\frac{n}{m}} W^T & \mathbf{0}_{n\times n}
    \end{bmatrix}, \label{eq: P def}
\end{equation}
where $\widetilde{L}$ is from~\eqref{eq: W_hat and L def}. 
Note that the eigenvectors of $P$ are the eigenvectors $\psi_k$ of $\widetilde{L}$ (see~\eqref{eq: embedding interpretation with eigenvectors of P}) with corresponding eigenvalues $\mu_k = 1 - \lambda_k$, namely \sloppy $\{\mu_1,\ldots,\mu_{m+n}\} = \left\{1,{s_2},\ldots,{s_m},0,\ldots,0,-{s_m},\ldots,-{s_2},-1\right\}$. Hence, the embedding~\eqref{eq: embedding formula} is given by the leading $q\leq m-1$ nontrivial eigenvectors of $P$ with largest eigenvalues, which are always nonnegative.
Recall that the sum of each row of $W$ is $\sqrt{n/m}$ and the sum of each column is $\sqrt{m/n}$. Consequently, $P$ is row-stochastic (i.e., the sum of each row of $P$ is $1$) and can be interpreted as the transition probability matrix of a random walk on the bipartite graph $\mathcal{G}$. At each step of this random walk, the random walker transitions from a node in $\mathcal{X}$ to a node in $\mathcal{Y}$ or vice versa according to the transition probability matrices $W\sqrt{m/n}$ or $W^T\sqrt{n/m}$, respectively. The transition probabilities after $t$ consecutive steps of this random walk are 
\begin{equation}
    P^t = 
    \begin{dcases}
        \begin{bmatrix}
        P_\mathcal{XX}^{(t)} & \mathbf{0}_{m\times n} \\
        \mathbf{0}_{n\times m} & P_\mathcal{YY}^{(t)}
        \end{bmatrix}
    & \text{$t$ is even}, \\  
        \begin{bmatrix}
        \mathbf{0}_{m\times m} & P_\mathcal{XY}^{(t)} \\
        P_\mathcal{YX}^{(t)} & \mathbf{0}_{n\times n}
    \end{bmatrix}
    & \text{$t$ is odd},
    \end{dcases} \label{eq: bipartite transition probability matrix}
\end{equation}
where
$
    P_\mathcal{XX}^{(t)} = U S^t U^T,  P_\mathcal{XY}^{(t)} = \sqrt{\frac{m}{n}} U S^t V^T,
    P_\mathcal{YY}^{(t)} = V S^t V^T, P_\mathcal{YX}^{(t)} = \sqrt{\frac{n}{m}} V S^t U^T.
$
The matrices $P_\mathcal{XX}^{(t)}$, $P_\mathcal{XY}^{(t)}$, $P_\mathcal{YX}^{(t)}$, and $P_\mathcal{YY}^{(t)}$, which depend only on the transport plan $W$ and its SVD, describe the probabilities to transition from a node in $\mathcal{X}$ to a node in $\mathcal{X}$, a node in $\mathcal{X}$ to a node in $\mathcal{Y}$, a node in $\mathcal{Y}$ to a node in $\mathcal{X}$, and a node in $\mathcal{Y}$ to a node in $\mathcal{Y}$, respectively, after $t$ steps of the random walk. The $i$'th row of $P^t$ represents the probability distribution of a random walker's location across all nodes after $t$ steps, starting at node $i$.

Given a graph over a set of data points, Coifman and Lafon~\cite{coifman2006diffusionMaps} proposed a robust way to quantify similarity by comparing the distributions of a random walker starting at different points after several steps. Concretely, they define the \emph{diffusion distance} as a weighted Euclidean distance between rows \(i\) and \(j\) of the \(t\)-step transition probability matrix, which captures the extent to which the two starting points ``diffuse'' to the rest of the dataset in a similar way. Here we construct an analogous diffusion distance for the random walk induced by \(P\). Our setting differs from the classical diffusion-maps construction because the underlying graph \(\mathcal{G}\) is bipartite, which makes the associated random walk periodic and requires a corresponding adaptation of the diffusion distance definition.

Given a pair of points $x_i, x_{i^{'}} \in \mathcal{X}$, we initiate random walks at these locations and assess the similarity between $x_i$ and $x_{i^{'}}$ by comparing the corresponding distributions of the random walker's location across $\mathcal{X}$ for even $t$ and across $\mathcal{Y}$ for odd $t$ (since the graph is bipartite). Specifically, we use the distance function
\begin{equation}
    D_{\mathcal{X}}^{(t)}({x}_i,{x}_{i^{'}}) = 
    \begin{dcases}
        \sqrt{m}\left\Vert \left[P_{\mathcal{XX}}^{(t)}\right]_{i,\cdot} - \left[P_{\mathcal{XX}}^{(t)}\right]_{i^{'},\cdot} \right\Vert_2, & \text{$t$ is even}, \\
        \sqrt{n}\left\Vert \left[P_{\mathcal{XY}}^{(t)}\right]_{i,\cdot} - \left[P_{\mathcal{XY}}^{(t)}\right]_{i^{'},\cdot} \right\Vert_2, & \text{$t$ is odd},
    \end{dcases} \label{eq:distance function D_x}
\end{equation}
where the notation $A_{i,\cdot}$ in~\eqref{eq:distance function D_x} refers to the $i$'th row of a matrix $A$. Analogously, for a pair of points $y_j, y_{j^{'}} \in \mathcal{Y}$, we use the distance function
\begin{equation}
    D_{\mathcal{Y}}^{(t)}({y}_j,{y}_{j^{'}}) = 
    \begin{dcases}
        \sqrt{n}\left\Vert \left[P_{\mathcal{YY}}^{(t)}\right]_{j,\cdot} - \left[P_{\mathcal{YY}}^{(t)}\right]_{j^{'},\cdot} \right\Vert_2, & \text{$t$ is even}, \\
        \sqrt{m}\left\Vert \left[P_{\mathcal{YX}}^{(t)}\right]_{j,\cdot} - \left[P_{\mathcal{YX}}^{(t)}\right]_{j^{'},\cdot} \right\Vert_2, & \text{$t$ is odd}.
    \end{dcases} \label{eq:distance function D_y}
\end{equation}
To assess the similarity between a pair of points from distinct datasets, i.e., $x_i\in\mathcal{X}$ and $y_j \in\mathcal{Y}$, we have to account for the fact that $P_{i,\cdot}^t$ and $P_{m+j,\cdot}^t$ have disjoint supports for all $t$. Therefore, these random walk distributions cannot be directly compared using the traditional diffusion distance evaluated from $P^t$. Instead, we propose to use the distance functions
\begin{align} 
    D_{\mathcal{X}\mathcal{Y}}^{(t)}(x_i,y_j) &= 
    \sqrt{n}\left\Vert \left[P_{\mathcal{XY}}^{(t)}\right]_{i,\cdot} - \left[P_{\mathcal{YY}}^{(t)}\right]_{j,\cdot}  \right\Vert_2,\label{eq:distance function D_yx} \\
    D_{\mathcal{Y}\mathcal{X}}^{(t)}(y_j,x_i) &= \sqrt{m}\left\Vert \left[P_{\mathcal{YX}}^{(t)}\right]_{j,\cdot} 
 - \left[P_{\mathcal{XX}}^{(t)}\right]_{i,\cdot} \right\Vert_2, 
 \label{eq:distance function D_xy}
\end{align}
where we extend the definitions of $P_\mathcal{XX}^{(t)}$, $P_\mathcal{XY}^{(t)}$, $P_\mathcal{YX}^{(t)}$, and $P_\mathcal{YY}^{(t)}$ (given after~\eqref{eq: bipartite transition probability matrix}) to all integers $t\geq 0$. 
These distances bypass the aforementioned issue arising from the graph's bipartite nature by comparing random walk distributions from $P^t$ with different parities of $t$.

The following proposition shows that for $q=m-1$ and an integer $t>0$, the Euclidean distance between any pair of points in the embedded data $\{\widetilde{\mathcal{X}},\widetilde{\mathcal{Y}}\}$ is precisely the diffusion distance described in~\eqref{eq:distance function D_x}--\eqref{eq:distance function D_xy}, which constitutes a proper metric over $\{\mathcal{X},\mathcal{Y}\}$.
\begin{prop} \label{prop: diffusion distance identities}
    Let $q=m-1$. Then, for any integer $t>0$, we have 
    \begin{align}
        \Vert \widetilde{x}_i - \widetilde{x}_{i^{'}} \Vert_2 &= D_{\mathcal{X}}^{(t)}({x}_i,{x}_{i^{'}}) 
        = \sqrt{\sum_{k=2}^m {s_k^{2t}}  \left( \sqrt{m} \mathbf{u}_k[i] - \sqrt{m} \mathbf{u}_k[i^{'}] \right)^2}, \label{eq: diffusion distances x-x}\\
        \Vert \widetilde{y}_j - \widetilde{y}_{j^{'}} \Vert_2 &= D_{\mathcal{Y}}^{(t)}({y}_j,{y}_{j^{'}}) 
        = \sqrt{\sum_{k=2}^m {s_k^{2t}} \left( \sqrt{n} \mathbf{v}_k[j] - \sqrt{n} \mathbf{v}_k[j^{'}] \right)^2}, \label{eq: diffusion distances y-y}\\
        \Vert \widetilde{x}_i - \widetilde{y}_{j} \Vert_2 &= D_{\mathcal{XY}}^{(t)}({x}_i,{y}_{j}) = D_{\mathcal{YX}}^{(t)}({y}_j,{x}_{i}) 
        = \sqrt{\sum_{k=2}^m {s_k^{2t}}  \left( \sqrt{m} \mathbf{u}_k[i] - \sqrt{n} \mathbf{v}_k[j] \right)^2}, \label{eq: diffusion distances x-y}
    \end{align}
for all $i,i^{'} \in \{1,\ldots,m\}$ and $j,j^{'}\in \{1,\ldots,n\}$, where $\widetilde{x}_i$ and $\widetilde{y}_j$ are from~\eqref{eq: embedding formula}.
\end{prop}
Hence, the pairwise distances in the embedded space naturally encode the similarities between the points in ambient space according to the bipartite random walk generated by $P$ after $t$ steps. Although the proposition is stated for the embedding dimension $q = m-1$, we show in Appendix~\ref{sec: truncating the diffusion distance} that it also implies an approximate version of this identity for general embedding dimensions $q$, with a controlled approximation error.

\section{Analysis under a latent manifold model with distortions} \label{sec: analysis in the latent manifold model}
In this section, we consider a setup where the points in each dataset are first sampled from a common latent manifold $\mathcal{M}$ in a low-dimensional subspace $\mathbb{R}^r$ and then embedded in a high-dimensional space $\mathbb{R}^p$ with dataset-specific deformations and corruptions. We characterize the large-sample behavior of the matrices $W$ and $P$ from Section~\ref{sec: approach and properties} in this setup and establish their convergence to suitable population quantities. We then analyze these population analogs and their spectral decompositions to justify our approach and explain our embedding. 

\subsection{Model and assumptions} \label{sec: model and assumptions}
Let $\overline{\mathcal{X}} = \{\overline{x}_1,\ldots,\overline{x}_m\} \subset \mathbb{R}^r$ and $\overline{\mathcal{Y}}= \{\overline{y}_1,\ldots,\overline{y}_n \} \subset \mathbb{R}^r$ be i.i.d. samples from a probability measure $\omega(x) d\mu(x)$ supported on a $d$-dimensional compact Riemannian manifold $\mathcal{M} \subset \mathbb{R}^r$, where $\omega(x)$ is a positive and continuous probability density function on $\mathcal{M}$ and $d\mu(x)$ is the volume form of $\mathcal{M}$ at $x\in\mathcal{M}$ (induced by the Euclidean metric in $\mathbb{R}^r$). We consider $\overline{\mathcal{X}}$ and $\overline{\mathcal{Y}}$ as clean datasets containing latent variables of interest. The observed datasets $\mathcal{X} = \{x_1,\ldots,x_m\} \subset \mathbb{R}^p$ and $\mathcal{Y} = \{y_1,\ldots,y_n\}\subset\mathbb{R}^p$ are modeled as
\begin{equation}
    x_i = \nu_1 + \mathcal{U} \mathcal{A}_1 \overline{x}_i + \mathcal{V}_1 z_i^{(1)} + \eta_i^{(1)}, \qquad\qquad 
    y_j = \nu_2 + \mathcal{U} \mathcal{A}_2\overline{y}_j + \mathcal{V}_2 z_j^{(2)} + \eta_j^{(2)}, \label{eq: shared manifold observation model}
\end{equation}
for $i=1,\ldots,m$ and $j=1,\ldots,n$, where:
 $\nu_1,\nu_2\in\mathbb{R}^p$ are arbitrary translation vectors;
 $\mathcal{A}_1,\mathcal{A}_2 \in\mathbb{R}^{r\times r}$ are commuting symmetric positive definite (SPD) matrices describing dataset-specific geometric distortions (and re-scaling);
$\mathcal{U}\in\mathbb{R}^{p\times r}$ is a matrix with orthonormal columns representing a latent subspace shared between the datasets;
$\mathcal{V}_1 \in \mathbb{R}^{p\times r_1}$ and $\mathcal{V}_2 \in \mathbb{R}^{p\times r_2}$ are matrices with orthonormal columns representing orthogonal nuisance subspaces specific to each dataset with $\mathcal{U}^T \mathcal{V}_1 = \mathbf{0}_{r\times r_1}$, $\mathcal{U}^T \mathcal{V}_2 = \mathbf{0}_{r\times r_2}$, and $\mathcal{V}_1^T \mathcal{V}_2 = \mathbf{0}_{r_1\times r_2}$;
$\{z_i^{(1)}\}_{i=1}^m\subset \mathbb{R}^{r_1}$ and $\{z_j^{(2)}\}_{j=1}^n\subset\mathbb{R}^{r_2}$ are arbitrary nuisance variables; and $\{\eta_i^{(1)}\}_{i=1}^m\subset\mathbb{R}^{p}$ and $\{\eta_j^{(2)}\}_{j=1}^n\subset\mathbb{R}^{p}$ are data-specific sub-Gaussian random noise vectors (see~\cite{vershynin2018high}) with zero means, which are pairwise independent between the two datasets, namely $\eta_i^{(1)}$ and $\eta_j^{(2)}$ are independent for all $i,j$.

The model~\eqref{eq: shared manifold observation model} aims to capture several types of deformations and corruptions commonly arising in data integration applications. It is also partly motivated by invariance properties of unregularized optimal transport: for squared Euclidean cost, the OT plan is known to be invariant under translations and dilations of either distribution~\cite{kuang2017preconditioning}, and any transport plan is optimal when the two distributions are supported on orthogonal subspaces~\cite{villani2021topics}.

\begin{myremark}
Model~\eqref{eq: shared manifold observation model} can be equivalently reparameterized by absorbing $\mathcal{A}_1$ into the latent manifold, reducing to $\mathcal{A}_1=I$ and a single relative distortion $\mathcal{B}:=\mathcal{A}_2\mathcal{A}_1^{-1}$ (which is SPD since $\mathcal{A}_1$ and $\mathcal{A}_2$ commute). Under this viewpoint, $\overline{\mathcal{X}}$ is sampled from an arbitrary manifold $\mathcal{M}_1$, while $\overline{\mathcal{Y}}$ is sampled from $\mathcal{M}_2=\mathcal{B}\mathcal{M}_1$. We use the form in~\eqref{eq: shared manifold observation model} with two commuting SPD maps for a symmetric presentation of the model (treating the two datasets the same) while allowing an arbitrary SPD relative distortion between the latent representations.
\end{myremark}

For a sub-Gaussian random vector $\eta\in\mathbb{R}^p$, we denote by $\Vert \eta \Vert_{\Psi_2} = \sup_{\Vert v \Vert_2 = 1} \Vert  \eta^T v \Vert_{\Psi_2}$ its sub-Gaussian norm, where $\Vert \eta^T v \Vert_{\Psi_2}$ is the sub-Gaussian norm of the random variable $\eta^T v$; see~\cite{vershynin2018high}. We define the maximal sub-Gaussian norm among all noise vectors in~\eqref{eq: shared manifold observation model} as $E := \max \{ \Vert \eta_1^{(1)} \Vert_{\Psi_2},\ldots, \Vert \eta_m^{(1)} \Vert_{\Psi_2}, \Vert \eta_1^{(2)} \Vert_{\Psi_2},\ldots, \Vert \eta_n^{(2)} \Vert_{\Psi_2}\}$. We now make the following assumption on the boundedness of the quantities in the model~\eqref{eq: shared manifold observation model}, where $\Vert \cdot \Vert_2$ denotes the standard $\ell^2$ norm for a vector and the spectral (operator) norm for a matrix.
\begin{assump} \label{assump: noise magnitude}
There exists a global constant $C>0$ such that $\Vert \mathcal{A}_1 \Vert_2 \leq C$, $\Vert \mathcal{A}_2 \Vert_2 \leq C$, $\Vert z_i^{(1)} \Vert_2 \leq C$ for  $i=1,\ldots,m$, $\Vert z_j^{(2)} \Vert_2 \leq C$ for  $j=1,\ldots,n$, $\Vert x \Vert_2 \leq C$ for all $x\in \mathcal{M}$, and $E \leq C/(p^{1/4} \sqrt{\log p})$.
\end{assump}
Assumption~\ref{assump: noise magnitude} permits nuisance components in~\eqref{eq: shared manifold observation model} whose magnitudes are comparable to, or larger than, the latent manifold signal; the same is true for the noise vectors $\eta_i^{(1)}$ and $\eta_j^{(2)}$. For example, if $\eta_i^{(1)},\eta_j^{(2)}\sim\mathcal{N}(\mathbf{0}_p,\Sigma^2)$ with $\Sigma^2=I_p/(p^{1/2}\log p)$, then the bound on $E$ holds (since for a Gaussian vector, $\|\eta\|_{\Psi_2}$ is proportional to the largest eigenvalue of $\Sigma$; see~\cite{vershynin2018high}), while $\mathbb{E}\|\eta_i^{(1)}\|_2^2=\mathbb{E}\|\eta_j^{(2)}\|_2^2=\mathrm{Trace}\{\Sigma^2\}=\sqrt{p}/\log p$, which grows unbounded with $p$, whereas $\|\overline{x}_i\|_2\le C$ and $\|\overline{y}_j\|_2\le C$ remain uniformly bounded. Finally, the noise vectors $\eta^{(1)}_1,\ldots,\eta^{(1)}_m$ and $\eta^{(2)}_1,\ldots,\eta^{(2)}_n$ need not be identically distributed, and may be dependent and highly correlated across coordinates and data points; such effects can degrade standard embedding methods applied to each dataset separately or to the concatenated data $\{\mathcal{X},\mathcal{Y}\}$~\cite{landa2021doubly}.

We further assume that the ambient dimension $p$ and the smaller dataset size $m$ (recalling that $m\leq n$) are increasing with at least some fractional power of $n$.
\begin{assump} \label{assump: dimensions}
There exists global constants $\kappa,\gamma > 0$ such that $\min\{p,m\}\geq \kappa n^\gamma$.
\end{assump}
This assumption enables us to study the EOT plan $W$ under the model~\eqref{eq: shared manifold observation model} in the large-sample, high-dimensional regime, and establish concentration around a suitable population form.
Note that the size of the smaller dataset $m$ and the ambient dimension $p$ are allowed to grow very slowly with $n$, e.g., $m,p \sim n^{0.01}$, so they can practically be much smaller than $n$ for large $n$. Alternatively, $p$ can also grow much faster than $m$ and $n$.
The quantities $\kappa$, $\gamma$, and $C$ are considered as fixed global constants in our setup. All constants in our results may implicitly depend on these global constants, while other quantities can vary freely within our assumptions (unless stated otherwise). Our focus is on the high-dimensional regime where $m$, $n$, and $p$ are sufficiently large but otherwise arbitrary.

\subsection{Concentration of the EOT plan $W$ around a population form} \label{sec: concentration of EOT plan}
To state our main result, we require several definitions. First, we define $\mathcal{A}\in\mathbb{R}^{r\times r}$ as the positive-definite square root of $ \mathcal{A}_1 \mathcal{A}_2$ (or equivalently $ \mathcal{A}_2 \mathcal{A}_1$, since $ \mathcal{A}_1$ and $\mathcal{A}_2$ commute), namely
\begin{equation}
    \mathcal{A} = \left( \mathcal{A}_1 \mathcal{A}_2 \right)^{1/2}, \label{eq: A_mathcal def}
\end{equation}
where $ \mathcal{A}_1$ and $\mathcal{A}_2$ are the individual distortion matrices in the model~\eqref{eq: shared manifold observation model}.
Let $\mathcal{A}(\cdot):\mathcal{M}\to\mathbb{R}^r$ denote the linear map $x \mapsto \mathcal{A}x$, and set $\mathcal{N} = \mathcal{A}(\mathcal{M})$, which is a $d$-dimensional compact Riemannian manifold embedded in $\mathbb{R}^r$. Geometrically, $\mathcal{N}$ is a deformed version of $\mathcal{M}$, obtained by stretching and 
compressing along the eigen-directions of $\mathcal{A}$ in $\mathbb{R}^r$.
We equip $\mathcal{N}$ with the pushforward volume form $d\nu$ of $d\mu$ under $\mathcal{A}$, i.e.,
\begin{equation}
    \int_{\mathcal{N}} g(y)\, d\nu(y) = \int_{\mathcal{M}} g(\mathcal{A}x)\, d\mu(x),
\end{equation}
for all measurable $g:\mathcal{N}\rightarrow\mathbb{R}$, and define $\widetilde{\omega}(y) = \omega(\mathcal{A}^{-1} y)$ for $y\in\mathcal{N}$ as the corresponding sampling density on $\mathcal{N}$ with respect to $d\nu$. 
We then define the kernels 
\begin{equation}
    \mathcal{W}_\varepsilon(x,y) = \rho_{\varepsilon}(x) \mathcal{K}_{\varepsilon}(x,y) \rho_{\varepsilon}(y), \qquad\mathcal{K}_{\varepsilon}(x,y) 
    = \frac{1}{(\pi \varepsilon)^{d/2}} \operatorname{exp} \left\{\frac{-\Vert x - y \Vert_2^2}{\varepsilon} \right\}, \qquad x,y \in \mathcal{N} \label{eq: W integral def}
\end{equation}
where $\rho_{\varepsilon}(x):\mathcal{N}\to(0,\infty)$ is the function that solves the integral equation 
\begin{equation}
    1 = \int_\mathcal{N} \mathcal{W}_\varepsilon(x,y) \widetilde{\omega}(y) d\nu(y) = \rho_{\varepsilon}(x) \int_\mathcal{N} \mathcal{K}_{\varepsilon}(x,y) \rho_{\varepsilon}(y) \omega(\mathcal{A}^{-1} y) d\nu(y), \label{eq:integral scaling eq with density}
\end{equation}    
for all $x\in \mathcal{N}$. 
Hence, $\mathcal{W}_\varepsilon(x,y)$ is obtained by symmetrically scaling the Gaussian kernel $\mathcal{K}_\varepsilon(x,y)$ to be doubly stochastic with respect to the probability measure $\widetilde{\omega} \, d\nu$ on $\mathcal{N}$. The scaling function $\rho_{\varepsilon}(x)$ is guaranteed to exist and is a unique positive and continuous function on $\mathcal{N}$; see~\cite{borwein1994entropy} and~\cite{knopp1968note}. 
The doubly stochastic kernel $\mathcal{W}_\varepsilon (x,y)$ can be interpreted as the entropic self-transport plan of the measure $\widetilde{\omega} \, d\nu$, i.e., the continuous solution to the entropic optimal transport problem from $\widetilde{\omega} \, d\nu$ to itself~\cite{peyre2019computational}. Such doubly stochastic kernels have been studied in the context of manifold learning for a single dataset~\cite{marshall2019manifold,wormell2021spectral,landa2023robust,cheng2024bi}. 

Our main result below characterizes the behavior of the EOT plan under the model~\eqref{eq: shared manifold observation model}. 
\begin{thm} \label{thm: concentration of transport plan}
Let $W$ be the EOT plan~\eqref{eq: optimal transport optimization problem} computed from the datasets $\mathcal{X}$ and $\mathcal{Y}$ according to the model~\eqref{eq: shared manifold observation model}. Under Assumptions~\ref{assump: noise magnitude} and~\ref{assump: dimensions}, there exist $\tau_0,n_0(\varepsilon), C^{'}({\varepsilon})>0$, 
such that for all $n\geq n_0(\varepsilon)$, 
\begin{align}
    \left\vert \sqrt{mn} W_{i,j} - \mathcal{W}_{\varepsilon}(\mathcal{A} \overline{x}_i, \mathcal{A} \overline{y}_j) \right\vert \leq {\tau} C^{'}({\varepsilon}) \max \left\{ E\sqrt{\log p}, E^2 \sqrt{p\log p}, \sqrt{\frac{\log m}{m}}\right\}, \label{eq: W_{i,j} variance error}
\end{align}
for all $i=1,\ldots,m$ and $j=1,\ldots,n$ with probability at least $1-n^{-\tau}$, for all $\tau \geq \tau_0$.
\end{thm}

Theorem~\ref{thm: concentration of transport plan} shows that under the latent manifold model~\eqref{eq: shared manifold observation model}, for any fixed bandwidth $\varepsilon$, sufficiently large dataset sizes $m$ and $n$, and sufficiently high ambient dimension $p$, the transport-plan entries $W_{i,j}$ concentrate around \sloppy$(mn)^{-1/2}\mathcal{W}_{\varepsilon}(\mathcal{A}\overline{x}_i,\mathcal{A}\overline{y}_j)$, which we view as the population analog of $W_{i,j}$. In particular, we have a probabilistic bound on the deviation between $\sqrt{mn}\,W_{i,j}$ and $\mathcal{W}_{\varepsilon}(\mathcal{A}\overline{x}_i,\mathcal{A}\overline{y}_j)$, with explicit dependence on $m$, $p$, and the sub-Gaussian norm of the noise $E$. In the noiseless case $E=0$, this error converges almost surely to zero at rate $\sqrt{\log m / m}$, reflecting sample-to-population convergence governed by the smaller dataset (recall $m\le n$). In fact, when $E=0$, the growth condition on $p$ in Assumption~\ref{assump: dimensions} is unnecessary: the ambient dimension plays no role in the concentration of the EOT plan.

Under Assumptions~\ref{assump: noise magnitude} and~\ref{assump: dimensions}, all terms on the right-hand side of~\eqref{eq: W_{i,j} variance error} vanish as $n\to\infty$ for fixed $\varepsilon$. Consequently, as $n\to\infty$, we have almost surely
\begin{equation}
    \sqrt{mn}\, W_{i,j} \sim \mathcal{W}_{\varepsilon}(\mathcal{A} \overline{x}_i,\mathcal{A} \overline{y}_j)
    = \frac{1}{(\pi \varepsilon)^{d/2}} \rho_{\varepsilon} (\mathcal{A} \overline{x}_i ) \exp \left\{ - \frac{ \left\Vert \mathcal{A} \left( \overline{x}_i - \overline{y}_j \right) \right\Vert_2^2}{\varepsilon} \right\} \rho_{\varepsilon} (\mathcal{A} \overline{y}_j ),
\end{equation}
so $\sqrt{mn}\,W_{i,j}$ encodes affinities between $\overline{x}_i$ and $\overline{y}_j$ through the map $x\mapsto \mathcal{A}x$, i.e., between $\mathcal{A}\overline{x}_i$ and $\mathcal{A}\overline{y}_j$ on the effective manifold $\mathcal{N}$. These affinities take the form of a doubly stochastic Gaussian kernel using the Mahalanobis distance $(\overline{x}_i-\overline{y}_j)^{T}\mathcal{A}^{2}(\overline{x}_i-\overline{y}_j)$ rather than the squared Euclidean distance $\|\overline{x}_i-\overline{y}_j\|_2^2$. Thus, latent points that are close on $\mathcal{M}$ are still encoded as similar, but with distances distorted to reflect the geometry of $\mathcal{N}$ instead of $\mathcal{M}$. In the special case of pure global re-scaling, where $\mathcal{A}_1$ and $\mathcal{A}_2$ are (possibly different) multiples of the identity, we have $\mathcal{A}=aI$ for some $a>0$; the geometry is unchanged, and the affinities reduce to those based on Euclidean distance up to a scalar that can be absorbed into $\varepsilon$.

The convergence of $\sqrt{mn}\,W_{i,j}$ to $\mathcal{W}_{\varepsilon}(\mathcal{A} \overline{x}_i,\mathcal{A} \overline{y}_j)$ holds despite dataset-specific translations, SPD deformations, orthogonal nuisance structures, and noise in~\eqref{eq: shared manifold observation model}---even when the noise is heteroskedastic and grows unbounded with dimension (see text after Assumption~\ref{assump: noise magnitude}).
This robustness is non-trivial and is not shared by common cross-dataset affinity measures such as k-NN graphs or row-stochastic Gaussian kernels. Indeed, expanding $\|x_i-y_j\|_2^2$ under~\eqref{eq: shared manifold observation model} reveals substantial deviations from $\|\mathcal{A}(\overline{x}_i-\overline{y}_j)\|_2^2$. However, the EOT plan is robust to these effects by virtue of the scaling factors $\alpha$ and $\beta$ in~\eqref{eq: W and K def} and their behavior under the marginal constraints~\eqref{eq: B def}. The proof of Theorem~\ref{thm: concentration of transport plan} (Appendix~\ref{appendix: proof of theorem on concentration of transport plan}) constructs deterministic surrogates for $\alpha$ and $\beta$ using the scaling function $\rho_\varepsilon$ and other model quantities, and then performs a stability analysis showing that the scaling factors (more precisely, the dual potentials $\mathbf{f}_i=\varepsilon\log(\alpha_i)$ and $\mathbf{g}_j=\varepsilon\log(\beta_j)$) effectively correct the raw distances $\|x_i-y_j\|_2^2$ in~\eqref{eq: W and K def} to approximate the latent distances $\|\mathcal{A}(\overline{x}_i-\overline{y}_j)\|_2^2$. In this sense, the scaling factors absorb the adverse effects of dataset-specific deformations and corruptions.

\subsection{Population interpretation of the spectral embedding} \label{sec: population interpretation of embedding under latent manifold model}
Next, we interpret our low-dimensional embedding by linking it to the eigenfunctions of a certain population-level integral operator.
Let $\mathcal{L}^2_{\tilde{\omega}}$ be the Hilbert space of functions over $\mathcal{N}$ endowed with the inner product \sloppy$\langle g, h \rangle_{\tilde{\omega}} = \int_\mathcal{N} g(x) h(x) \tilde{\omega}(x) d\nu(x)$ for any $g,h:\mathcal{N} \rightarrow \mathbb{R}$ (for the definitions of $\tilde{w}$ and $d\nu$ see the beginning of Section~\ref{sec: concentration of EOT plan}). We define the integral operator $\mathscr{W}_\varepsilon: \mathcal{L}^2_{\tilde{\omega} } \rightarrow \mathcal{L}^2_{\tilde{\omega} }$ via
\begin{equation}
    \{\mathscr{W}_\varepsilon g\}(x) = \langle \mathcal{W}_\varepsilon (x,\cdot), g \rangle_{\tilde{\omega}} = \int_\mathcal{N} \mathcal{W}_\varepsilon(x,y) g(y) \tilde{\omega}(y) d\nu(y), \label{eq: W_tidle integral operator def}
\end{equation}
for all $x\in\mathcal{N}$ and any $g\in \mathcal{L}^2_{\tilde{\omega}}$. 
The integral operator $\mathscr{W}_\varepsilon$ is compact, self-adjoint, and positive definite over $\mathcal{L}^2_{\tilde{\omega}}$~\cite{marshall2019manifold}, admitting a sequence of positive eigenvalues $\{\nu_k^{(\varepsilon)}\}_{k=1}^\infty$, sorted in descending order, and corresponding eigenfunctions $\{\xi_k^{(\varepsilon)}\}_{k=1}^\infty$, which are orthonormal and complete in $\mathcal{L}^2_{\tilde{\omega}}$.
The first eigenpair of $\mathscr{W}_\varepsilon$ is trivial, namely $\nu_1 = 1$ and $\xi_1(x) = 1$ for all $x\in\mathcal{N}$. Moreover, $\mathscr{W}_\varepsilon$ is a transition probability operator with respect to the measure $\tilde{\omega} \, d\nu$, i.e., for any nonnegative function $g$ satisfying $\int_\mathcal{N} g(y) \tilde{\omega}(y) d\nu(y) = 1$, we have that $\mathscr{W}_\varepsilon g $ is nonnegative and satisfies $\int_\mathcal{N} \{\mathscr{W}_\varepsilon g\}(x) \tilde{\omega}(x) d\nu(x) = 1$. Hence, $\mathscr{W}_\varepsilon$ generates a random walk on the manifold $\mathcal{N}$, where $\mathscr{W}_\varepsilon^t g$ is the probability distribution of the random walker's location across $\mathcal{N}$ after $t$ steps with the initial location distribution $g$. 

We proceed to define the Hilbert space $\mathcal{H} = \mathcal{L}^2_{\tilde{\omega}} \times \mathcal{L}^2_{\tilde{\omega}}$, where each function $f\in \mathcal{H}$ can be written as $f = \begin{bmatrix} g \\ h
\end{bmatrix}$ for $g,h\in \mathcal{L}^2_{\tilde{\omega}}$, endowed with the inner product $\langle f_1, f_2 \rangle_\mathcal{H} = \langle g_1, g_2 \rangle_{\tilde{\omega}} + \langle h_1, h_2 \rangle_{\tilde{\omega}}$ for any $f_1 = \begin{bmatrix} g_1 \\ h_1 \end{bmatrix}$ and $f_2 = \begin{bmatrix} g_2 \\ h_2 \end{bmatrix}$. Additionally, we define the operator $\mathcal{P}_\varepsilon: \mathcal{H} \rightarrow \mathcal{H}$ such that for any $f = \begin{bmatrix} g \\ h
\end{bmatrix} \in \mathcal{H}$, we have
\begin{equation}
    \mathcal{P}_\varepsilon f = 
    \begin{bmatrix}
        0 & \mathscr{W}_\varepsilon\\
        \mathscr{W}_\varepsilon & 0
    \end{bmatrix} 
    \begin{bmatrix}
        g\\
        h
    \end{bmatrix} 
    = \begin{bmatrix}
        \mathscr{W}_\varepsilon h\\
        \mathscr{W}_\varepsilon g
    \end{bmatrix}. \label{eq: operator P_cal def}
\end{equation}
The operator $\mathcal{P}_\varepsilon$ is compact and self adjoint over $\mathcal{H}$, with eigenvalues $\{\nu_k^{(\varepsilon)}\}_{k=1}^\infty \cup \{-\nu_k^{(\varepsilon)}\}_{k=1}^\infty$ and corresponding eigenfunctions 
\begin{equation}
    \left\{ \frac{1}{\sqrt{2}}\begin{bmatrix}
    \xi_k^{(\varepsilon)} \\
    \xi_k^{(\varepsilon)}
    \end{bmatrix}
    \right\}_{k=1}^{\infty}
    \bigcup 
     \left\{ \frac{1}{\sqrt{2}} \begin{bmatrix}
    \xi_k^{(\varepsilon)} \\
    -\xi_k^{(\varepsilon)}
    \end{bmatrix}
    \right\}_{k=1}^{\infty}, \label{eq: eigenfunctions of P_cal}
\end{equation}
which are orthonormal and complete in $\mathcal{H}$, where $\{\nu_k^{(\varepsilon)}\}$ and $\{\xi_k^{(\varepsilon)}\}$ are the eigenvalues and eigenfunctions of $\mathscr{W}_\varepsilon$, respectively. Since $\mathscr{W}_\varepsilon$ is a transition probability operator over $\mathcal{N}$, then $\mathcal{P}_\varepsilon$ is a transition probability operator over $\mathcal{N}\times \mathcal{N}$, describing a random walk between two copies of $\mathcal{N}$. At each step of this random walk, the random walker can transition from a location in one copy of the manifold $\mathcal{N}$ to another location in the other copy, but not within the same copy of $\mathcal{N}$, akin to a random walk on a bipartite graph. 

For any $f=\begin{bmatrix} g\\ h \end{bmatrix} \in\mathcal{H}$, we define the evaluation of $f$ at the latent datasets $(\overline{\mathcal{X}},\overline{\mathcal{Y}})$ as
    \begin{equation}
        [f]_{(\overline{\mathcal{X}},\overline{\mathcal{Y}})} = 
        \begin{bmatrix}
            [g]_{\overline{\mathcal{X}}} \\
            [h]_{\overline{\mathcal{Y}}}
        \end{bmatrix}
        \qquad\qquad 
        [g]_{\overline{\mathcal{X}}} = 
        \begin{bmatrix}
            g(\mathcal{A}\overline{x}_1) \\
            \vdots \\
            g(\mathcal{A}\overline{x}_m)
        \end{bmatrix},
        \qquad\qquad 
        [h]_{\overline{\mathcal{Y}}} = 
        \begin{bmatrix}
            h(\mathcal{A} \overline{y}_1) \\
            \vdots \\
            h(\mathcal{A} \overline{y}_n)
        \end{bmatrix},
    \end{equation}
recalling that $\mathcal{A} x \in\mathcal{N}$ for any $x\in\mathcal{M}$. The following corollary of Theorem~\ref{thm: concentration of transport plan} establishes pointwise operator convergence of powers of the transition probability matrix $P$ from Section~\ref{sec: inter-data diffusion maps} (see~\eqref{eq: P def}) to the corresponding power of the integral operator $\mathcal{P}_{\varepsilon}$.
\begin{cor} \label{cor: pointwise convergence of P}
    Fix any integer $t > 0$ and $f = \begin{bmatrix} g\\ h \end{bmatrix} \in \mathcal{H}$, where $g,h$ are bounded functions on $\mathcal{N}$. Then, under Assumptions~\ref{assump: noise magnitude} and~\ref{assump: dimensions}, as $n\rightarrow \infty$, we have almost surely that
\begin{equation}
    \left\Vert P^t \left[ {f} \right]_{(\overline{\mathcal{X}},\overline{\mathcal{Y}})} - \left[ \mathcal{P}^t_{\varepsilon} f \right]_{(\overline{\mathcal{X}},\overline{\mathcal{Y}})} 
 \right\Vert_\infty  \longrightarrow 0,
\end{equation}
where the matrix $P\in\mathbb{R}^{(m+n)\times (m+n)}$ is from~\eqref{eq: P def}.
\end{cor}

The proof of Corollary~\ref{cor: pointwise convergence of P} is given in Appendix~\ref{appendix: proof of corollary on pointwise convergence of P}. 
Under the latent manifold model~\eqref{eq: shared manifold observation model}, the corollary implies that the random walk on the bipartite graph $\mathcal{G}$ generated by $P$ (Section~\ref{sec: inter-data diffusion maps}) converges, in a pointwise operator sense, to the continuum random walk on two copies of the effective manifold $\mathcal{N}$ generated by the transition operator $\mathcal{P}_{\varepsilon}$. 
Since our embedding admits an inter-data diffusion-distance interpretation in terms of $P^t$ (Proposition~\ref{prop: diffusion distance identities}), we view $\mathcal{P}_{\varepsilon}$ as the population analog of $P$ in our setting.

Recall that our embedding uses the leading nontrivial eigenvectors of $P$ with positive eigenvalues as coordinates (equivalently, eigenvectors of $\widetilde L$); see~\eqref{eq: embedding interpretation with eigenvectors of P}. 
We thus consider the population counterpart in terms of the leading nontrivial eigenfunctions of $\mathcal{P}_{\varepsilon}$ with positive eigenvalues (see~\eqref{eq: eigenfunctions of P_cal}), evaluated at the latent samples mapped to $\mathcal{N}$. 
Concretely, the $(k-1)$st coordinate of this population-level embedding is given (up to a scalar multiple) by
\begin{equation}
    \frac{1}{\sqrt{2}}
\begin{bmatrix}
            \xi_k^{(\varepsilon)}(\mathcal{A} \overline{x}_1) \\
            \vdots \\
            \xi_k^{(\varepsilon)}(\mathcal{A} \overline{x}_m) \\
            \xi_k^{(\varepsilon)}(\mathcal{A} \overline{y}_1) \\
            \vdots \\
            \xi_k^{(\varepsilon)}(\mathcal{A} \overline{y}_n)
        \end{bmatrix}, 
        \qquad k=2,\ldots,q+1\le m,
    \label{eq: sampled eigenfunctions of P_cal}
\end{equation}
where $\xi_k^{(\varepsilon)}$ denotes the $k$th eigenfunction of $\mathscr{W}_{\varepsilon}$.
Since $\mathscr{W}_{\varepsilon}$ has a continuous kernel on the compact manifold $\mathcal{N}$, its eigenfunctions $\xi_k^{(\varepsilon)}$ are continuous. 
Because $\mathcal{A}$ is a fixed bounded linear map, continuity implies that nearby latent points on $\mathcal{M}$ yield nearby embedding coordinates:
\begin{equation}
    \vert \xi_k^{(\varepsilon)}(\mathcal{A} \overline{x}) - \xi_k^{(\varepsilon)}(\mathcal{A} \overline{y}) \vert \rightarrow 0, 
    \qquad \forall \; \overline{x},\overline{y} \in \mathcal{M}: \; \Vert \overline{x} - \overline{y} \Vert_2 \rightarrow 0.  
\end{equation}

Thus, for sufficiently large $m$, $n$, $p$ in our setup, Corollary~\ref{cor: pointwise convergence of P} suggests that the embedding from Section~\ref{sec: approach and properties} aligns the two datasets by their intrinsic latent locations: if $\overline{x}_i$ and $\overline{y}_j$ are close on $\mathcal{M}$, then $\widetilde x_i$ and $\widetilde y_j$ are close in $\mathbb{R}^q$, despite dataset-specific translations, orthogonal nuisance structure, noise, and distinct deformations.

\subsection{Behavior under small entropic regularization} \label{sec: behavior under small bandwidth}
Lastly, we provide a more refined geometric interpretation of the EOT plan and the embedding for small entropic regularization $\varepsilon$ (while the sample size and dimension are sufficiently large). To this end, we combine our previous results in Sections~\ref{sec: concentration of EOT plan} and~\ref{sec: population interpretation of embedding under latent manifold model} with existing results on the doubly stochastic Gaussian kernel and the scaling function $\rho_\varepsilon$ (see~\eqref{eq: W integral def}) for small bandwidth~\cite{landa2023robust,cheng2024bi,wormell2021spectral}.

Let the columns of $\mathcal{T}_{\mathcal{M}}(x)\in \mathbb{R}^{r\times d}$ be an orthonormal basis for the tangent space of $\mathcal{M}$ at ${x}\in\mathcal{M}$ and denote $\widetilde{\mathcal{T}}_{\mathcal{M}}(y) = {\mathcal{T}}_{\mathcal{M}}(\mathcal{A}^{-1} y)$ for $y\in\mathcal{N}$. We define for all $x,y\in\mathcal{N}$, 
\begin{equation}
    \hat{\omega}(y) = \frac{\tilde{\omega}(y)}{\sqrt{\det\left\{ \widetilde{\mathcal{T}}_{\mathcal{M}}^T(y) \, \mathcal{A}^2 \, \widetilde{\mathcal{T}}_{\mathcal{M}}(y) \right\}}}, \qquad \qquad 
    \hat{\mathcal{W}}_\varepsilon(x,y) = \frac{\mathcal{K}_{\varepsilon}(x,y)}{\sqrt{\hat{\omega}(x) \, \hat{\omega}(y)}}, \label{eq: W_tilde_hat kernel def}
\end{equation}
where $\det\{\cdot\}$ denotes the determinant of a matrix, and $\tilde{\omega}(y) = \omega(\mathcal{A}^{-1} y)$ for $y\in\mathcal{N}$ is the density function of $\mathcal{N}$ with respect to the pushforward volume form $d\nu$ (see the beginning of Section~\ref{sec: concentration of EOT plan}). Note that the denominator of $\hat{\omega}$ in~\eqref{eq: W_tilde_hat kernel def} is precisely the Jacobian of the linear map $ x \mapsto \mathcal{A}x$ for $x\in\mathcal{M}$, evaluated at $y = \mathcal{A} x \in \mathcal{N}$. Hence, $\hat{\omega}$ is the density on $\mathcal{N}$ with respect to the natural volume form induced by the Euclidean metric in $\mathbb{R}^{r}$, and $\hat{\mathcal{W}}_\varepsilon(x,y)$ in~\eqref{eq: W_tilde_hat kernel def} is the Gaussian kernel normalized symmetrically by the square root of the sampling density, which is commonly used for spectral clustering and manifold learning; see, e.g.,~\cite{von2007tutorial,hein2007graph,hoffmann2022spectral,trillos2021geometric}.

The following corollary of Theorem~\ref{thm: concentration of transport plan} shows that the EOT plan $W$ approximates the kernel $\hat{\mathcal{W}}_\varepsilon(x,y)$ from~\eqref{eq: W_tilde_hat kernel def} for small bandwidth parameters $\varepsilon$. The proof can be found in Appendix~\ref{appendix: proof of corollary on relative error of W for small bandwidth} and relies on an asymptotic expansion of the function $\rho_\varepsilon$ for small $\varepsilon$~\cite{landa2023robust}. 
\begin{cor} \label{cor: relative error of W for small bandwidth}
    Suppose Assumptions~\ref{assump: noise magnitude} and~\ref{assump: dimensions} hold, $\mathcal{M}$ is smooth without boundary, $\omega \in \mathcal{C}^6(\mathcal{M})$, and there exists a global constant $c>0$ such that $\omega(x) \leq c$ for all $x\in \mathcal{M}$ and $\Vert \mathcal{A}^{-1} \Vert_2 \leq c$. Then, there exist $\tau_0,\varepsilon_0, c^{'},n_0(\varepsilon),C^{'}({\varepsilon})>0$, 
such that 
\begin{equation}
    \frac{1}{mn} \sum_{i=1}^m \sum_{j=1}^n \left\vert \frac{ \sqrt{mn} W_{i,j} - \hat{\mathcal{W}}_{\varepsilon} (\mathcal{A} \overline{x}_i,\mathcal{A} \overline{y}_j) }{\hat{\mathcal{W}}_{\varepsilon} (\mathcal{A} \overline{x}_i,\mathcal{A} \overline{y}_j)} \right\vert \leq  
    c^{'} \varepsilon 
    +  {\tau} C^{'}({\varepsilon}) \max \left\{ E\sqrt{\log p}, E^2 \sqrt{p\log p}, \sqrt{\frac{\log m}{m}}\right\}\label{eq: W_hat bias and variance errors}
\end{equation}  
for all $n\geq n_0(\varepsilon)$ and $\varepsilon \leq \varepsilon_0$, with probability at least $1-n^{-\tau}$, for all $\tau\geq \tau_0$. Here, the constants $c^{'}$ and $\varepsilon_0$ may depend also on $\mathcal{M}$, $\omega$, and $\mathcal{A}$ in addition to $c$ and the global constants in Assumptions~\ref{assump: noise magnitude} and~\ref{assump: dimensions}.
\end{cor}
Corollary~\ref{cor: relative error of W for small bandwidth} bounds the relative error between $\sqrt{mn}W_{i,j}$ and $\hat{\mathcal{W}}_{\varepsilon} (\mathcal{A} \overline{x}_i,\mathcal{A} \overline{y}_j)$. This bound consists of an $\mathcal{O}(\varepsilon)$ term and the same probabilistic error bound from Theorem~\ref{thm: concentration of transport plan}, where the latter stems from the noise and the sample-to-population convergence. The error tends to zero almost surely if $m,n,p \rightarrow \infty$ sufficiently quickly while $\varepsilon\rightarrow 0$ sufficiently slowly. We note that the relative (normalized) error used in Corollary~\ref{cor: relative error of W for small bandwidth} is more informative than the absolute (un-normalized) error due to the rapid exponential decay of $\hat{\mathcal{W}}_\varepsilon(x,y)$ in $\varepsilon$ as $\varepsilon$ decreases.

Let us denote by $\hat{\mathscr{W}}_\varepsilon$ the integral operator defined analogously to $\mathscr{W}_\varepsilon$ from~\eqref{eq: W_tidle integral operator def} by replacing the kernel $\mathcal{W}_\varepsilon(x,y)$ with $\hat{\mathcal{W}}_\varepsilon(x,y)$ from~\eqref{eq: W_tilde_hat kernel def}. We then consider the Laplacian-type operators $\hat{\mathscr{L}}_\varepsilon = 4(\mathcal{I} - \hat{\mathscr{W}}_\varepsilon)/\varepsilon$ and ${\mathscr{L}}_\varepsilon = 4(\mathcal{I} - {\mathscr{W}}_\varepsilon)/\varepsilon$, where $\mathcal{I}$ is the identity operator. The operator $\hat{\mathscr{L}}_\varepsilon$ is the population analog of the popular symmetric normalized graph Laplacian (on $\mathcal{N}$); see, e.g.,~\cite{von2007tutorial,hein2007graph,hoffmann2022spectral,trillos2021geometric} and references therein. For sufficiently small $\varepsilon$, both operators $\hat{\mathscr{L}}_\varepsilon$ and ${\mathscr{L}}_\varepsilon$ approximate the same differential operator $\Delta_{(\mathcal{N},\hat{\omega})}$~\cite{wormell2021spectral,landa2023robust,cheng2024bi,coifman2006diffusionMaps,hein2007graph,garcia2020error}---the weighted manifold Laplacian on $\mathcal{N}$ with the density $\hat{\omega}$ from~\eqref{eq: W_tilde_hat kernel def}, defined by
\begin{equation}
    \{\Delta_{(\mathcal{N},\hat{\omega})} g\}(x) = \frac{\Delta_{\mathcal{N}}\{ g \sqrt{\hat{\omega}} \} (x)}{\sqrt{\hat{\omega}(x)}} - \frac{\Delta_{\mathcal{N}}\{\sqrt{\hat{\omega}} \} (x)}{\sqrt{\hat{\omega}(x)}} g(x), \label{eq: differential operator Delta omega}
\end{equation}
for all $x\in\mathcal{N}$ and $g\in \mathcal{C}^2(\mathcal{N})$, where $\Delta_{\mathcal{N}}$ is the negative Laplace-Beltrami operator on $\mathcal{N}$~\cite{grigor2006heat}.
Specifically, under suitable regularity conditions and a sufficiently smooth function $g(x)$ on $\mathcal{N}$, we have the pointwise approximation $(4/\varepsilon)\{\hat{\mathscr{L}}_\varepsilon g\}(x) =  \{\Delta_{(\mathcal{N},\hat{\omega})} g\}(x) + \mathcal{O}(\varepsilon)$~\cite{coifman2006diffusionMaps,hein2007graph} and similarly $(4/\varepsilon)\{{\mathscr{L}}_\varepsilon g\}(x) =  \{\Delta_{(\mathcal{N},\hat{\omega})} g\}(x) + \mathcal{O}(\varepsilon)$~\cite{landa2023robust,cheng2024bi}. Moreover, spectral convergence (i.e., convergence of eigenvalues and eigenvectors) of $\hat{\mathscr{L}}_\varepsilon$ to $\Delta_{(\mathcal{N},\hat{\omega})}$ as $\varepsilon\rightarrow 0$ was established in~\cite{garcia2020error}, while the spectral convergence of ${\mathscr{L}}_\varepsilon$ was established in~\cite{wormell2021spectral} for hyper-torus manifolds.

Our analysis suggests that under the model~\eqref{eq: shared manifold observation model}, for sufficiently large sample sizes $m$ and $n$, high dimension $p$, and sufficiently small bandwidth parameter $\varepsilon$, our proposed method approximates an embedding using the leading eigenfunctions of the operator $\Delta_{(\mathcal{N},\hat{\omega})}$ evaluated at the combined latent dataset (mapped to $\mathcal{N}$) $\{\mathcal{A} \overline{x}_1,\ldots,\mathcal{A} \overline{x}_m,\mathcal{A} \overline{y}_1,\ldots,\mathcal{A} \overline{y}_n\}$.
This operator has been extensively studied in the literature~\cite{grigor2006heat},
and the benefits of utilizing its eigenfunctions for clustering and manifold learning are well-established; see, e.g.,~\cite{hoffmann2022spectral,trillos2021geometric}.

\section{Simulation Studies} \label{simu.sec}

\paragraph{Noisy manifold alignment} We first focus on the alignment and joint embedding of a pair of high-dimensional datasets, where the goal is to preserve and align the shared low-dimensional manifold structure. We generate a pair of datasets  $\{x_1,...,x_m\}$ and $\{y_1,...,y_n\}$ according to the latent manifold model (\ref{eq: shared manifold observation model}). In the first simulation setup,  we define 
$ \label{UV}
\mathcal{U}=\begin{bmatrix} {\bf I}_r\\ {\bf 0}\end{bmatrix}\in\mathbb{R}^{p\times r},\quad \mathcal{V}_1 =0, \quad \mathcal{V}_2=\begin{bmatrix}{\bf e}_{r+1}&{\bf e}_{r+2} & ... & {\bf e}_{p} \end{bmatrix}\in \mathbb{R}^{p\times (p-r)},
$
where $\{{\bf e}_{i}\}$ is the ordinary Euclidean basis of $\mathbb{R}^p$. 
Setting $r=3$, we sample the latent variables $\{\bar x_1, ..., \bar x_m, \bar y_1, ..., \bar y_n\}\in \mathbb{R}^{r}$ uniformly from a torus $\mathbb{T}$ in $\mathbb{R}^3$, so that
$
\{\bar x_1, ..., \bar x_m, \bar y_1, ..., \bar y_n\}\subset\mathbb{T}\equiv \{(2+0.8\cos u) \cos v, (2+0.8\cos u)\sin v, 0.8\sin u): u,v\in[0,2\pi)\}.$
We generate the  datasets $\mathcal{X}$ and $\mathcal{Y}$ so that there are scale differences between them with various levels of mean shifts. Specifically, we set $z_i^{(1)}=z_i^{(2)}=0$, $\mathcal{A}_1=0.2\theta A$, $\mathcal{A}_2=\theta A^{-1}$ for $\theta=166$ and $A=\textup{diag}(0.2,5,0.2)$, and set 
$\nu_1=\tau\cdot \theta \text{\bf e}_1\in \mathbb{R}^p$ for different values of $\tau\in[1,20]$, and $\nu_2=0$. Further, we generate the noises $\eta_i^{(1)}$ and $\eta_i^{(2)}$  independently from $N(0,\sigma^2{\bf I}_p)$ with $\sigma=0.002\theta$. In the second setting, we fix $\tau=1$ and generate data-specific nuisance structures by setting $z_i^{(1)}=0$ for all $i$ and by drawing each component of $z_i^{(2)}\in \mathbb{R}^{p-r}$ independently from a uniform distribution on $[\theta/5,\theta/4]$. We set $\mathcal{A}_1=\textup{diag}(0.2,\gamma,0.2)$ and $\mathcal{A}_2=\mathcal{A}_1^{-1}$ for various levels of $\gamma\in[1,20]$. We generate $\eta_i^{(1)}=[\eta_{i1}^{(1)},...,\eta_{ip}^{(1)}]^T$ such that $\eta_{ik}^{(1)}\sim N(0,\sigma^2)$ and
heteroskedastic noise  $\eta_{j}^{(2)}=[\eta_{j1}^{(2)},...,\eta_{jp}^{(2)}]^T$ such that
\begin{equation} \label{eta2}
\eta_{jk}^{(2)} \sim    \begin{cases}
      N(0,10\sigma^2) & \text{if $2\le k\le r$ and $1\le j\le \lfloor \frac{n}{3} \rfloor$}\\
      N(0,5\sigma^2) & \text{if $1\le k\le r$ and $\frac{n}{3}< j\le \lfloor \frac{2n}{3} \rfloor$}\\
      N(0,\sigma^2) & \text{otherwise}
    \end{cases}.
\end{equation}
The other model parameters are the same as the first setting. {In the above simulation setup, the key parameters are $\tau$ and $\gamma$, with the former indicating the magnitude of mean shift, and the latter indicating the level of scaling discrepancy between the two datasets. We will evaluate the performance against different values of these parameters}. Throughout our simulations, we set $n=m=600$ and $p=1000$. We set the embedding dimension $q=3$.

We compare the following methods in terms of their embedding  and alignment quality: 
\begin{itemize}
\item the proposed EOT eigenmaps with $t=0$ (``EOT-0") and $t=1$ (``EOT-1"); 
\item the Seurat integration method \cite{stuart2019comprehensive} (``seurat"), which computes a joint embedding by performing SVD on the product of the two data matrices (Appendix~\ref{supp.simu.sec}); 
\item the Roseland \cite{shen2022scalability} (``rl") and LBDM (``lbdm'')~\cite{pham2018large} algorithms, which obtain a joint embedding via a landmark-based bipartite diffusion process between the two datasets. These methods are algorithmically similar to our approach but use different normalizations of the Gaussian kernel motivated by diffusion maps~\cite{coifman2006diffusionMaps} for a single dataset; see a more detailed discussion at the end of Section~\ref{sec: related work};
\item PCA applied to each dataset separately, forming the $q$-dimensional embedding from the first $q$ principal components, i.e., the first $q$ eigenvectors of the sample covariance matrix of each dataset (``pca"); 
\item kernel-based PCA applied to each dataset separately, forming the $q$-dimensional embedding from the eigenvectors of the kernel matrix described in~\cite{ding2023learning}; 
\item PCA applied to the concatenated dataset $\{x_1,...,x_{m}, y_1,...,y_{n}\}$, forming the embedding from the first $q$ principal components (``j-pca"); 
\item kernel-based PCA applied to the concatenated dataset, forming the embedding from the first $q$ eigenvectors of the kernel matrix described in~\cite{ding2023learning} (``j-kpca").
\end{itemize}
For ``EOT-0", ``EOT-1", ``lbdm" and ``rl", following the recommendation of  \cite{ding2023learning}, we set the bandwidth parameter $\varepsilon$ for the corresponding kernel matrices as the median of all pairwise distances $\{\|x_i-y_j\|^2\}_{i\in[m], j\in[n]}$; see Appendix~\ref{supp.simu.sec} for empirical justification. To evaluate the performance of each method, we compare the embedded data points $\widetilde T=\{\widetilde x_1, ..., \widetilde x_m, \widetilde y_1, ..., \widetilde y_n\}$ to their corresponding latent variables $\bar{T}=\{\bar x_1, ..., \bar x_m, \bar y_1, ..., \bar y_n\}$ in terms of their local neighborhood structures. Specifically, for each $\iota\in \bar T$ and corresponding $\widetilde{\iota} \in \widetilde T$, we denote by $S_\iota$ the set of 50-nearest neighbors of $\iota$ in  $\bar T$, and denote by $W_\iota$ the set of the 50-nearest neighbors of $\widetilde \iota$ in $\widetilde T$. We then compute 
\begin{equation}  \label{jaccard}
\text{concordance} = \frac{1}{m+n}\sum_{\iota\in \bar T}\text{[Jaccard Index]}_\iota,\qquad \text{[Jaccard Index]}_\iota\equiv\frac{|S_\iota\cap W_\iota|}{|S_\iota\cup W_\iota|},
\end{equation}
where the denominator $(n+m)$ reflects the total number of data points. A higher concordance score suggests a better alignment of the two datasets in terms of the shared torus structure and a better recovery of the local neighborhood structure of the latent torus samples. 
For each setting, the simulation is repeated 500 times to calculate the averaged concordance.

\begin{figure}[bt]
	\centering
\includegraphics[angle=0,width=15cm]{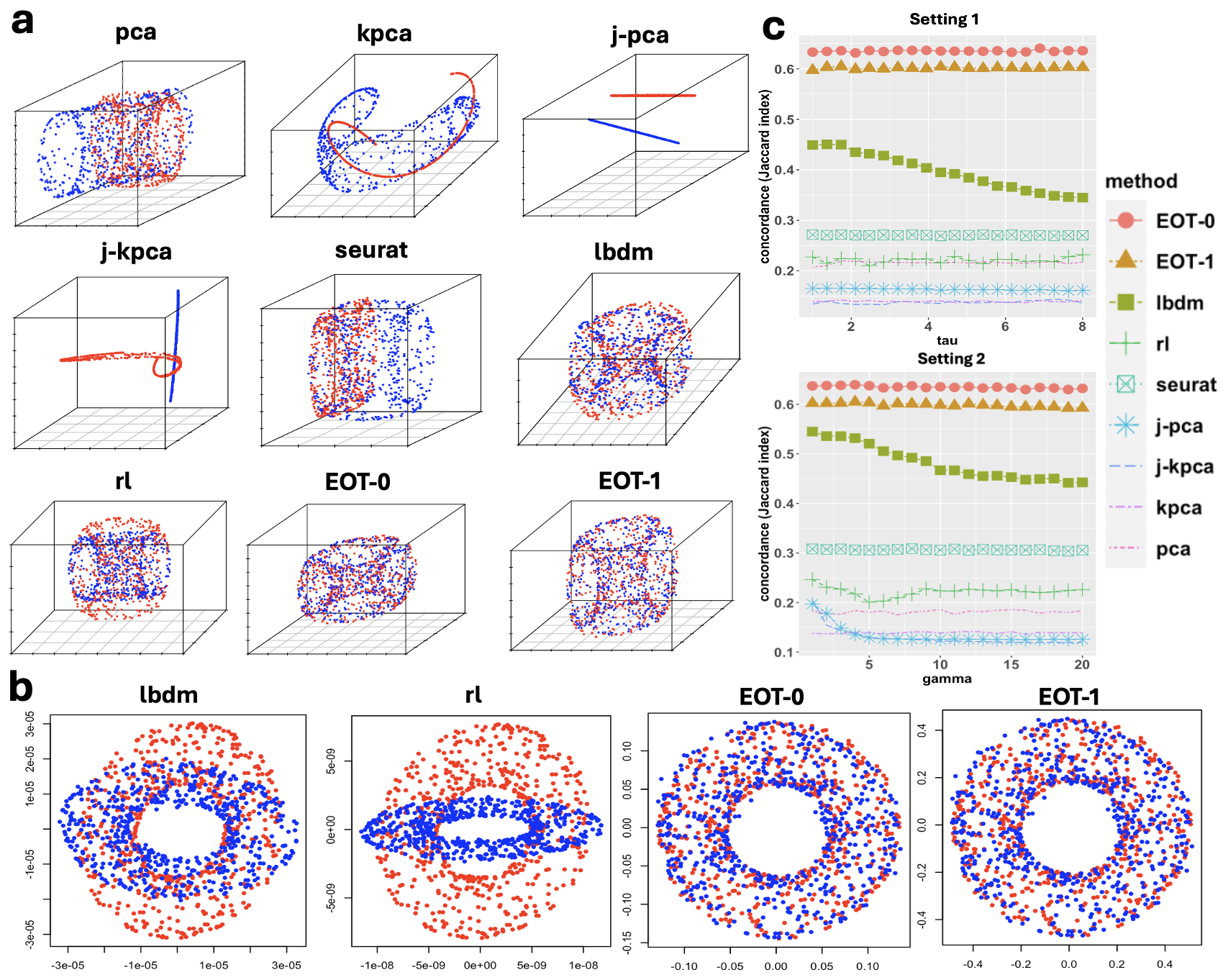}
	\caption{\footnotesize Comparison of nine integration methods based on simulations. (a) Visualization of the first three coordinates of each integrated low-dimensional embedding under Setting 2 of noisy manifold alignment experiments with $\gamma=20$, where the data points are colored according to datasets; (b) Closer comparison of  four methods in (a)  based on the first two coordinates of their integrated low-dimensional embeddings; (c) Numerical evaluation of different methods in terms of noisy manifold alignment across the two simulation settings.} 
	\label{simu.fig}
\end{figure}

We find that our proposed methods (``EOT-0" and ``EOT-1") in general have superior performance across all the settings over the alternative methods, with ``EOT-0" slightly outperforming ``EOT-1" (Figure~\ref{simu.fig}a-c). Similar observations can be made under an alternative metric (i.e., Davies-Bouldin index, see Appendix~\ref{supp.simu.sec}) focusing on the performance in alignment (Figure~\ref{supp.fig}, left and middle panels).  As the magnitude of the mean shifts ($\tau$) or the coordinate distortions ($\gamma$) increases, the existing methods demonstrate poor to mediocre performance, whereas the proposed methods are significantly better and more stable (Figure~\ref{simu.fig}c). From the visualization of the first three coordinates of each integrated low-dimensional embedding (Figure~\ref{simu.fig}a) under Setting 2 with $\gamma=20$, we observe that most existing methods fail to capture and align the latent torus structures. Among the best four joint embedding methods indicated in Figure~\ref{simu.fig}a, ``lbdm" and ``rl" achieve moderate performance in capturing the shared torus structure that is comparable with ``EOT-0" and ``EOT-1", but a closer examination of the first two coordinates shows discernible misalignment of the two datasets by ``lbdm" and ``rl" (Figure~\ref{simu.fig}b).

\paragraph{Joint clustering} Next, we consider the task of joint clustering of the samples in the two datasets, containing shared cluster patterns but also data-specific nuisance structures. %
Based on (\ref{eq: shared manifold observation model}), we generate $\bar x_i, \bar y_i\in\mathbb{R}^r$, $r=6$, from a Gaussian mixture model with $6$ classes of equal proportions, where each Gaussian component is from $N(\mu_j=5 \mathbf{e}_j,{\bf I}_r)$, where $\{\mathbf{e}_j\}_{1\le j\le r}$ are the standard Euclidean basis in $\mathbb{R}^r$. The noises $\eta^{(1)}_{ik}$ are generated from $N(0,1)$, whereas $\eta_{jk}^{(2)}$ are generated from (\ref{eta2}) with $\sigma^2=1$. We set $\nu_1=15{\bf e}_1+15{\bf e}_2$, $\nu_2=0$, and $\mathcal{A}_1=\mathcal{A}_2=\theta I_r$ for various values of $\theta\in[1,3]$. We set $z_i^{(1)}=0$, draw each component of $z_i^{(2)}\in\mathbb{R}^{p-r}$ independently from a uniform distribution on $[\theta/2,\theta]$, and set $\mathcal{U}$, $\mathcal{V}_1$, $\mathcal{V}_2$ as in (\ref{UV}). 
In this way, both datasets contain a latent low-dimensional Gaussian mixture cluster structure with the common cluster centers, whereas discrepancies exist between the two datasets in terms of a mean shift, a nuisance structure, and heteroskedastic noises. {The key parameter in this simulation setup $\theta$ indicates the overall signal-to-noise ratio in the two datasets.} We set $n=m=600$ and $p=1000$, and set the embedding dimension $q=6$.
We compare the performance of nine different joint clustering algorithms, each corresponding to one of the joint embedding algorithms evaluated above, with an additional $k$-means algorithm ($k=6$) applied to the integrated $r$-dimensional embeddings for both datasets. For each $\theta$, we evaluate the performance of each method using the Rand index \cite{rand1971objective} between the estimated and true cluster memberships. %
We repeat each setting 500 times to calculate the averaged Rand index.

\begin{figure}[bt]
	\centering
\includegraphics[angle=0,width=8cm]{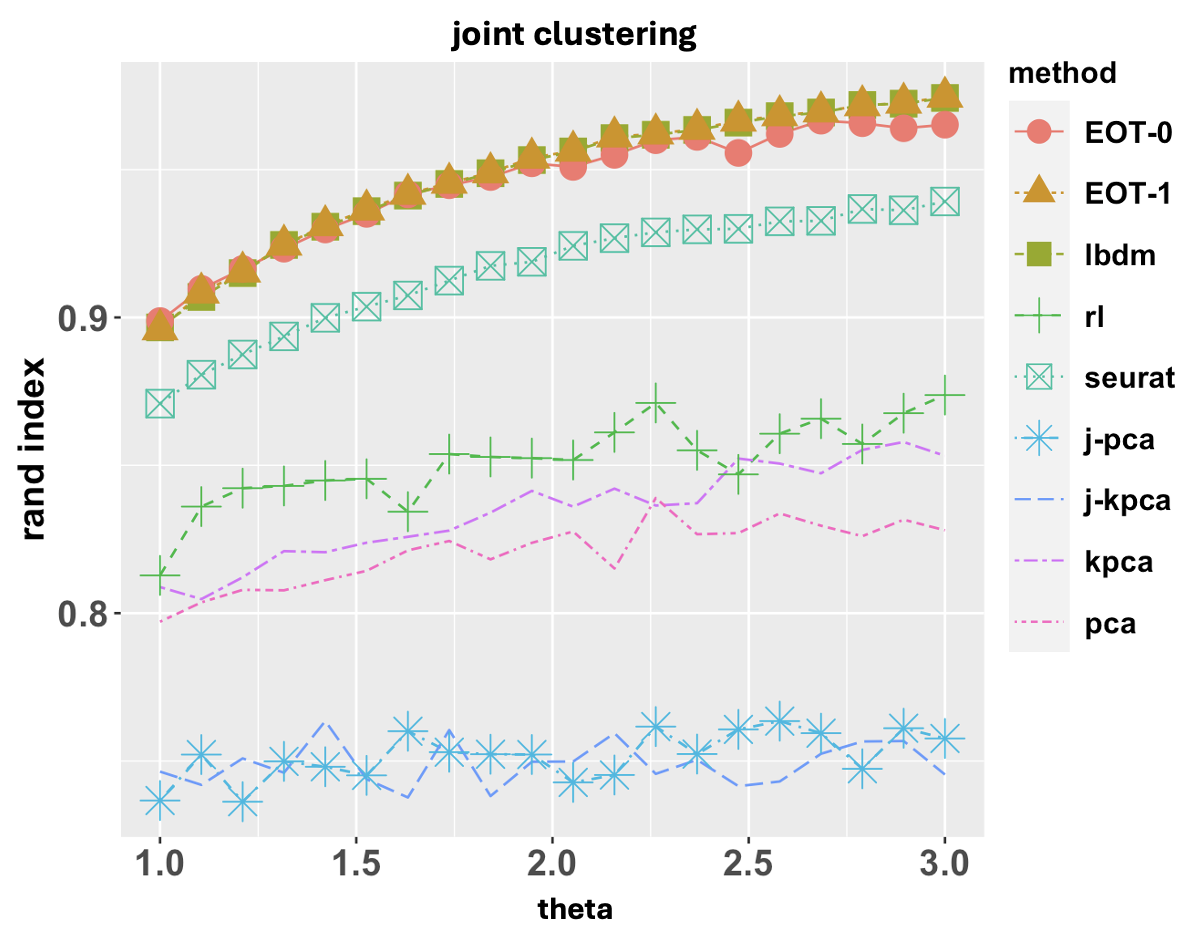}
	\caption{\footnotesize Comparison of nine integration methods in terms of joint clustering performance.} 
	\label{simu2.fig}
\end{figure} 

Figure~\ref{simu2.fig} illustrates the behavior of the Rand index as a function of $\theta$ in our experiment, which controls the magnitudes of the latent variables and the nuisance structure. Evidently, as $\theta$ increases, all methods except for ``j-pca" and ``k-jpca" show improved clustering accuracy. Importantly, ``EOT-0", ``EOT-1", and ``lbdm" exhibit superior performance in identifying the shared cluster patterns across the two datasets. In addition, we find that ``EOT-0" and ``EOT-1" perform significantly better than ``lbdm" and the other methods in aligning the latent cluster patterns across datasets (Figure~\ref{supp.fig}, rightmost panel).

\begin{figure}[t!]
	\centering
\includegraphics[angle=0,width=15cm]{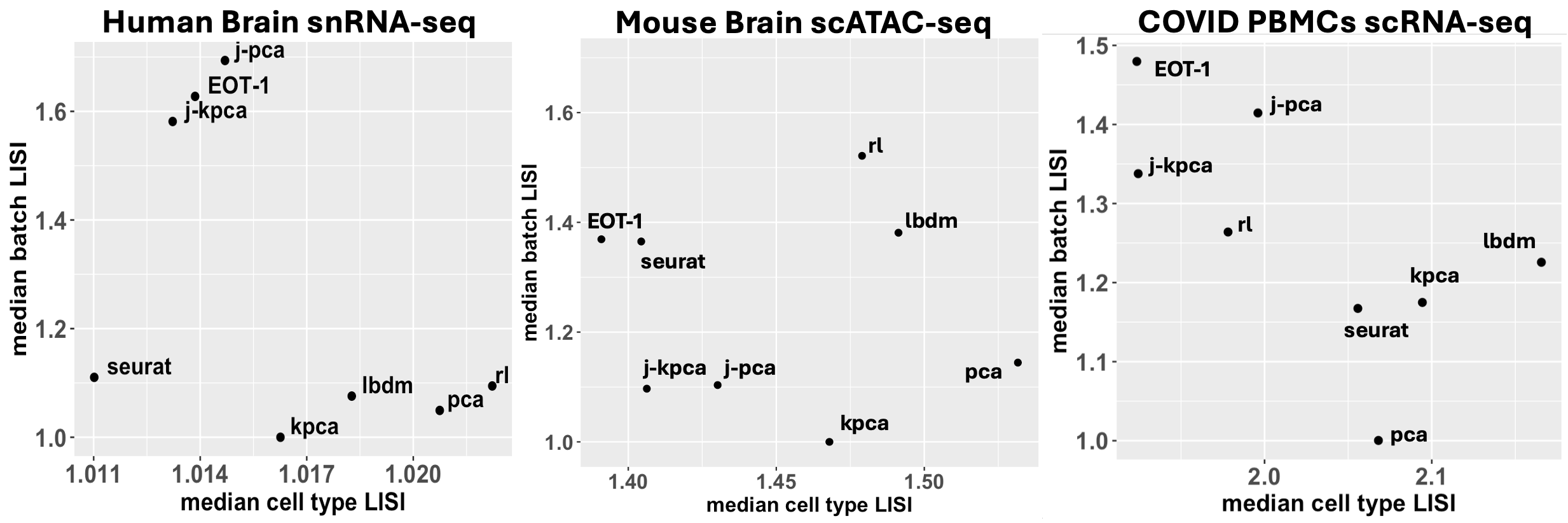}%
	\caption{\footnotesize Integrative analyses of single-cell omics data. A comparison of eight methods across three single-cell omics data integration tasks, evaluated using two metrics, where methods with better performance are expected to appear in the top left corner.} 
	\label{real.fig}
\end{figure}

\section{Applications to Single-Cell Data Integration} \label{sec: applications to single-cell data integration}

We evaluate our methods on three integration tasks. In each task we integrate a pair of single-cell omics datasets from different studies in order to identify the common underlying cell types. {The first pair of datasets concerns the human brain cells from two biological samples under the same clinical conditions \cite{ament2024single}. The first  sample contains $n_1=3126$ cells, whereas the second sample contains $n_2 = 6034$ cells. In this example, we expect that the differences between the datasets are primarily due to experimental artifacts, or batch effects. We are interested in obtaining a joint embedding of all cells from both biological samples into the same low-dimensional space, so as to align the corresponding cell types shared between the two datasets. Although the latent manifold model analyzed in Section~\ref{sec: analysis in the latent manifold model} may not hold exactly, we expect that key assumptions—such as the orthogonality between the shared latent manifold  (cell type variations) subspace and the subspace of nuisance structures (batch effects)—are approximately satisfied in these single-cell datasets \cite{haghverdi2018batch}.} We preprocess and normalize each dataset using the standard pipeline \cite{ding2023learning,ma2024principled}
described in Appendix~\ref{supp.simu.sec} and select the $p = 1000$ most variable genes for subsequent analysis. We apply our proposed method along with the seven alternative methods evaluated previously in our simulations. For each method, for various choices of $q$ (from 2 to 20), we obtain a $q$-dimensional joint embedding of both datasets and then evaluate the embedding and alignment quality based on the local inverse Simpson index (LISI), a popular metric for single-cell data integration developed by \cite{korsunsky2019fast}. Our first evaluation score is the LISI with respect to the original cell type annotations \cite{kang2018multiplexed,stuart2019comprehensive}. The LISI, calculated based on the neighborhood structure of the low-dimensional embeddings, measures the effective number of different categories in each cell's local neighborhood ($k=30$ neighbors). In this case, a smaller value (closer to 1) indicates more significant separation of data points corresponding to distinct cell types. In our analysis, for each method and each $q$, we compute the median cell type LISI across all the cells. The second evaluation score is LISI with respect to batch labels, and a larger value (closer to 2) indicates a better mixing of the two batches. For each method and each $q$, we compute the median batch LISI across all cells in the embedding space. 

Our second analysis concerns integrating a pair of single-cell ATAC-seq (scATAC-seq) datasets ($n_1=3618$ and $n_2=3715$) of mouse brain cells from different studies \cite{luecken2022benchmarking}. ATAC-seq is a biotechnology that quantifies genome-wide chromatin accessibility, which contains important information about epigenome dynamics and gene regulations. In our analysis, each dataset contains a matrix of ATAC-seq gene activity scores, characterizing gene-specific chromatin accessibility for individual cells. The third analysis concerns a pair of scRNA-seq datasets ($n_1=2839$ and $n_2=1221$) of human PBMCs from a COVID patient sampled from different time points after hospitalization \cite{zhu2020single}. For each pair of datasets, we perform the same analysis as before to obtain joint embeddings of all cells. For the proposed method, we choose $t=1$ {(see Figure~\ref{real.t.fig} for the results under alternative values of $t$).}
Figure~\ref{real.fig} and Figure~\ref{box.fig2} contain the evaluation results for the integrative analyses of all three pairs of datasets. For each analysis and each method, we present the median of the two metrics across different embedding dimensions $q$ in a scatter plot. %
Our results indicate that our proposed method exhibits overall the best performance in terms of both metrics, staying at the left upper corner of each scatter plot, indicating its better performance in both preserving the biological information (separating distinct cell types), and removing the batch effects (mixing the same cell types across batches). In contrast, the alternative methods have varying performance across the three analytical tasks. {Moreover, Figure~\ref{box.fig2} contains boxplots of the performance metrics  across varying embedding dimensions $q$. These boxplots illustrate the distribution and concentration of the metrics for each method, revealing smaller within-method variation compared to between-method variation.} This analysis also suggests that the proposed approach is robust to the choice of the embedding dimension $q$.

\section{Discussion} \label{appendix: discussion}
Our results suggest several potential future research directions. Firstly, it is desirable to extend our approach beyond the case of two datasets, namely, when the goal is to align and jointly embed an arbitrary number of datasets simultaneously. One natural approach is to consider all possible transport plans between pairs of datasets. Yet, it is not immediately obvious how to obtain a joint embedding from all pairwise transport plans and what are the interpretations and analytical properties of such an embedding. Another direction of interest is to allow for variable bandwidth kernels~\cite{berry2016variable} beyond a single bandwidth parameter $\varepsilon$. Several variants of optimal transport were recently proposed to account for this~\cite{van2024snekhorn,matsumoto2022beyond}. Such extensions can better adapt to local variations in the sampling density and further automate the embedding procedure. However, these approaches require substantial analytical and numerical investigation from the perspective of alignment and joint embedding under deformations and corruptions. Finally, it is desirable to expand the theoretical results of Section~\ref{sec: analysis in the latent manifold model} by considering more general deformation models and establishing more refined analytical results, e.g., pointwise error rates that depend explicitly on $\varepsilon$ and spectral convergence guarantees.
We leave such extensions for future work.

\section{Acknowledgements}
BL and YK acknowledge funding support from NIH grants R01GM131642, UM1PA051410, R33DA047037, U54AG076043, U54AG079759, U01DA053628, P50CA121974, and R01GM135928. RM would like to thank Xiucai Ding, Jason Buenrostro, Rafael Irizarry, and Chang Lu for their helpful discussions.

\bibliographystyle{plain}
\bibliography{mybib}

\begin{appendices}
    \section{Proof of Proposition~\ref{prop: embedding formula}}\label{appendix: proof of proposition on embedding formula for two datasets}
    Under the constraints~\eqref{eq: constraints for embedding}, the objective function~\eqref{eq: cost function for embedding} can be simplified as
\begin{align}
    J(\mathcal{X}^{'},\mathcal{Y}^{'}) 
    &= \sum_{i} \Vert {x}_i^{'} \Vert_2^2 \sum_j W_{i,j} + \sum_j \Vert {y}_j^{'} \Vert_2^2 \sum_i W_{i,j} - 2 \sum_{i,j} \left( ({x}_i^{'})^T {y}_j^{'} \right) W_{i,j} \nonumber \\
    &= 2q\sqrt{mn} - 2 \sum_{i,j} \left( ({x}_i^{'})^T {y}_j^{'} \right) W_{i,j} 
    = 2q\sqrt{mn} - 2 \sum_{k=1}^q \left[x_1^{'}[k],\ldots,x_m^{'}[k]\right] W \left[y_1^{'}[k],\ldots,y_n^{'}[k]\right]^T, \label{eq: simplified objective for embedding}
\end{align}
where we used the fact that $W\in \mathcal{B}_{m,n}$. By the properties of the SVD and the fact that the first pair of singular vectors of $W$ is trivial (see text after~\eqref{eq: SVD of W}), it follows that taking $\left[x_1^{'}[k],\ldots,x_m^{'}[k]\right]^T = \sqrt{m}\mathbf{u}_k$ and $\left[y_1^{'}[k],\ldots,y_n^{'}[k]\right]^T = \sqrt{n}\mathbf{v}_k$ for $k=2,\ldots,q\leq m-1$ minimizes~\eqref{eq: simplified objective for embedding} under the constraints~\eqref{eq: constraints for embedding}.

\section{Proof of Proposition~\ref{prop: spectrum of L}} \label{appendix: proof of spectrum of L}
To prove the first part of Proposition~\ref{prop: spectrum of L}, it can be verified directly that the vectors 
        \begin{align}
        \{\phi_1,\ldots,\phi_{m+n}\} = 
        \bigg\{
            &\begin{bmatrix}
                \frac{1}{\sqrt{2m}}\mathbf{1}_m \\ \frac{1}{\sqrt{2n}}\mathbf{1}_n   
            \end{bmatrix},
            \begin{bmatrix}
                \frac{1}{\sqrt{2}} \mathbf{u}_2 \\ \frac{1}{\sqrt{2}} \mathbf{v}_2 
            \end{bmatrix},
            \ldots,
            \begin{bmatrix}
                \frac{1}{\sqrt{2}} \mathbf{u}_m \\ \frac{1}{\sqrt{2}} \mathbf{v}_m 
            \end{bmatrix},
            \nonumber \\
            &\begin{bmatrix}
                \mathbf{0}_m \\ {\mathbf{v}}_{m+1}
            \end{bmatrix},
            \ldots,
            \begin{bmatrix}
                \mathbf{0}_m \\ {\mathbf{v}}_{m+n} 
            \end{bmatrix}, 
            \begin{bmatrix}
                \frac{1}{\sqrt{2}} \mathbf{u}_m \\ -\frac{1}{\sqrt{2}} \mathbf{v}_m 
            \end{bmatrix},
            \ldots,
            \begin{bmatrix}
                \frac{1}{\sqrt{2}} \mathbf{u}_2 \\ -\frac{1}{\sqrt{2}} \mathbf{v}_2 
            \end{bmatrix},
            \begin{bmatrix}
                \frac{1}{\sqrt{2m}}\mathbf{1}_m \\ -\frac{1}{\sqrt{2n}}\mathbf{1}_n   
            \end{bmatrix}
            \bigg\}, \label{eq: eigenvectors of L}
        \end{align}
are eigenvectors of $L$ with eigenvalues \sloppy $\{\lambda_1,\ldots,\lambda_{m+n}\} = \{ 0, 1-{s_2},\ldots,1-{s_m},1,\ldots,1,1+{s_m},\ldots,1+{s_2},2\}$,
where $\{\mathbf{v}_{m+1},\ldots,\mathbf{v}_{m+n}\} \subset \mathbb{R}^n$ is any set of orthonormal vectors that are orthogonal to $\{\mathbf{v}_1,\ldots,\mathbf{v}_m\}$. To prove the quadratic form, we write
\begin{align}
    &\frac{1}{\sqrt{mn}} \sum_{{i,j}} \left( \sqrt{m} \mathbf{g}[i] - \sqrt{n} \mathbf{h}[j] \right)^2 W_{i,j} \nonumber \\
    &= \sqrt{\frac{m}{n}} \sum_i \left( \mathbf{g}[i]\right)^2\sum_j W_{i,j} + \sqrt{\frac{n}{m}} \sum_j \left( \mathbf{h}[j]\right)^2\sum_i W_{i,j} - 2 \sum_{{i,j}} \mathbf{g}[i] W_{i,j} \mathbf{h}[j] \nonumber \\
    &= \Vert \mathbf{g} \Vert_2^2 + \Vert \mathbf{h} \Vert_2^2 - 2 \mathbf{g}^T W \mathbf{h}
    = \Vert \mathbf{f} \Vert_2^2 - \mathbf{f}^T \hat{W} \mathbf{f}
    = \mathbf{f}^T L \mathbf{f},
\end{align}
where we used the fact that $W\in \mathcal{B}_{m,n}$ (see~\eqref{eq: B def}). For the second part of Proposition~\ref{prop: spectrum of L}, the facts that $\widetilde{L}$ shares the eigenvalues of $L$ and $\psi_k = D \phi_k/\Vert D \phi_k\Vert_2$ follow from the similarity of $\widetilde{L} = D L D^{-1}$ to $L$. Therefore, eq.~\eqref{eq: embedding interpretation with eigenvectors of P} follows immediately from~\eqref{eq: eigenvectors of L} and the embedding formula~\eqref{eq: embedding formula}.

\section{Proof of Proposition~\ref{prop: diffusion distance identities}} \label{appendix: proof of diffusion distance identities}
For even $t$, we have
\begin{align}
        D_{\mathcal{X}}^{(t)}(\mathbf{x}_i,\mathbf{x}_{i^{'}}) 
        &= \sqrt{m} \left\Vert \left[P_{\mathcal{XX}}^{(t)}\right]_{i,\cdot} - \left[P_{\mathcal{XX}}^{(t)}\right]_{i^{'},\cdot} \right\Vert_2 
        = \sqrt{m} \left\Vert \sum_{k=1}^m \mathbf{u}_k[i] s_k \mathbf{u}_k^T - \sum_{k=1}^m \mathbf{u}_k[i^{'}] s_k \mathbf{u}_k^T \right\Vert_2 \nonumber \\
        &= \sqrt{\sum_{k=1}^m {s_k^{2t}}  \left( \sqrt{m} \mathbf{u}_k[i] - \sqrt{m} \mathbf{u}_k[i^{'}] \right)^2} 
        = \sqrt{\sum_{k=2}^m {s_k^{2t}}  \left( \sqrt{m} \mathbf{u}_k[i] - \sqrt{m} \mathbf{u}_k[i^{'}] \right)^2} = \Vert \widetilde{x}_i - \widetilde{x}_{i^{'}} \Vert_2, \label{eq: D_X_t event t formula} \\
        D_{\mathcal{Y}}^{(t)}(\mathbf{y}_j,\mathbf{y}_{j^{'}}) 
        &= \sqrt{n}\left\Vert \left[P_{\mathcal{YY}}^{(t)}\right]_{j,\cdot} - \left[P_{\mathcal{YY}}^{(t)}\right]_{j^{'},\cdot} \right\Vert_2 
        = \sqrt{n} \left\Vert \sum_{k=1}^m \mathbf{v}_k[j] s_k \mathbf{v}_k^T - \sum_{k=1}^m \mathbf{v}_k[j^{'}] s_k \mathbf{v}_k^T \right\Vert_2 \nonumber \\
        &=\sqrt{\sum_{k=1}^m {s_k^{2t}} \left( \sqrt{n} \mathbf{v}_k[j] - \sqrt{n} \mathbf{v}_k[j^{'}] \right)^2} = 
        \sqrt{\sum_{k=2}^m {s_k^{2t}} \left( \sqrt{n} \mathbf{v}_k[j] - \sqrt{n} \mathbf{v}_k[j^{'}] \right)^2}
        = \Vert \widetilde{y}_j - \widetilde{y}_{j^{'}} \Vert_2, \label{eq: D_Y_t even t formula}
    \end{align}
where we used the fact that $\mathbf{u}_1 = \mathbf{1}_m/\sqrt{m}$ and $\mathbf{v}_1 = \mathbf{1}_n/\sqrt{n}$. Similarly, for odd $t$,
\begin{align}
        D_{\mathcal{X}}^{(t)}(\mathbf{x}_i,\mathbf{x}_{i^{'}}) 
        &= \sqrt{n} \left\Vert \left[P_{\mathcal{XY}}^{(t)}\right]_{i,\cdot} - \left[P_{\mathcal{XY}}^{(t)}\right]_{i^{'},\cdot} \right\Vert_2 
        = \sqrt{n} \left\Vert \sqrt{\frac{m}{n}} \sum_{k=1}^m \mathbf{u}_k[i] s_k \mathbf{v}_k^T - \sqrt{\frac{m}{n}} \sum_{k=1}^m \mathbf{u}_k[i^{'}] s_k \mathbf{v}_k^T \right\Vert_2 \nonumber \\
        &= \sqrt{\sum_{k=1}^m {s_k^{2t}}  \left( \sqrt{m} \mathbf{u}_k[i] - \sqrt{m} \mathbf{u}_k[i^{'}] \right)^2}  =\sqrt{\sum_{k=2}^m {s_k^{2t}}  \left( \sqrt{m} \mathbf{u}_k[i] - \sqrt{m} \mathbf{u}_k[i^{'}] \right)^2}
        = \Vert \widetilde{x}_i - \widetilde{x}_{i^{'}} \Vert_2, \label{eq: D_X_t odd t formula} \\
        D_{\mathcal{Y}}^{(t)}(\mathbf{y}_j,\mathbf{y}_{j^{'}}) 
        &= \sqrt{m}\left\Vert \left[P_{\mathcal{YX}}^{(t)}\right]_{j,\cdot} - \left[P_{\mathcal{YX}}^{(t)}\right]_{j^{'},\cdot} \right\Vert_2 
        = \sqrt{m} \left\Vert \sqrt{\frac{n}{m}} \sum_{k=1}^m \mathbf{v}_k[j] s_k \mathbf{u}_k^T - \sqrt{\frac{n}{m}} \sum_{k=1}^m \mathbf{v}_k[j^{'}] s_k \mathbf{u}_k^T \right\Vert_2 \nonumber \\
        &=\sqrt{\sum_{k=1}^m {s_k^{2t}} \left( \sqrt{n} \mathbf{v}_k[j] - \sqrt{n} \mathbf{v}_k[j^{'}] \right)^2} 
        = \sqrt{\sum_{k=2}^m {s_k^{2t}} \left( \sqrt{n} \mathbf{v}_k[j] - \sqrt{n} \mathbf{v}_k[j^{'}] \right)^2}
        =\Vert \widetilde{y}_j - \widetilde{y}_{j^{'}} \Vert_2. \label{eq: D_Y_t odd t formula}
    \end{align}
Lastly, we have
\begin{align}
D_{\mathcal{X}\mathcal{Y}}^{(t)}(x_i,y_j) &= \sqrt{m}\left\Vert \left[P_{\mathcal{XX}}^{(t)}\right]_{i,\cdot} - \left[P_{\mathcal{YX}}^{(t)}\right]_{j,\cdot} \right\Vert_2 
= \sqrt{m} \left\Vert \sum_{k=1}^m \mathbf{u}_k[i] s_k \mathbf{u}_k^T - \sqrt{\frac{n}{m}} \sum_{k=1}^m \mathbf{v}_k[j] s_k \mathbf{u}_k^T \right\Vert_2 \nonumber \\
&= \sqrt{\sum_{k=1}^m {s_k^{2t}} \left( \sqrt{m} \mathbf{u}_k[i] - \sqrt{n} \mathbf{v}_k[j] \right)^2} 
= \sqrt{\sum_{k=2}^m {s_k^{2t}} \left( \sqrt{m} \mathbf{u}_k[i] - \sqrt{n} \mathbf{v}_k[j] \right)^2}
= \Vert \widetilde{x}_i - \widetilde{y}_{j} \Vert_2, \label{eq: D_XY_t formula} \\
    D_{\mathcal{Y}\mathcal{X}}^{(t)}(y_j,x_i) &= 
    \sqrt{n}\left\Vert \left[P_{\mathcal{YY}}^{(t)}\right]_{j,\cdot} - \left[P_{\mathcal{XY}}^{(t)}\right]_{i,\cdot} \right\Vert_2
    = \sqrt{n} \left\Vert \sum_{k=1}^m \mathbf{v}_k[j] s_k \mathbf{v}_k^T - \sqrt{\frac{m}{n}} \sum_{k=1}^m \mathbf{u}_k[i] s_k \mathbf{v}_k^T \right\Vert_2 \nonumber \\
&= \sqrt{\sum_{k=1}^m {s_k^{2t}} \left( \sqrt{m} \mathbf{u}_k[i] - \sqrt{n} \mathbf{v}_k[j] \right)^2} 
= \sqrt{\sum_{k=2}^m {s_k^{2t}} \left( \sqrt{m} \mathbf{u}_k[i] - \sqrt{n} \mathbf{v}_k[j] \right)^2}
= \Vert \widetilde{x}_i - \widetilde{y}_{j} \Vert_2, \label{eq: D_YX_t formula}
\end{align}
which concludes the proof.

\section{Truncating the embedding and diffusion distance for $t>0$} \label{sec: truncating the diffusion distance}
{While proposition~\ref{prop: diffusion distance identities} is stated for the embedding dimension $m-1$, we can obtain embeddings with lower dimensions by simply discarding embedding coordinates whose magnitudes are sufficiently small. As we increase $t$, we can discard more embedding coordinates, relying on the fact that $s_k^{2t}$ decays rapidly with $t$ for any $s_k<1$. For instance, to obtain an approximate version of the identity~\eqref{eq: diffusion distances x-x} for $q<m-1$ with controlled error, we can take the embedding dimension $q$ such that the residual error satisfies
\begin{equation}
    \sum_{k=q+1}^m {s_k^{2t}}  \left( \sqrt{m} \mathbf{u}_k[i] - \sqrt{n} \mathbf{v}_k[j] \right)^2 \leq (\sqrt{m}+\sqrt{n})^2 s_{q+1}^{2t} \leq \delta, \label{eq: diffusion distance approx}
\end{equation}
where $\delta$ is some prescribed tolerance threshold and we used the orthonormality of the singular vectors to obtain the first inequality. Thus, the identity~\eqref{eq: diffusion distances x-x} would be preserved up to an error of $\delta$. Similarly, we can approximate~\eqref{eq: diffusion distances y-y} and~\eqref{eq: diffusion distances x-y} up to an error of $\delta$ using the leading $q<m-1$ embedding coordinates if $4{m} s_{q+1}^{2t}$ and $4{n} s_{q+1}^{2t}$ are smaller than $\delta$. This approximation can achieve high accuracy for small embedding dimensions $q$, even for moderate values of $t$, due to the fast exponential decay of the residual error with $t$. This discussion is analogous to the one in~\cite{coifman2006diffusionMaps} on the classical diffusion distance when embedding a single dataset.}

\section{Supporting lemmas and definitions for Theorem~\ref{thm: concentration of transport plan}}
\subsection{Matrix scaling} \label{appendix: stability of matrix scaling}
We now provide some basic results on the stability of matrix scaling~\cite{sinkhorn1964relationship,sinkhorn1967diagonal}, namely, the process of diagonally scaling a positive matrix $A\in\mathbb{R}^{m\times n}$ to have certain prescribed row and column sums. The definitions and results below are adapted from~\cite{landa2022scaling}. Let $\alpha = [\alpha_1,\ldots,\alpha_m]$, $\beta = [\beta_1,\ldots,\beta_n]$, $\mathbf{r}=[r_1,\ldots,r_m]$, and $\mathbf{c}=[c_1,\ldots,c_n]$ be (entrywise) positive vectors with $\Vert \mathbf{r} \Vert_1 = \Vert \mathbf{c} \Vert_1$, and denote $B = \operatorname{diag}\{\alpha\} A \operatorname{diag}\{\beta\}$, where $\operatorname{diag}\{\alpha\}$ denotes a diagonal matrix with $\alpha$ on its main diagonal. We say that the pair of vectors $(\alpha,\beta)$ \textit{scales} $A$ to row sums $\mathbf{r}$ and column sums $\mathbf{c}$, if
\begin{equation}
    r_i = \sum_{j=1}^n B_{i,j} = \sum_{j=1}^n \alpha_i A_{i,j} \beta_j, \qquad \text{and} \qquad c_j = \sum_{i=1}^m B_{i,j} = \sum_{i=1}^m \alpha_i A_{i,j} \beta_j, \label{eq:equations for x and y in matrix scaling definition}
\end{equation}
for all $i\in[m]$ and  $j\in [n]$. We refer to $\alpha$ and $\beta$ from~\eqref{eq:equations for x and y in matrix scaling definition} (or their entries) as \textit{scaling factors} of $A$.
Let us define
\begin{equation}
    \overline{r}_i = \frac{r_i}{\sqrt{s}}, \qquad\qquad \overline{c}_j = \frac{c_j}{\sqrt{s}}, \qquad\qquad s = \Vert \mathbf{r} \Vert_1 = \Vert \mathbf{c} \Vert_1, \label{eq:r_overline and c_overline def}
\end{equation}
for  $i\in [m]$ and  $j\in [n]$. 
The following lemma describes a useful normalization of the scaling factors and the resulting bounds on their entries; see~\cite{landa2022scaling}.
\begin{lem}[Boundedness of scaling factors~\cite{landa2022scaling}] \label{lemma:boundedness of scaling factors}
Let $A\in\mathbb{R}^{m\times n}$ be a matrix with positive entries and denote $a = \min_{i,j} A_{i,j}$ and $b = \max_{i,j} A_{i,j}$. Then, there exists a unique pair of positive vectors $(\alpha,\beta)$ with $\Vert \alpha \Vert_1 = \Vert \beta \Vert_1$ that scales $A$ to row sums $\mathbf{r}$ and column sums $\mathbf{c}$. Moreover, for all $i\in [m]$ and $j\in [n]$, we have
    \begin{equation}
        \frac{\sqrt{a}}{b} \leq \frac{\alpha_i}{\overline{r}_i} \leq \frac{\sqrt{b}}{a}, \qquad \qquad  \frac{\sqrt{a}}{b} \leq \frac{\beta_j}{\overline{c}_j} \leq \frac{\sqrt{b}}{a}. \label{eq:x_i and y_i bounds}
    \end{equation}
\end{lem}

The following lemma characterizes the stability of the scaling factors to perturbations in the row and column sums. In particular, it states that if there exists a pair of positive vectors $(\hat{\alpha},\hat{\beta})$ that approximately scales $A$ to certain row and column sums, then $(\hat{\alpha},\hat{\beta})$ must be close to a true pair of scaling factors for these row and column sums. 
\begin{lem} [Stability of scaling factors under approximate scaling] \label{lem:closeness of scaling factors}
Let $A\in\mathbb{R}^{m\times n}$ be a matrix with positive entries and denote $a = \min_{i,j} {{A}}_{i,j}$, $b = \max_{i,j} {A}_{i,j}$. Suppose that there exists $\epsilon\in (0,1)$ and positive vectors $\hat{\alpha} = [\hat{\alpha}_1,\ldots,\hat{\alpha}_m]$ and $\hat{\beta} = [\hat{\beta}_1,\ldots,\hat{\beta}_n]$, such that
\begin{equation}
\left\vert \frac{1}{c_j} \sum_{i=1}^m \hat{\alpha}_i A_{i,j} \hat{\beta}_j - 1 \right\vert \leq \epsilon, \qquad \left\vert \frac{1}{r_i} \sum_{j=1}^n \hat{\alpha}_i A_{i,j} \hat{\beta}_j -1 \right\vert \leq \epsilon, \label{eq:approximate scaling condition in lemma}
\end{equation}
for all $i\in [m]$ and $j\in [n]$. Then, there exists a pair of positive vectors $({\alpha},{\beta})$ that scales $A$ to row sums $\mathbf{r}$ and column sums $\mathbf{c}$, satisfying
\begin{align}
     \frac{\vert{\alpha}_i - \hat{\alpha}_i \vert}{\hat{\alpha}_i} &\leq \epsilon \left(\frac{1}{1-\epsilon} + \frac{4 s \sqrt{b} }{a^2 C_1^{3/2} C_2^{3/2} m \min_i {r}_i } \right), \\
    \frac{\vert{\beta}_j - \hat{\beta}_j \vert}{\hat{\beta}_j} &\leq \epsilon \left( \frac{1}{1-\epsilon} + \frac{4  s \sqrt{b} }{a^2 C_1^{3/2} C_2^{3/2} n \min_j {c}_j } \right),
\end{align}
for all $i\in [m]$ and $j\in [n]$, where $C_1 = \min_{i} \{\hat{\alpha}_i/\overline{r}_i\}$ and $C_2 = \min_j \{\hat{\beta}_j/\overline{c}_j\}$.
\end{lem}
An argument similar to Lemma~\ref{lem:closeness of scaling factors} is outlined inside the proof of Theorem 2 in~\cite{landa2022scaling} in a more specialized setting. For completeness, we provide the full proof of Lemma~\ref{lem:closeness of scaling factors} below.
\begin{proof}
Let $(\widetilde{\alpha},\widetilde{\beta})$ be the unique pair of scaling factors of $A$ with $\Vert \widetilde{\alpha} \Vert_1 = \Vert \widetilde{\beta} \Vert_1$ (see Lemma~\ref{lemma:boundedness of scaling factors}), and define
\begin{equation}
    B_{i,j} = \hat{\alpha}_i A_{i,j} \hat{\beta}_j, \qquad \widetilde{B}_{i,j} = \widetilde{\alpha}_i A_{i,j} \widetilde{\beta}_j = u_i B_{i,j} v_j, \qquad u_i = \frac{\widetilde{\alpha}_i}{\hat{\alpha}_i}, \qquad v_j = \frac{\widetilde{\beta}_j}{\hat{\beta}_j}, \label{eq:P,P_hat,u, and v def}
\end{equation}
for all $i\in [m]$ and $j\in [n]$. Observe that $\sum_i \widetilde{B}_{i,j} = c_j$, $\sum_j \widetilde{B}_{i,j} = r_i$, and
\begin{equation}
    \left\vert \frac{1}{c_j} \sum_{i=1}^m B_{i,j} - 1 \right\vert \leq \epsilon, \qquad\qquad \left\vert \frac{1}{r_i} \sum_{j=1}^n B_{i,j} -1 \right\vert \leq \epsilon, \label{eq:u_i P_{i,j} v_j sum closeness bounds}
\end{equation}
for all $i\in[m]$, $j\in[n]$.
Let
\[
j_{\min} \in \arg\min_{k\in[n]} v_k,\qquad j_{\max} \in \arg\max_{k\in[n]} v_k,
\qquad
i_{\min} \in \arg\min_{\ell\in[m]} u_\ell,\qquad i_{\max} \in \arg\max_{\ell\in[m]} u_\ell.
\]
Using the first inequality in~\eqref{eq:u_i P_{i,j} v_j sum closeness bounds} with $j=j_{\min}$, we have
\begin{equation}
    1 = \frac{1}{c_{j_{\min}}} \sum_{i=1}^m \widetilde{B}_{i,j_{\min}}
    = \frac{1}{c_{j_{\min}}}\sum_{i=1}^m u_i B_{i,j_{\min}} v_{j_{\min}}
    \leq \big(\min_j v_j\big)\big(\max_i u_i\big)\frac{1}{c_{j_{\min}}} \sum_{i=1}^m B_{i,j_{\min}}
    \leq (1+\epsilon)\min_j v_j \max_i u_i. \label{eq:min v max u lower bound}
\end{equation}
Similarly, using the second inequality in~\eqref{eq:u_i P_{i,j} v_j sum closeness bounds} with $i=i_{\max}$ gives
\begin{equation}
    1 = \frac{1}{r_{i_{\max}}}\sum_{j=1}^n \widetilde{B}_{i_{\max},j}
    = \frac{1}{r_{i_{\max}}}\sum_{j=1}^n u_{i_{\max}} B_{i_{\max},j} v_j
    \geq \big(\max_i u_i\big)\big(\min_j v_j\big)\frac{1}{r_{i_{\max}}} \sum_{j=1}^n B_{i_{\max},j}
    \geq (1-\epsilon)\max_i u_i \min_j v_j, \label{eq: max u min v upper bound}
\end{equation}
and by combining~\eqref{eq:min v max u lower bound} and~\eqref{eq: max u min v upper bound} we obtain
\begin{equation}
    \frac{1}{1+\epsilon} \leq \max_i u_i \min_j v_j \leq \frac{1}{1-\epsilon}. \label{eq:max u min v upper and lower bounds}
\end{equation}
Analogously, using the first inequality in~\eqref{eq:u_i P_{i,j} v_j sum closeness bounds} with $j=j_{\max}$ and the second inequality with $i=i_{\min}$ yields
\begin{equation}
    \frac{1}{1+\epsilon} \leq \min_i u_i \max_j v_j \leq \frac{1}{1-\epsilon}. \label{eq:min u max v upper and lower bounds}
\end{equation}

Note that~\eqref{eq:min u max v upper and lower bounds} can also be obtained directly from~\eqref{eq:max u min v upper and lower bounds} by a symmetry argument, that is, by considering~\eqref{eq:max u min v upper and lower bounds} in the setting when $A$ is replaced with its transpose, thereby interchanging the roles of $\mathbf{u}$ and $\mathbf{v}$.  
In addition, according to Lemma~\ref{lemma:boundedness of scaling factors} and using the fact that $\hat{\alpha}_i\geq C_1 \overline{r}_i$ and $\hat{\beta}_j\geq C_2 \overline{c}_j$ (from the definition of $C_1$ and $C_2$ in Lemma~\ref{lem:closeness of scaling factors}), it follows that for all $i\in [m]$ and $j\in [n]$
\begin{equation}
    u_i \leq \frac{\sqrt{b}}{a C_1}, \qquad \qquad v_j \leq \frac{\sqrt{b}}{a C_2}. \label{eq:u_i and v_j bounds}
\end{equation}

Let us denote $\ell = \operatorname{argmax}_i u_i$. By the second inequality in~\eqref{eq:u_i P_{i,j} v_j sum closeness bounds} together with~\eqref{eq:max u min v upper and lower bounds}, we can write
\begin{equation}
    1 = \frac{1}{r_\ell}\sum_{j=1}^n \widetilde{B}_{\ell,j} = \frac{1}{r_\ell}\sum_{j=1}^n u_\ell B_{\ell,j} v_j \geq \frac{1}{(1+\epsilon) r_\ell}\sum_{j=1}^n B_{\ell,j} \frac{v_j}{\min_j v_j}, \label{eq:bounding max v - min v step 1}
\end{equation}
implying that
\begin{equation}
    \frac{1}{r_\ell} \sum_{j=1}^n B_{\ell,j} \left( \frac{v_j}{\min_j v_j} -1 \right) \leq {1+\epsilon} - \frac{1}{r_\ell}\sum_{j=1}^n B_{\ell,j} \leq {2\epsilon}. \label{eq:preliminary bound 1}
\end{equation}
Multiplying~\eqref{eq:preliminary bound 1} by $\min_j v_j /\min_j  B_{\ell,j}$, it follows that
\begin{equation}
   \frac{1}{r_\ell} \sum_{j=1}^n ({v_j} - \min_j v_j ) \leq \frac{1}{r_\ell}  \sum_{j=1}^n \frac{B_{\ell,j}}{\min_j B_{\ell,j} } ({v_j} - \min_j v_j ) \leq 2\epsilon \frac{\min_j v_j}{\min_j B_{\ell,j} } \leq  \frac{2\epsilon \min_j v_j}{a C_1 C_2 \overline{r}_\ell \min_j \overline{c}_j }, \label{eq:sum v_j -min v_j bound intermediate}
\end{equation}
where we used the definition of $B$ together with the conditions in Lemma~\ref{lem:closeness of scaling factors}.
Multiplying~\eqref{eq:sum v_j -min v_j bound intermediate} by $r_\ell/n$ and employing the definitions of $\overline{r}_i$ and $\overline{c}_j$ (see~\eqref{eq:r_overline and c_overline def}) gives
\begin{equation}
    \frac{1}{n} \sum_{j=1}^n ({v_j} - \min_j v_j ) \leq \frac{2\epsilon {s} \min_j v_j}{a C_1 C_2 n \min_j {c}_j } \leq \frac{2\epsilon {s} \max_j v_j}{a C_1 C_2 n \min_j {c}_j }. \label{eq:bounding max v - min v step 2}
\end{equation}

We next provide a derivation analogous to~\eqref{eq:bounding max v - min v step 1}--\eqref{eq:bounding max v - min v step 2} to obtain a bound for $\frac{1}{n} \sum_{j=1}^n (\max_j v_j - v_j )$. 
Let us denote $t = \operatorname{argmin}_i u_i$. Using the second inequality in~\eqref{eq:u_i P_{i,j} v_j sum closeness bounds} together with~\eqref{eq:min u max v upper and lower bounds}, we have
\begin{equation}
    1 = \frac{1}{r_t}\sum_{j=1}^n \widetilde{B}_{t,j} = \frac{1}{r_t}\sum_{j=1}^n u_t B_{t,j} v_j \leq \frac{1}{(1-\epsilon)r_t}\sum_{j=1}^n B_{t,j} \frac{v_j}{\max_j v_j},
\end{equation}
and therefore
\begin{equation}
    \frac{1}{r_t} \sum_{j=1}^n B_{t,j} \left( 1-\frac{v_j}{\max_j v_j}\right) \leq \frac{1}{r_t}\sum_{j=1}^n B_{t,j} - (1-\epsilon) \leq 2\epsilon.
\end{equation}
Multiplying the above by $\max_j v_j /\min_j  B_{t,j}$, it follows that
\begin{equation}
    \frac{1}{r_t} \sum_{j=1}^n (\max_j v_j  - v_j) \leq \frac{1}{r_t} \sum_{j=1}^n \frac{B_{t,j}}{\min_j  B_{t,j}} (\max_j v_j  - v_j) \leq 2\epsilon \frac{\max_j v_j}{\min_j  B_{t,j}} \leq \frac{2\epsilon \max_j v_j}{a C_1 C_2 \overline{r}_t \min_j \overline{c}_j }.
\end{equation}
Furthermore, multiplying the above by $r_t/n$ and using the definitions of $\overline{r}_i$ and $\overline{c}_j$ (see~\eqref{eq:r_overline and c_overline def}), we get
\begin{equation}
   \frac{1}{n} \sum_{j=1}^n (\max_j v_j  - v_j) \leq 2\epsilon \frac{\max_j v_j}{\min_j B_{t,j} } \leq \frac{2\epsilon s \max_j v_j }{a C_1 C_2 n \min_j {c}_j }. \label{eq:bounding max v - min v step 3}
\end{equation}

Lastly, summing~\eqref{eq:bounding max v - min v step 2} and~\eqref{eq:bounding max v - min v step 3} gives
\begin{equation}
    \max_j v_j - \min_j v_j \leq \frac{4 \epsilon s \max_j v_j }{a C_1 C_2 n \min_j {c}_j }. \label{eq:max v_j - min v_j bound}
\end{equation}
It is easy to verify that by repeating the derivation of~\eqref{eq:bounding max v - min v step 1} --~\eqref{eq:max v_j - min v_j bound} analogously for $u_i$ instead of $v_j$, we get
\begin{equation}
    \max_i u_i - \min_i u_i \leq \frac{4 \epsilon s \max_i u_i }{a C_1 C_2 m \min_i {r}_i }. \label{eq:max u_i - min u_i bound}
\end{equation}
We omit the full derivation for the sake of brevity. Note that~\eqref{eq:max u_i - min u_i bound} can also be obtained directly from~\eqref{eq:max v_j - min v_j bound} by a symmetry argument, namely by considering~\eqref{eq:max v_j - min v_j bound} in the setting where $A$ is replaced with its transpose, so that $n$ is replaced with $m$, $\mathbf{c}$ is replaced with $\mathbf{r}$, and $\mathbf{v}$ is replaced with $\mathbf{u}$.

Observe that $|\tau - v_j| \leq \max_j v_j - \min_j v_j$ for any $\tau \in [\min_j v_j,\max_j v_j]$ and all $j\in[n]$. Taking $\tau$ as the geometric mean of $\max_j v_j$ and $\min_j v_j$, together with~\eqref{eq:max v_j - min v_j bound} gives
\begin{equation}
    \left\vert\sqrt{\max_j v_j \min_j v_j} -  v_j \right\vert \leq \frac{4 \epsilon s \max_j v_j }{a C_1 C_2 n \min_j {c}_j }, \label{eq:max v min u geometric mean bound}
\end{equation}
for all $j\in [n]$. Multiplying both hand sides of~\eqref{eq:max v min u geometric mean bound} by $\gamma^{-1} = \sqrt{\max_i u_i / \max_j v_j}$ we get
\begin{equation}
    \left\vert\sqrt{\max_i u_i \min_j v_j} - \gamma^{-1} v_j \right\vert 
    \leq \frac{4 \epsilon s \sqrt{\max_j v_j \max_i  u_i} }{a C_1 C_2 n \min_j {c}_j } 
    \leq \frac{4 \epsilon s \sqrt{b} }{a^2 C_1^{3/2} C_2^{3/2} n \min_j {c}_j },
\end{equation}
where we also used~\eqref{eq:u_i and v_j bounds} in the last inequality. According to~\eqref{eq:max u min v upper and lower bounds}, we have for all $\epsilon\in (0,1)$ that
\begin{equation}
    1-\frac{\epsilon}{1-\epsilon} \leq {\frac{1}{1+\epsilon}} \leq \sqrt{\frac{1}{1+\epsilon}} \leq \sqrt{\max_i u_i \min_j v_j} \leq  \sqrt{\frac{1}{1-\epsilon}} \leq \frac{1}{1-\epsilon} = 1 + \frac{\epsilon}{1-\epsilon},
\end{equation}
which together with~\eqref{eq:max v min u geometric mean bound} implies that
\begin{equation}
    \left\vert 1 - \gamma^{-1} v_j \right\vert 
    \leq \frac{\epsilon}{1-\epsilon} + \frac{4 \epsilon s \sqrt{b} }{a^2 C_1^{3/2} C_2^{3/2} n \min_j {c}_j }. \label{eq:1 - alpha_inv v_j bound}
\end{equation}

Analogously to~\eqref{eq:max v min u geometric mean bound}, from~\eqref{eq:max u_i - min u_i bound} we obtain
\begin{equation}
    \left\vert\sqrt{\max_i u_i \min_i u_i} -  u_i \right\vert \leq \frac{4 \epsilon s \max_i u_i }{a C_1 C_2 m \min_i {r}_i }, \label{eq:max u min u geometric mean bound}
\end{equation}
for all $i\in [m]$. Multiplying both hand sides of~\eqref{eq:max u min u geometric mean bound} by $\gamma = \sqrt{ \max_j v_j / \max_i u_i }$ we get
\begin{equation}
    \left\vert\sqrt{\max_j v_j \min_i u_i} - \gamma u_i \right\vert 
    \leq \frac{4 \epsilon s \sqrt{\max_j v_j \max_i u_i} }{a C_1 C_2 m \min_i {r}_i } 
    \leq \frac{4 \epsilon s \sqrt{b} }{a^2 C_1^{3/2} C_2^{3/2} m \min_i {r}_i }.
\end{equation}
Consequently, using~\eqref{eq:min u max v upper and lower bounds}, and analogously to the derivation of~\eqref{eq:1 - alpha_inv v_j bound}, it follows that
\begin{equation}
    \left\vert 1 - \gamma u_i \right\vert 
    \leq \frac{\epsilon}{1-\epsilon} + \frac{4 \epsilon s \sqrt{b} }{a^2 C_1^{3/2} C_2^{3/2} m \min_i {r}_i }, \label{eq:1 - alpha u_i bound}
\end{equation}
which together with the definition of $\mathbf{u}$ and $\mathbf{v}$ in~\eqref{eq:P,P_hat,u, and v def} concludes the proof, taking $\alpha = \gamma \widetilde{\alpha}$ and $\beta = \gamma^{-1} \widetilde{\beta}$.

\end{proof}

\subsection{Boundedness of the scaling function}
The following lemma provides upper and lower bounds on the scaling function $\rho_{\varepsilon} (x)$ satisfying~\eqref{eq:integral scaling eq with density} for $x\in \mathcal{N}$.
\begin{lem} \label{lem: boundedness of the scaling function}
    Under Assumption~\ref{assump: noise magnitude}, we have for all $x\in \mathcal{N}$,
    \begin{equation}
        \exp \left\{ -\frac{2C^4}{\varepsilon} \right\} \leq \frac{\rho_{\varepsilon} (x)}{(\pi \varepsilon)^{d/4}} \leq \exp \left\{ \frac{4C^4}{\varepsilon} \right\}.
    \end{equation}
\end{lem}
\begin{proof}
    Let $x,y \in \mathcal{N}$ and define $x = \mathcal{A} \tilde{x} $ and $y = \mathcal{A} \tilde{y} $ for $\tilde{x}, \tilde{y} \in \mathcal{M}$. Under Assumption~\ref{assump: noise magnitude} and the definition of $\mathcal{A}$ in~\eqref{eq: A_mathcal def},
\begin{equation}
    0 \leq \Vert x - y \Vert_2^2 \leq \Vert \mathcal{A} (\tilde{x} - \tilde{y}) \Vert_2^2 \leq \Vert \mathcal{A} \Vert_2^2 \, \Vert \tilde{x} - \tilde{y} \Vert_2^2 \leq \Vert \mathcal{A}_1 \Vert_2 \, \Vert \mathcal{A}_2 \Vert_2 \, 4C^2 \leq 4C^4,
\end{equation}
for all $x,y \in \mathcal{N}$. Hence,
    \begin{equation}
        \frac{1}{(\pi \varepsilon)^{d/2}}\exp\left\{ -\frac{4C^4}{\varepsilon} \right\} \leq \mathcal{K}_\varepsilon (x,y) \leq \frac{1}{(\pi \varepsilon)^{d/2}}, \label{eq: boundedness of the Gaussian kernel}
    \end{equation}
    and combining this with~\eqref{eq:integral scaling eq with density}, we have
    \begin{equation}
        {\rho_\varepsilon(x)} \exp\left\{ -\frac{4C^4}{\varepsilon} \right\} \int_{\mathcal{N}} \rho_\varepsilon(y) \tilde{w}(y) d\nu(y) 
        \leq (\pi \varepsilon)^{d/2}
        \leq {\rho_\varepsilon(x)} \int_{\mathcal{N}} \rho_\varepsilon(y) \tilde{w}(y) d\nu(y), \label{eq: integral bound for rho}
    \end{equation} 
for all $x\in \mathcal{N}$. Integrating all sides of the inequality above over $x\in \mathcal{N}$ with respect to the probability measure $\tilde{w}\, d\nu$ (recalling that $\int_{\mathcal{N}} \tilde{w}(x) d\nu(x) = 1$) gives
\begin{equation}
    \exp\left\{ -\frac{4C^4}{\varepsilon} \right\} \left( \int_{\mathcal{N}} \rho_\varepsilon(y) \tilde{w}(y) d\nu(y) \right)^2 
    \leq (\pi \varepsilon)^{d/2} \leq 
    \left( \int_{\mathcal{N}} \rho_\varepsilon(y) \tilde{w}(y) d\nu(y) \right)^2,
\end{equation}
which implies that
\begin{equation}
    (\pi \varepsilon)^{d/4} \leq \int_{\mathcal{N}} \rho_\varepsilon(y) \tilde{w}(y) d\nu(y) \leq (\pi \varepsilon)^{d/4} \exp\left\{ \frac{2C^4}{\varepsilon} \right\}.
\end{equation}
Substituting the above back into~\eqref{eq: integral bound for rho} gives the required result.
\end{proof}

\subsection{Concentration of inner products in the latent manifold model}
The following lemma characterizes the concentration of the inner products $\langle x_i - \nu_1,y_j - \nu_2 \rangle$ with respect to the quantities in the latent manifold model~\eqref{eq: shared manifold observation model}. 
\begin{lem} 
\label{lem:noise scalar product concentration}
Under Assumption~\ref{assump: noise magnitude}, there exist universal constants $c,p_0,\tau_0>0$, such that for any $i\in [m]$ and $j\in [n]$, and for all $\tau>\tau_0$ and $p>p_0$, we have
\begin{equation}
    \operatorname{Pr} \left\{ \left\vert  \langle x_i - \nu_1,y_j - \nu_2 \rangle - \langle \mathcal{A}_1\overline{x}_i,\mathcal{A}_2\overline{y}_j \rangle \right\vert > c\tau \sqrt{\log p} \cdot \operatorname{max}\{E,E^2\sqrt{p}\} \right\} \leq p^{-\tau}. \label{eq: inner product concentration}
\end{equation}
\end{lem}
The proof (provided below) is similar to the proof of Lemma SM1.1 in~\cite{landa2023robust}, adapted here to support the special deformations and corruptions in the latent manifold model~\eqref{eq: shared manifold observation model}.
\begin{proof}
According to the latent manifold model~\eqref{eq: shared manifold observation model}, we can write
\begin{equation}
    \langle x_i - \nu_1,y_j-\nu_2 \rangle = \langle \mathcal{A}_1\overline{x}_i,\mathcal{A}_2\overline{y}_j \rangle + \langle {\eta}_i^{(1)}, h_j^{(2)} \rangle + \langle {\eta}_j^{(2)}, h_i^{(1)} \rangle + \langle \eta_i^{(1)},\eta_j^{(2)} \rangle, \label{eq:noisy inner product decomposition}
\end{equation}
where we defined
\begin{equation}
   h_i^{(1)} = \mathcal{U}\mathcal{A}_1\overline{x}_i + \mathcal{V}_1 z_i^{(1)}, \qquad\qquad h_j^{(2)} =  \mathcal{U}\mathcal{A}_2\overline{y}_j + \mathcal{V}_2 z_j^{(2)},
\end{equation}
for $i=1,\ldots,m$ and $j=1,\ldots,n$.
Conditioning on $h^{(2)}_j$, the random variable $\langle {\eta}_i^{(1)},h^{(2)}_j \rangle$ is sub-Gaussian for each $i=1,\ldots,m$, satisfying
\begin{equation}
    \Vert \langle {\eta}_i^{(1)},h^{(2)}_j \rangle \Vert_{\psi_2}
    \leq \Vert h^{(2)}_j \Vert_2 \sup_{\Vert y \Vert_2=1} \Vert \langle {\eta}^{(1)}_i,y \rangle \Vert_{\psi_2} \leq C(C+1)E,
\end{equation}
where we used the definition of the sub-Gaussian norm~\cite{vershynin2018high} and the fact that 
\begin{equation}
    \Vert h^{(2)}_j \Vert_2 \leq \Vert \mathcal{A}_2 \Vert_2 \Vert\overline{y}_j\Vert_2 + \Vert z_j^{(2)} \Vert_2 \leq C(C+1),
\end{equation}
according to Assumption~\ref{assump: noise magnitude}.
Therefore, according to Proposition 2.5.2 in~\cite{vershynin2018high}, we have
\begin{equation}
    \operatorname{Pr}\left\{ \left\vert \langle {\eta}_i^{(1)},h^{(2)}_j \rangle \right\vert > \tau \right\} 
    \leq 2\operatorname{exp}\left\{ - \frac{\tau^2}{\widetilde{c}^2 \Vert \langle {\eta}_i^{(1)},h^{(2)}_j \rangle \Vert_{\psi_2}^2} \right\} 
    \leq 2\operatorname{exp}\left\{ -\frac{\tau^2}{\widetilde{c}^2 C^2 (C+1)^2 E^2} \right\},
\end{equation}
for a universal constant $\widetilde{c}>0$. Taking $\tau = \widetilde{c} C (C+1)  E  \sqrt{ t \log p}$, for any $t>0$, shows that 
\begin{equation}
    \operatorname{Pr}\left\{ \left\vert \langle {\eta}_i^{(1)},h^{(2)}_j \rangle \right\vert > \widetilde{c} C (C+1) E  \sqrt{ t \log p} \right\} 
    \leq 2 p^{-t}. \label{eq: langle eta, x rangle bound }
\end{equation}
Since~\eqref{eq: langle eta, x rangle bound } holds conditionally on any realization of $h^{(2)}_j$, it also holds unconditionally. Analogously to the derivation of the probabilistic bound~\eqref{eq: langle eta, x rangle bound }, it can be verified that
\begin{equation}
    \operatorname{Pr}\left\{ \left\vert \langle {\eta}_j^{(2)},h^{(1)}_i \rangle \right\vert > \widetilde{c} C (C+1) E  \sqrt{ t \log p} \right\} 
    \leq 2 p^{-t}. \label{eq: langle eta, x rangle bound 2}
\end{equation}
For the last term in~\eqref{eq:noisy inner product decomposition}, we have that
\begin{equation}
    \operatorname{Pr}\left\{ \vert \langle {\eta}_i^{(1)},{\eta}^{(2)}_j \rangle \vert > E^2 \left( \widetilde{c}_1 \sqrt{{t p \log p}} + \widetilde{c}_2 { t \log p} \right) \right\}\leq 3p^{-t}, \label{eq:bound for langle eta_i , eta_j, rangle}
\end{equation}
for all $t>0$, where $\widetilde{c}_1,\widetilde{c}_2>0$ are universal constants; see the proof of Lemma SM1.1 in~\cite{landa2023robust} and specifically the probabilistic bound on $\langle \eta_i, \eta_j \rangle$ in eq. SM1.11 therein. Overall, combining~\eqref{eq:bound for langle eta_i , eta_j, rangle},~\eqref{eq: langle eta, x rangle bound },~\eqref{eq: langle eta, x rangle bound 2}, and~\eqref{eq:noisy inner product decomposition}, applying the union bound, absorbing any constants multiplying $p^{-t}$ into the constant $c$ in~\eqref{eq: inner product concentration}, and replacing $t$ with $\tau$, provides the required result.
\end{proof}

\subsection{Order with high probability in $n$} \label{sec: order with high probability}
For brevity of our proofs, we will use the following definition.
\begin{defn}[Order with high probability in $n$] \label{def: order with high probability}
 Let $X$ be a real-valued random variable that may depend on $\varepsilon$, $m$, $n$, and $p$, and suppose that $f(m,n,p)$ is a real-valued function of $m,n,p$. We say that $X = \mathcal{O}_{n}^{(\varepsilon)}\left(f(m,n,p)\right)$ if there exist global constants $\tau_0,n_0(\varepsilon),C^{'}({\varepsilon})>0$, such that for all $n\geq n_0(\varepsilon)$, $m\geq n^\gamma$, and $p\geq n^\gamma$ (where $\gamma$ is from Assumption~\ref{assump: dimensions}), we have
\begin{equation}
    |X| \leq {\tau} C^{'}({\varepsilon}) f(m,n,p),
\end{equation}
with probability at least $1-n^{-\tau}$, for all $\tau \geq \tau_0$.
\end{defn}
Definition~\ref{def: order with high probability} is convenient for our subsequent analysis. First, if we have a collection of random variables $X_1,\ldots,X_{P(n)}$, where $X_i = \mathcal{O}_{m,n,p}^{(\varepsilon)}(f(m,n,p))$ for $i=1,\ldots,P(n)$ and $P(n)$ is a fixed polynomial in $n$ whose coefficients are global constants, then we immediately get (by applying the union bound $P(n)$ times)
\begin{equation}
    \max_{i=1,\ldots,P(n)} \vert X_i \vert = \mathcal{O}_{n}^{(\varepsilon)}(f(m,n,p)). \label{eq: definition order with high probability property 1}
\end{equation}
Second, if $X = \mathcal{O}_{n}^{(\varepsilon)}(f(m,n,p))$ and $Y = g^{(\varepsilon)}(X)$, where $g^{(\varepsilon)}\in\mathcal{C}^1(\mathbb{R})$ for all $\varepsilon>0$ and $\lim_{m,n,p\rightarrow \infty} f(m,n,p) = 0$, then by a Taylor expansion of $g^{(\varepsilon)}(x)$ around zero,
\begin{equation}
    Y = g^{(\varepsilon)}(0) + \mathcal{O}_{n}^{(\varepsilon)}(f(m,n,p)). \label{eq: definition order with high probability property 2}
\end{equation}
We will use properties~\eqref{eq: definition order with high probability property 1} and~\eqref{eq: definition order with high probability property 2}  of Definition~\ref{def: order with high probability} seamlessly throughout the remaining proofs.

\section{Proof of Theorem~\ref{thm: concentration of transport plan}} \label{appendix: proof of theorem on concentration of transport plan}
Let us write
\begin{align}
    K_{i,j} &=  \operatorname{exp}\left\{ -\frac{\Vert x_i - y_j \Vert_2^2}{\varepsilon}  \right\}  
    = \operatorname{exp}\left\{ -\frac{\Vert x_i \Vert_2^2}{\varepsilon}  \right\} \operatorname{exp}\left\{ \frac{2 \langle x_i , y_j \rangle}{\varepsilon}  \right\}  \operatorname{exp}\left\{ -\frac{\Vert y_j \Vert_2^2}{\varepsilon}  \right\} \nonumber \\
    &= \operatorname{exp}\left\{ \frac{-\Vert x_i \Vert_2^2 + \langle 2x_i - \nu_1, \nu_2 \rangle}{\varepsilon}  \right\} \operatorname{exp}\left\{ \frac{2 \langle x_i - \nu_1 , y_j - \nu_2 \rangle}{\varepsilon}  \right\}  \operatorname{exp}\left\{ \frac{-\Vert y_j \Vert_2^2 + \langle \nu_1, 2y_j-\nu_2 \rangle}{\varepsilon}  \right\}.
\end{align}
Therefore, we have
\begin{equation}
    W_{i,j} = \alpha_i K_{i,j} \beta_j = \widetilde{\alpha}_i \widetilde{K}_{i,j} \widetilde{\beta}_j, \label{eq: W alternative form using inner product kernel}
\end{equation}
where we defined
\begin{align}
    \widetilde{\alpha}_i &= \alpha_i \operatorname{exp}\left\{ \frac{-\Vert x_i \Vert_2^2 + \langle 2x_i - \nu_1, \nu_2 \rangle}{\varepsilon}  \right\}, \\
    \widetilde{\beta}_j &= \beta_j \operatorname{exp}\left\{ \frac{-\Vert y_j \Vert_2^2 + \langle 2y_j - \nu_2, \nu_1 \rangle}{\varepsilon}  \right\}, \\
    \widetilde{K}_{i,j} &= \operatorname{exp}\left\{ \frac{2 \langle x_i - \nu_1 , y_j - \nu_2 \rangle}{\varepsilon}  \right\}.
\end{align}
According to Lemma~\ref{lem:noise scalar product concentration}, Assumptions~\ref{assump: noise magnitude} and~\ref{assump: dimensions}, and Definition~\ref{def: order with high probability}, we have
\begin{align}
    \widetilde{K}_{i,j} &= \operatorname{exp}\left\{ \frac{2 \langle \mathcal{A}_1\overline{x}_i , \mathcal{A}_2 \overline{y}_j\rangle + \mathcal{O}_{n}^{(\varepsilon)}\left( \max \left\{ E\sqrt{\log p}, E^2 \sqrt{p\log p}\right\} \right)}{\varepsilon}  \right\} \nonumber \\
    &= \operatorname{exp}\left\{ \frac{2 \langle \mathcal{A}_1\overline{x}_i , \mathcal{A}_2 \overline{y}_j\rangle}{\varepsilon}  \right\} \operatorname{exp}\left\{ \frac{\mathcal{O}_{n}^{(\varepsilon)}\left( \max \left\{ E\sqrt{\log p}, E^2 \sqrt{p\log p}\right\} \right)}{\varepsilon}  \right\} \\
    &= \operatorname{exp}\left\{ \frac{2 \langle \mathcal{A}_1\overline{x}_i , \mathcal{A}_2 \overline{y}_j\rangle}{\varepsilon }  \right\} \left[ 1 + \mathcal{O}_{n}^{(\varepsilon)}\left( \max \left\{ E\sqrt{\log p}, E^2 \sqrt{p\log p}\right\} \right) \right], \label{eq: K_tilde expression}
\end{align}
for all $i\in[m]$ and $j\in[n]$, where we used the two properties of Definition~\ref{def: order with high probability} mentioned in Section~\ref{sec: order with high probability} (noting that $\max \left\{ E\sqrt{\log p}, E^2 \sqrt{p\log p}\right\} \rightarrow 0$ as $n\rightarrow \infty$ due to Assumptions~\ref{assump: noise magnitude} and~\ref{assump: dimensions}).

According to~\eqref{eq: W alternative form using inner product kernel} and our definition of $\mathcal{B}_{m,n}$ from~\eqref{eq: B def}, the matrix $\widetilde{K}$ is scaled to prescribed row sums $r_i = \sqrt{n/m}$ and column sums $c_j = \sqrt{m/n}$ (see definitions and notation in Section~\ref{appendix: stability of matrix scaling}) by the pair of scaling factors $(\widetilde{\alpha},\widetilde{\beta})$. We now proceed to show that $\widetilde{K}$ is also approximately scaled to the same prescribed row and column sums by the pair $(\hat{\alpha},\hat{\beta})$, defined by
\begin{equation}
    \hat{\rho}_\varepsilon(x) = \frac{\rho_\varepsilon(x)}{(\pi {\varepsilon})^{d/4}}, \qquad \hat{\alpha}_i = \frac{1}{\sqrt{ m}} \hat{\rho}_{{\varepsilon}}(\mathcal{A} \overline{x}_i) \operatorname{exp}\left\{ - \frac{\Vert {\mathcal{A}} \overline{x}_i \Vert_2^2}{{\varepsilon}} \right\}, \qquad 
    \hat{\beta}_j = \frac{1}{\sqrt{ n}} \hat{\rho}_{{\varepsilon}}(\mathcal{A} \overline{y}_j) \operatorname{exp}\left\{ - \frac{\Vert {\mathcal{A}} \overline{y}_j \Vert_2^2}{{\varepsilon}} \right\}, \label{eq: alpha_hat and beta_hat def}
\end{equation}
for all $x\in \mathcal{N}$, $i\in[m]$, and $j\in [n]$.
Since $\langle \mathcal{A}_1\overline{x}_i , \mathcal{A}_2 \overline{y}_j\rangle = \overline{x}_i^T \mathcal{A}_1^T \mathcal{A}_2 \overline{y}_j = \overline{x}_i^T \mathcal{A}^2 \overline{y}_j$, we have 
\begin{align}
    \frac{1}{c_j}\sum_{i=1}^m \hat{\alpha}_i \widetilde{K}_{i,j} \hat{\beta}_j 
    &= \frac{1}{m} \sum_{i=1}^m \hat{\rho}_{{\varepsilon}}(\mathcal{A} \overline{x}_i) \operatorname{exp}\left\{ - \frac{\Vert \mathcal{A}\left(\overline{x}_i - \overline{y}_j\right)\Vert_2^2}{{\varepsilon} }  \right\} \hat{\rho}_{{\varepsilon}}(\mathcal{A} \overline{y}_j) \left[ 1 + \mathcal{O}_{n}^{(\varepsilon)}\left( \max \left\{ E\sqrt{\log p}, E^2 \sqrt{p\log p}\right\} \right) \right] \nonumber \\
    &= \frac{1}{m} \sum_{i=1}^m \hat{\rho}_{{\varepsilon}}(\mathcal{A} \overline{x}_i) \operatorname{exp}\left\{ - \frac{\Vert \mathcal{A} \left( \overline{x}_i - \overline{y}_j \right) \Vert_2^2}{{\varepsilon} }  \right\} \hat{\rho}_{{\varepsilon}}(\mathcal{A} \overline{y}_j) + \mathcal{O}_{n}^{(\varepsilon)}\left( \max \left\{ E\sqrt{\log p}, E^2 \sqrt{p\log p}\right\} \right), 
\end{align}
for all $j\in [n]$, where we used Assumption~\ref{assump: noise magnitude} and the fact that $\hat{\rho}_{\hat{\varepsilon}}$ is bounded according to Lemma~\ref{lem: boundedness of the scaling function}. Conditioning on $\overline{y}_j$, we apply Hoeffding's inequality and utilize Assumption~\ref{assump: dimensions}, obtaining
\begin{align}
    \frac{1}{c_j}\sum_{i=1}^m \hat{\alpha}_i \widetilde{K}_{i,j} \hat{\beta}_j 
    &= \int_{\mathcal{M}} \rho_{{\varepsilon}}(\mathcal{A} \tilde{x}) \mathcal{K}_{{\varepsilon}} (\mathcal{A} \tilde{x},\mathcal{A} \overline{y}_j) \rho_{{\varepsilon}}(\mathcal{A} \overline{y}_j) \omega(\tilde{x}) d\mu(\tilde{x}) + \mathcal{O}_{n}^{(\varepsilon)}\left( \sqrt{\frac{\log m}{m}} \right) \nonumber \\ 
    &\phantom{==} + \mathcal{O}_{n}^{(\varepsilon)}\left( \max \left\{ E\sqrt{\log p}, E^2 \sqrt{p\log p}\right\} \right),
\end{align}
recalling that $d\mu(\tilde{x})$ is the volume form of $\mathcal{M}$ at $\tilde{x}\in \mathcal{M}$. Using the change of variables $x = \mathcal{A}\tilde{x} \in \mathcal{N}$, we integrate over $x\in \mathcal{N}$ using the pushforward volume form $d\nu (x)$ (see the beginning of Section~\ref{sec: concentration of EOT plan}), obtaining
\begin{align}
    \frac{1}{c_j}\sum_{i=1}^m \hat{\alpha}_i \widetilde{K}_{i,j} \hat{\beta}_j 
    &= \int_{\mathcal{N}} \rho_{{\varepsilon}}(x) \mathcal{K}_{{\varepsilon}} (x,\mathcal{A} \overline{y}_j) \rho_{{\varepsilon}}(\mathcal{A} \overline{y}_j) \omega(\mathcal{A}^{-1} x) d\nu(x) + \mathcal{O}_{n}^{(\varepsilon)}\left( \sqrt{\frac{\log m}{m}} \right) \nonumber \\ 
    &\phantom{==} + \mathcal{O}_{n}^{(\varepsilon)}\left( \max \left\{ E\sqrt{\log p}, E^2 \sqrt{p\log p}\right\} \right) \nonumber \\
    &= \int_{\mathcal{N}} \mathcal{W}_\varepsilon(x,\mathcal{A} \overline{y}_j)  \tilde{\omega}(x) d\nu(x) + \mathcal{O}_{n}^{(\varepsilon)}\left( \sqrt{\frac{\log m}{m}} \right) + \mathcal{O}_{n}^{(\varepsilon)}\left( \max \left\{ E\sqrt{\log p}, E^2 \sqrt{p\log p}\right\} \right) \nonumber \\
    &= 1 + \mathcal{O}_{n}^{(\varepsilon)}\left( \max \left\{ E\sqrt{\log p}, E^2 \sqrt{p\log p}, \sqrt{\frac{\log m}{m}} \right\} \right), \label{eq: approximate column scaling of K_tilde}
\end{align}
for all $i\in [m]$, where we used the double stochasticity of $\mathcal{W}_\varepsilon$~\eqref{eq:integral scaling eq with density} in the last equality. Similarly, we have
\begin{align}
    \frac{1}{r_i}\sum_{j=1}^n \hat{\alpha}_i \widetilde{K}_{i,j} \hat{\beta}_j 
    &= \frac{1}{n} \sum_{j=1}^n \hat{\rho}_{{\varepsilon}}(\mathcal{A}\overline{x}_i) \operatorname{exp}\left\{ - \frac{\Vert \mathcal{A} \left( \overline{x}_i - \overline{y}_j \right) \Vert_2^2}{{\varepsilon} }  \right\} \hat{\rho}_{{\varepsilon}}(\mathcal{A} \overline{y}_j) + \mathcal{O}_{n}^{(\varepsilon)}\left( \max \left\{ E\sqrt{\log p}, E^2 \sqrt{p\log p}\right\} \right) \nonumber \\
    &=\int_{\mathcal{M}} \rho_{{\varepsilon}}(\mathcal{A} \overline{x}_i) \mathcal{K}_{{\varepsilon}} (\mathcal{A} \overline{x}_i,\mathcal{A} \tilde{y}) \rho_{{\varepsilon}}(\mathcal{A} \tilde{y}) \omega(\tilde{y}) d\mu(\tilde{y})  \nonumber \\
    &\phantom{==} + \mathcal{O}_{n}^{(\varepsilon)}\left( \sqrt{\frac{\log n}{n}} \right)  +\mathcal{O}_{n}^{(\varepsilon)}\left( \max \left\{ E\sqrt{\log p}, E^2 \sqrt{p\log p}\right\} \right) \nonumber \\
    &=\int_{\mathcal{N}} \rho_{{\varepsilon}}(\mathcal{A} \overline{x}_i) \mathcal{K}_{{\varepsilon}} (\mathcal{A} \overline{x}_i,y) \rho_{{\varepsilon}}(y) \omega(\mathcal{A}^{-1} y) d\nu(y) \nonumber \\
    &\phantom{==} + \mathcal{O}_{n}^{(\varepsilon)}\left( \sqrt{\frac{\log n}{n}} \right) + \mathcal{O}_{n}^{(\varepsilon)}\left( \max \left\{ E\sqrt{\log p}, E^2 \sqrt{p\log p}\right\} \right) \nonumber \\
    &= 1 + \mathcal{O}_{n}^{(\varepsilon)}\left( \max \left\{ E\sqrt{\log p}, E^2 \sqrt{p\log p}, \sqrt{\frac{\log m}{m}} \right\} \right), \label{eq: approximate row scaling of K_tilde}
\end{align}
for all $i\in [m]$, where we used the change of variables $y = \mathcal{A} \tilde{y}$ and the fact that $\sqrt{\log n / n} \leq 2\sqrt{\log m / m}$ for all sufficiently large $m$ under Assumption~\ref{assump: dimensions} (recalling also that 
$m\leq n$).

We are now in a position to apply Lemma~\ref{lem:closeness of scaling factors} using the approximate scaling of $\widetilde{K}$ according to~\eqref{eq: approximate column scaling of K_tilde} and~\eqref{eq: approximate row scaling of K_tilde}. We first evaluate and bound the different quantities appearing in Lemma~\ref{lem:closeness of scaling factors}. Recall that $r_i = \sqrt{n/m}$ and $c_j = \sqrt{m/n}$. Consequently, $s = \Vert \mathbf{r} \Vert_1 = \Vert \mathbf{c} \Vert_1 = \sqrt{m n}$ and
\begin{align}
    \overline{r}_i = \frac{r_i}{\sqrt{s}} = \frac{n^{1/4}}{m^{3/4}}, \qquad  \overline{c}_j = \frac{c_j}{\sqrt{s}} = \frac{m^{1/4}}{n^{3/4}}.
\end{align}
Hence, according to~\eqref{eq: alpha_hat and beta_hat def},
\begin{align}
    C_1 &= \min_i \left\{ \frac{\hat{\alpha}_i}{\overline{r}_i} \right\} \geq C_1^{'}(\varepsilon) \left( \frac{m}{n} \right)^{1/4}, \\
    C_2 &= \min_j \left\{ \frac{\hat{\beta}_j}{\overline{c}_j} \right\} \geq C_2^{'}(\varepsilon) \left( \frac{n}{m} \right)^{1/4},
\end{align}
for some constants $C_1^{'}(\varepsilon), C_2^{'}(\varepsilon) >0$ (which may depend on $\varepsilon$), where we used the facts that $\hat{\rho}_\varepsilon$ is lower bounded according to Lemma~\ref{lem: boundedness of the scaling function} and that $\Vert \overline{x}_i\Vert_2, \Vert \overline{y}_j\Vert_2, \Vert \mathcal{A}_1\Vert_2, \Vert \mathcal{A}_2\Vert_2 \leq C$ by Assumption~\ref{assump: noise magnitude}. Note that by~\eqref{eq: K_tilde expression} and Assumption~\ref{assump: noise magnitude},
\begin{equation}
    C^{'}_3(\varepsilon) \left[ 1 + \mathcal{O}_{n}^{(\varepsilon)}\left( \max \left\{ E\sqrt{\log p}, E^2 \sqrt{p\log p}\right\} \right) \right] \leq \widetilde{K}_{i,j} \leq C^{'}_4(\varepsilon) \left[ 1 + \mathcal{O}_{n}^{(\varepsilon)}\left( \max \left\{ E\sqrt{\log p}, E^2 \sqrt{p\log p}\right\} \right) \right], 
\end{equation}
for some constants $C_3^{'}(\varepsilon), C_4^{'}(\varepsilon) >0$ for all sufficiently large $n$. Overall, applying Lemma~\ref{lem:closeness of scaling factors} with $A=\widetilde{K}$ and $(\hat{\alpha},\hat{\beta})$ from~\eqref{eq: alpha_hat and beta_hat def}, and after some manipulation, it can be verified that there exists a pair of positive vectors $({\alpha}^{'},{\beta}^{'})$ that scales $\widetilde{K}$ to row sums $\mathbf{r} = \sqrt{n/m}$ and column sums $\mathbf{c} = \sqrt{m/n}$, such that
\begin{align}
     \frac{\vert{\alpha}^{'}_i - \hat{\alpha}_i \vert}{\hat{\alpha}_i} &\leq \mathcal{O}_{n}^{(\varepsilon)}\left( \max \left\{ E\sqrt{\log p}, E^2 \sqrt{p\log p}, \sqrt{\frac{\log m}{m}} \right\} \right), \label{eq: relative error of alpha_prime} \\
    \frac{\vert{\beta}^{'}_j - \hat{\beta}_j \vert}{\hat{\beta}_j} &\leq \mathcal{O}_{n}^{(\varepsilon)}\left( \max \left\{ E\sqrt{\log p}, E^2 \sqrt{p\log p}, \sqrt{\frac{\log m}{m}} \right\} \right), \label{eq: relative error of beta_prime}
\end{align}
for all $i\in [m]$ and $j\in [n]$.

Finally, since both $(\widetilde{\alpha},\widetilde{\beta})$ from~\eqref{eq: W alternative form using inner product kernel} and $({\alpha}^{'},{\beta}^{'})$ are pairs of scaling factors of $\widetilde{K}$ for the same prescribed row/column sums, we can write
\begin{align}
    W_{i,j} &= \alpha_i K_{i,j} \beta_j = \widetilde{\alpha}_i \widetilde{K}_{i,j} \widetilde{\beta}_j = {\alpha^{'}}_i \widetilde{K}_{i,j} {\beta^{'}}_j \nonumber \\
    &= \hat{\alpha}_i \widetilde{K}_{i,j} \hat{\beta}_j \left[ 1 + \mathcal{O}_{n}^{(\varepsilon)}\left( \max \left\{ E\sqrt{\log p}, E^2 \sqrt{p\log p}, \sqrt{\frac{\log m}{m}} \right\} \right) \right] \nonumber \\
    &= \frac{1}{(\pi {\varepsilon})^{d/2}\sqrt{m n}} \rho_{{\varepsilon}}(\mathcal{A}\overline{x}_i) \operatorname{exp}\left\{ - \frac{\Vert \mathcal{A} \overline{x}_i \Vert_2^2}{{\varepsilon}} \right\} \operatorname{exp}\left\{ \frac{2 \langle \mathcal{A} \overline{x}_i , \mathcal{A} \overline{y}_j\rangle}{{\varepsilon} }  \right\} \operatorname{exp}\left\{ - \frac{\Vert \mathcal{A} \overline{y}_j \Vert_2^2}{{\varepsilon}} \right\} \rho_{{\varepsilon}}(\mathcal{A} \overline{y}_j) \nonumber \\ &\phantom{==} \times \left[ 1 + \mathcal{O}_{n}^{(\varepsilon)}\left( \max \left\{ E\sqrt{\log p}, E^2 \sqrt{p\log p}, \sqrt{\frac{\log m}{m}} \right\} \right) \right],  \nonumber \\
    &= \frac{1}{\sqrt{m n}} \rho_{{\varepsilon}}(\mathcal{A}\overline{x}_i) \mathcal{K}_{\varepsilon}(\mathcal{A} \overline{x}_i, \mathcal{A} \overline{y}_j )  \rho_{{\varepsilon}}(\mathcal{A} \overline{y}_j)  \left[ 1 + \mathcal{O}_{n}^{(\varepsilon)}\left( \max \left\{ E\sqrt{\log p}, E^2 \sqrt{p\log p}, \sqrt{\frac{\log m}{m}} \right\} \right) \right],
\end{align}
for all $i\in[m]$ and $j\in [n]$, where we also used~\eqref{eq: relative error of alpha_prime} and~\eqref{eq: relative error of beta_prime}. The proof of Theorem~\ref{thm: concentration of transport plan} follows immediately from the definitions of $\mathcal{W}_\varepsilon$ and $\mathcal{K}_\varepsilon$ in~\eqref{eq: W integral def} and Assumption~\ref{assump: noise magnitude}.

\section{Proof of Corollary~\ref{cor: pointwise convergence of P}} \label{appendix: proof of corollary on pointwise convergence of P}
According to the definition of $P$ in~\eqref{eq: P def}, we can write
\begin{align}
    &\left\Vert P \left[ {f} \right]_{(\overline{\mathcal{X}},\overline{\mathcal{Y}})} - \left[ \mathcal{P}_{\varepsilon} f \right]_{(\overline{\mathcal{X}},\overline{\mathcal{Y}})} 
 \right\Vert_\infty \nonumber \\
 &=  \max \left\{\left\Vert 
     \sqrt{\frac{m}{n}} W [h]_{\overline{\mathcal{Y}}} - 
     [\mathcal{W}_{\varepsilon} h]_{\overline{\mathcal{X}}} \right\Vert_\infty
     , \left\Vert \sqrt{\frac{n}{m}} W^T [g]_{\overline{\mathcal{X}}} - [\mathcal{W}_{\varepsilon} g]_{\overline{\mathcal{Y}}} \right\Vert_\infty \right\}.
\end{align}
Using Theorem~\ref{thm: concentration of transport plan}, we have
\begin{align}
    \sqrt{\frac{m}{n}} W_{i,\cdot} [h]_{\overline{\mathcal{Y}}} 
    &= \frac{1}{n} \sum_{j=1}^n \mathcal{W}_{\varepsilon}(\mathcal{A} \overline{x}_i, \mathcal{A} \overline{y}_j) h(\mathcal{A} \overline{y}_j) + \mathcal{O}_{n}^{(\varepsilon)}\left( \max \left\{ E\sqrt{\log p}, E^2 \sqrt{p\log p}, \sqrt{\frac{\log m}{m}} \right\} \right),
\end{align}
where we used the fact that $\mathcal{W}_\varepsilon$ and $h$ are bounded over $\mathcal{N}$ by universal constants according to Assumption~\ref{assump: noise magnitude} and the conditions in Corollary~\ref{cor: pointwise convergence of P} (see Lemma~\ref{lem: boundedness of the scaling function} and~\eqref{eq: boundedness of the Gaussian kernel}). Next, we apply Hoeffding's inequality and obtain
\begin{align}
    \sqrt{\frac{m}{n}} W_{i,\cdot} [h]_{\overline{\mathcal{Y}}} 
    &= \int_{\mathcal{M}} \mathcal{W}_{\varepsilon}(\mathcal{A} \overline{x}_i, \mathcal{A} \tilde{y}) h(\mathcal{A} \tilde{y}) \omega(\tilde{y}) d\mu(\tilde{y})  \nonumber \\
    &\phantom{\int_{\mathcal{M}}}+ \mathcal{O}_{n}^{(\varepsilon)}\left( \sqrt{\frac{\log n}{n}}  \right) 
    + \mathcal{O}_{n}^{(\varepsilon)}\left( \max \left\{ E\sqrt{\log p}, E^2 \sqrt{p\log p}, \sqrt{\frac{\log m}{m}} \right\} \right).
\end{align}
Using the change of variables $y = \mathcal{A}\tilde{y} \in \mathcal{N}$ and the definition $\tilde{w}(y) = \omega(\mathcal{A}^{-1} y)$ (see the beginning of Section~\ref{sec: concentration of EOT plan}), we integrate over $y\in \mathcal{N}$ using the volume form $d\nu (x)$, and thus
\begin{align}
    \sqrt{\frac{m}{n}} W_{i,\cdot} [h]_{\overline{\mathcal{Y}}}  &= \int_{\mathcal{N}} \mathcal{W}_{\varepsilon}(\mathcal{A} \overline{x}_i, y) h(y) \tilde{\omega}(y) d\nu(y)  \nonumber \\
    &\phantom{\int_{\mathcal{M}}}+ \mathcal{O}_{n}^{(\varepsilon)}\left( \sqrt{\frac{\log n}{n}}  \right) 
    + \mathcal{O}_{n}^{(\varepsilon)}\left( \max \left\{ E\sqrt{\log p}, E^2 \sqrt{p\log p}, \sqrt{\frac{\log m}{m}} \right\} \right) \nonumber \\
    &= \{\mathscr{W}_{\varepsilon} h\}(\mathcal{A} \overline{x}_i) + \mathcal{O}_{n}^{(\varepsilon)}\left( \max \left\{ E\sqrt{\log p}, E^2 \sqrt{p\log p}, \sqrt{\frac{\log m}{m}} \right\} \right),
\end{align}
for any $i\in [m]$ according to Assumption~\ref{assump: dimensions}. Using the union bound, we have 
\begin{equation}
    \left\Vert 
     \sqrt{\frac{m}{n}} W [h]_{\overline{\mathcal{Y}}} - 
     [\mathscr{W}_{\varepsilon} h]_{\overline{\mathcal{X}}} \right\Vert_\infty = \mathcal{O}_{n}^{(\varepsilon)}\left( \max \left\{ E\sqrt{\log p}, E^2 \sqrt{p\log p}, \sqrt{\frac{\log m}{m}} \right\} \right).
\end{equation}
By an analogous derivation, it can be verified that 
\begin{equation}
    \left\Vert \sqrt{\frac{n}{m}} W^T [g]_{\overline{\mathcal{X}}} - [\mathscr{W}_{\varepsilon} g]_{\overline{\mathcal{Y}}} \right\Vert_\infty 
    = \mathcal{O}_{n}^{(\varepsilon)}\left( \max \left\{ E\sqrt{\log p}, E^2 \sqrt{p\log p}, \sqrt{\frac{\log m}{m}} \right\} \right).
\end{equation}
Overall, we obtain
\begin{equation}
    \left\Vert P \left[ {f} \right]_{(\overline{\mathcal{X}},\overline{\mathcal{Y}})} - \left[ \mathcal{P}_{\varepsilon} f \right]_{(\overline{\mathcal{X}},\overline{\mathcal{Y}})} 
 \right\Vert_\infty = \mathcal{O}_{n}^{(\varepsilon)}\left( \max \left\{ E\sqrt{\log p}, E^2 \sqrt{p\log p}, \sqrt{\frac{\log m}{m}} \right\} \right). \label{eq: one step error}
\end{equation}
Let us define $f_1 = \mathcal{P}_{\varepsilon} f $. Clearly, $f_1\in\mathcal{H}$ and is bounded on $\mathcal{N}$, hence~\eqref{eq: one step error} also holds when replacing $f$ with $f_1$. We now have 
\begin{align}
    &\left\Vert P^2 \left[ {f} \right]_{(\overline{\mathcal{X}},\overline{\mathcal{Y}})} - \left[ \mathcal{P}^2_{\varepsilon} f \right]_{(\overline{\mathcal{X}},\overline{\mathcal{Y}})} 
 \right\Vert_\infty 
 = \left\Vert P^2 \left[ {f} \right]_{(\overline{\mathcal{X}},\overline{\mathcal{Y}})} - \left[ \mathcal{P}_{\varepsilon} f_1 \right]_{(\overline{\mathcal{X}},\overline{\mathcal{Y}})} 
 \right\Vert_\infty \nonumber \\
 &\leq \left\Vert P^2 \left[ {f} \right]_{(\overline{\mathcal{X}},\overline{\mathcal{Y}})} - P \left[ f_1 \right]_{(\overline{\mathcal{X}},\overline{\mathcal{Y}})} 
 \right\Vert_\infty 
 + \left\Vert P \left[ f_1 \right]_{(\overline{\mathcal{X}},\overline{\mathcal{Y}})} - \left[ \mathcal{P}_{\varepsilon} f_1 \right]_{(\overline{\mathcal{X}},\overline{\mathcal{Y}})} 
 \right\Vert_\infty \nonumber \\
 &\leq \max_i \{ \sum_j P_{i,j} \} \left\Vert P \left[ {f} \right]_{(\overline{\mathcal{X}},\overline{\mathcal{Y}})} - \left[ f_1 \right]_{(\overline{\mathcal{X}},\overline{\mathcal{Y}})} 
 \right\Vert_\infty 
 + \mathcal{O}_{n}^{(\varepsilon)}\left( \max \left\{ E\sqrt{\log p}, E^2 \sqrt{p\log p}, \sqrt{\frac{\log m}{m}} \right\} \right) \nonumber \\
 &\leq \mathcal{O}_{n}^{(\varepsilon)}\left( \max \left\{ E\sqrt{\log p}, E^2 \sqrt{p\log p}, \sqrt{\frac{\log m}{m}} \right\} \right), \label{eq: two step error}
\end{align}
where we used~\eqref{eq: one step error} and the fact that $P$ is row-stochastic ($\sum_j P_{i,j} = 1$ for all $i$) in the last transition. We can proceed recursively by defining $f_{k} = \mathcal{P}_{\varepsilon} f_{k-1} $ for $k=2,\ldots,t$, repeating the process in~\eqref{eq: one step error} and~\eqref{eq: two step error} for each step. Eventually, we obtain
\begin{equation}
    \left\Vert P^t \left[ {f} \right]_{(\overline{\mathcal{X}},\overline{\mathcal{Y}})} - \left[ \mathcal{P}_{\varepsilon}^t f \right]_{(\overline{\mathcal{X}},\overline{\mathcal{Y}})} 
 \right\Vert_\infty = \mathcal{O}_{n}^{(\varepsilon)}\left( \max \left\{ E\sqrt{\log p}, E^2 \sqrt{p\log p}, \sqrt{\frac{\log m}{m}} \right\} \right).
\end{equation}
The proof of Corollary~\ref{cor: pointwise convergence of P} follows immediately from the fact that $\max \left\{ E\sqrt{\log p}, E^2 \sqrt{p\log p}, \sqrt{\frac{\log m}{m}} \right\} \longrightarrow 0$ as $n\rightarrow\infty$ by Assumptions~\ref{assump: noise magnitude} and~\ref{assump: dimensions}.

\section{Proof of Corollary~\ref{cor: relative error of W for small bandwidth}} \label{appendix: proof of corollary on relative error of W for small bandwidth}
According to Theorem~\ref{thm: concentration of transport plan}, we have
\begin{align}
    \frac{\sqrt{mn}W_{i,j}}{\hat{\mathcal{W}}_{{\varepsilon}}(\mathcal{A} \overline{x}_i,\mathcal{A} \overline{y}_j)} 
    &= \frac{\rho_{{\varepsilon}}(\mathcal{A} \overline{x}_i) \mathcal{K}_{{\varepsilon}}(\mathcal{A} \overline{x}_i,\mathcal{A} \overline{y}_j) \rho_{{\varepsilon}}(\mathcal{A} \overline{y}_j) \left[ 1 + \mathcal{O}_{n}^{(\varepsilon)}\left( \max \left\{ E\sqrt{\log p}, E^2 \sqrt{p\log p}, \sqrt{\frac{\log m}{m}} \right\} \right)\right]}{\frac{\mathcal{K}_{{\varepsilon}}(\mathcal{A} \overline{x}_i,\mathcal{A} \overline{y}_j) }{\sqrt{\hat{\omega}(\mathcal{A} \overline{x}_i) \hat{\omega}(\mathcal{A} \overline{y}_j)}}} \nonumber \\
    &= \rho_{{\varepsilon}}(\mathcal{A} \overline{x}_i)  \rho_{{\varepsilon}} (\mathcal{A} \overline{y}_j) \sqrt{\hat{\omega}(\mathcal{A} \overline{x}_i) \hat{\omega}(\mathcal{A} \overline{y}_j)} \left[ 1 +  \mathcal{O}_{n}^{(\varepsilon)}\left( \max \left\{ E\sqrt{\log p}, E^2 \sqrt{p\log p}, \sqrt{\frac{\log m}{m}} \right\} \right) \right], \nonumber \\
    &= \rho_{{\varepsilon}}(\mathcal{A} \overline{x}_i)  \rho_{{\varepsilon}} (\mathcal{A} \overline{y}_j) \sqrt{\hat{\omega}(\mathcal{A} \overline{x}_i) \hat{\omega}(\mathcal{A} \overline{y}_j)} + \mathcal{O}_{n}^{(\varepsilon)}\left( \max \left\{ E\sqrt{\log p}, E^2 \sqrt{p\log p}, \sqrt{\frac{\log m}{m}} \right\} \right),
\end{align}
for all $i\in [m]$ and $j\in [n]$, where we used the properties of Definition~\ref{def: order with high probability}, the boundedness of $\rho_\varepsilon$ according to Lemma~\ref{lem: boundedness of the scaling function}, and the fact that 
\begin{equation}
    \hat{\omega}(y) = \frac{\tilde{\omega}(y)}{\sqrt{\det\left\{ \widetilde{\mathcal{T}}_{\mathcal{M}}^T(y) \, \mathcal{A}^2 \, \widetilde{\mathcal{T}}_{\mathcal{M}}(y) \right\}}} \leq \frac{c}{c^d} = c^{1-d}, \label{eq: upper bound on omega_hat}
\end{equation}
for all $y\in \mathcal{N}$ by the assumptions of Corollary~\ref{cor: relative error of W for small bandwidth}.
Applying Hoeffding's inequality, we can write
\begin{align}
    \frac{1}{mn}\sum_{i,j} \left\vert \frac{\sqrt{mn}W_{i,j}}{\hat{\mathcal{W}}_{{\varepsilon}}(\mathcal{A} \overline{x}_i,\mathcal{A} \overline{y}_j)} - 1 \right\vert &= \int_\mathcal{N} \int_\mathcal{N} \left\vert \rho_{{\varepsilon}}(x)  \rho_{{\varepsilon}} (y) \sqrt{\hat{\omega}(x) \hat{\omega}(y)} - 1 \right\vert \tilde{\omega}(x) \tilde{\omega}(y) d\nu(x) d\nu(y) \nonumber \\
    &+ \mathcal{O}_{n}^{(\varepsilon)}\left( \max \left\{ E\sqrt{\log p}, E^2 \sqrt{p\log p}, \sqrt{\frac{\log m}{m}} \right\} \right).
\end{align}
Let $d\hat{\mu}$ be the natural volume form on $\mathcal{N} \subset \mathbb{R}^r$ induced by the Euclidean metric in $\mathbb{R}^r$. In particular, we have 
\begin{equation}
    \int_\mathcal{N} g(y) \tilde{\omega}(y) d\nu(y) 
    = \int_\mathcal{N} g(y) {\omega}(\mathcal{A}^{-1} y) d\nu(y) 
    = \int_\mathcal{M} g(\mathcal{A} x) {\omega}(x) d\mu(x)
    = \int_\mathcal{N} g(y) \hat{\omega}(y) d\hat{\mu}(y),
\end{equation}
 for any measurable function $g$ on $\mathcal{N}$, where we used the change of variables $y = \mathcal{A} x$ for $x\in\mathcal{M}$ that yields 
 \[
 d\hat{\mu} (y) = \sqrt{\det\left\{ \widetilde{\mathcal{T}}_{\mathcal{M}}^T(x) \, \mathcal{A}^2 \, \widetilde{\mathcal{T}}_{\mathcal{M}}(x) \right\}} d\mu (x) . 
 \]
 Therefore, 
\begin{align}
    \frac{1}{mn}\sum_{i,j} \left\vert \frac{\sqrt{mn}W_{i,j}}{\hat{\mathcal{W}}_{{\varepsilon}}(\mathcal{A} \overline{x}_i,\mathcal{A} \overline{y}_j)} - 1 \right\vert &= \int_\mathcal{N} \int_\mathcal{N} \left\vert \rho_{{\varepsilon}}(x)  \rho_{{\varepsilon}} (y) \sqrt{\hat{\omega}(x) \hat{\omega}(y)} - 1 \right\vert \hat{\omega}(x) \hat{\omega}(y) d\hat{\mu}(x) d\hat{\mu}(y) \nonumber \\
    &+ \mathcal{O}_{n}^{(\varepsilon)}\left( \max \left\{ E\sqrt{\log p}, E^2 \sqrt{p\log p}, \sqrt{\frac{\log m}{m}} \right\} \right).
\end{align}
According to the conditions in Corollary~\ref{cor: relative error of W for small bandwidth}, $\mathcal{M}$ is compact and smooth with no boundary. Since $\mathcal{N}$ is defined by applying the linear map $x \mapsto \mathcal{A} x$ to each point in $\mathcal{M}$, the new manifold $\mathcal{N}$ also satisfies the same properties. Additionally, since $\omega \in \mathcal{C}^6$ and $\mathcal{M}$ is smooth, then also $\hat{\omega} \in \mathcal{C}^6$ according to its definition~\eqref{eq: W_tilde_hat kernel def}. We now apply Theorem 2.6 in~\cite{landa2023robust} to the scaling function $\rho_\varepsilon$ defined on the manifold $\mathcal{N}$, obtaining
\begin{equation}
    \rho_\varepsilon(x) = \frac{1}{\sqrt{\hat{\omega}(x)}} + e_\varepsilon(x), \qquad \qquad \int_\mathcal{N} \vert e_\varepsilon(x) \vert d\hat{\mu}(x) \leq c_1 \varepsilon,
\end{equation}
for all $x\in\mathcal{N}$ and $\varepsilon \leq \varepsilon_0$, where $c_1,\varepsilon_0>0$ are some constants that may depend on $\mathcal{N}$ and $\hat{\omega}$ (in addition to the global constants in our assumptions), and hence by product on $\mathcal{M}$, $\omega$, and $\mathcal{A}$. We thus have
\begin{align}
    \frac{1}{mn}\sum_{i,j} \left\vert \frac{\sqrt{mn}W_{i,j}}{\hat{\mathcal{W}}_{{\varepsilon}}(\mathcal{A} \overline{x}_i,\mathcal{A} \overline{y}_j)} - 1 \right\vert &= \int_\mathcal{N} \int_\mathcal{N} \left\vert \frac{e_\varepsilon(x)}{\sqrt{\hat{\omega}(y)}} + \frac{e_\varepsilon(y)}{\sqrt{\hat{\omega}(x)}} + e_\varepsilon(x) e_\varepsilon(y) \right\vert \hat{\omega}(x) \hat{\omega}(y) d\hat{\mu}(x) d\hat{\mu}(y) \nonumber \\
    &\phantom{==}+ \mathcal{O}_{n}^{(\varepsilon)}\left( \max \left\{ E\sqrt{\log p}, E^2 \sqrt{p\log p}, \sqrt{\frac{\log m}{m}} \right\} \right) \nonumber \\
    &\leq 2\left( \int_\mathcal{N} \left\vert e_\varepsilon(x) \right\vert \hat{\omega}(x) d\hat{\mu}(x)\right) \left( \int_\mathcal{N} \sqrt{\hat{\omega}(x)} d\hat{\mu}(x) \right) + \left( \int_\mathcal{N} \left\vert e_\varepsilon(x) \right\vert \hat{\omega}(x) d\hat{\mu}(x)\right)^2 \nonumber \\
    &\phantom{==}+ \mathcal{O}_{n}^{(\varepsilon)}\left( \max \left\{ E\sqrt{\log p}, E^2 \sqrt{p\log p}, \sqrt{\frac{\log m}{m}} \right\} \right) \nonumber \\
    &\leq c_2 \varepsilon + c_3\varepsilon^2 +\mathcal{O}_{n}^{(\varepsilon)}\left( \max \left\{ E\sqrt{\log p}, E^2 \sqrt{p\log p}, \sqrt{\frac{\log m}{m}} \right\} \right),
\end{align}
for some constants $c_2,c_3>0$ that may depend on $\mathcal{M}$, $\omega$, and $\mathcal{A}$, where we used~\eqref{eq: upper bound on omega_hat}. Since $c_3 \varepsilon^2$ is smaller than $c_2 \varepsilon$ for all sufficiently small $\varepsilon$, we obtain the required result.

\section{Supplementary Notes on Numerical Results} \label{supp.simu.sec}

\paragraph{Additional alignment evaluations} For each of the simulation experiments, we also consider an alternative metric to evaluate the performance of each method in aligning the shared manifold  structures  (i.e., torus, or clusters) between the two datasets (batches). Specifically, once the joint embeddings are obtained, we calculate the Davies-Bouldin (D-B) index with respect to the dataset/batch labels. For a given integrated dataset consisting of $K$ batches $C_1, C_2, ..., C_K$, the D-B index is defined as
\begin{equation}
\text{D-B index} =  \frac{1}{K}\sum_{k=1}^K \max_{j\ne k}\frac{S_k+S_j}{M_{{k,j}}},
\end{equation}
where
\begin{equation}
S_i=\bigg[\frac{1}{|C_k|}\sum_{i\in C_k}\|X_i-A_k\|_2^2\bigg]^{1/2},\qquad M_{kj}=\|A_k-A_j\|_2,
\end{equation}
with $A_k$ being the centroid of batch $k$ of size $|C_k|$, $X_i$ is the embedding of $i$-th data point. A higher D-B index indicates stronger mixing or a better alignment of the batches.

\paragraph{Hyperparameters} The proposed EOT eigenmap requires specifying three major hyperparameters: the kernel bandwidth or entropy regularization parameter $\epsilon>0$, the diffusion steps $t\ge 0$, and the embedding dimension $q$.  In practice, these parameters are user-specified depending on the specific applications, whose determination should be subject to critical procedures described in community guidelines \cite{yu2020veridical,yu2020data} to enhance contextual fitness, stability, and reproducibility.

\begin{figure}[h!]
	\centering
\includegraphics[angle=0,width=14cm]{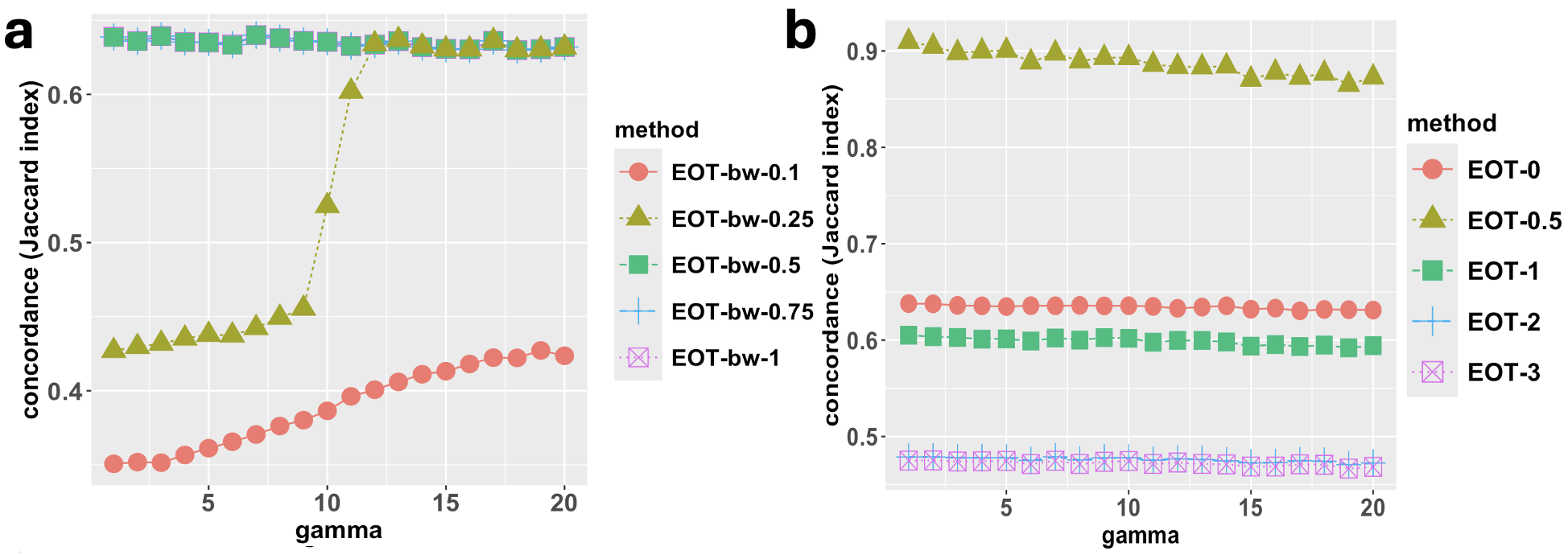}
	\caption{\footnotesize Evaluation of the performance of the proposed EOT joint embedding under the same simulation setup as the bottom panel of Figure~\ref{simu.fig}c. (a) Comparison of the performance of EOT ($t=0$) under various $\epsilon$. We set $\epsilon$ as $(\alpha\times $ median of $\{\|x_i-y_j\|\}_{i\in[m],j\in[n]})^2$, where $\alpha\in\{0.1, 0.25, 0.5, 0.75, 1\}$, and denote the methods as ``EOT-bw-$\alpha$". (b) Comparison of the performance of EOT under different $t\in\{0,0.5, 1,2,3\}$. 
    } 
	\label{epsilon.fig}
\end{figure}

Below we provide some general recommendations which may be considered as a default setting for exploratory analysis. 
Specifically, the bandwidth parameter $\varepsilon$ can be determined as the median of pairwise squared distances between the datasets, that is, the median of $\{\|x_i-y_j\|_2^2\}_{1\le i\le m, 1\le j\le n}$. This procedure has been found robust against high-dimensional noise and theoretically appealing \cite{ding2023learning}. We followed such a recommendation throughout our simulations and real data analysis, demonstrating its practical efficacy. {To demonstrate the practical advantage of such a heuristic, we conducted additional experiments under the noisy manifold alignment setup (setting 2), where we evaluated the performance of our proposed method using decreasing values of  $\epsilon$. Specifically, for each simulated dataset, we set $\epsilon$ to be $(\alpha\times $ median of the pairwise distances $\{\|x_i-y_j\|\}_{i\in[m],j\in[n]})^2$, where $\alpha\in\{0.1, 0.25, 0.5, 0.75, 1\}$. We found that the proposed EOT joint embedding with $t=0$ achieved the best performance when $\alpha\ge 0.5$, which outperformed the alternative methods (see Figure~\ref{simu.fig}c),  while the performance declined as $\alpha$ decreased further, especially when $\alpha=0.1$; see Figure~\ref{epsilon.fig}a. This analysis suggests the practical advantage of the proposed heuristic for choosing $\epsilon$, in contrast to choosing $\epsilon\to0$.}

\begin{figure}[h!]
	\centering
\includegraphics[angle=0,width=14cm]{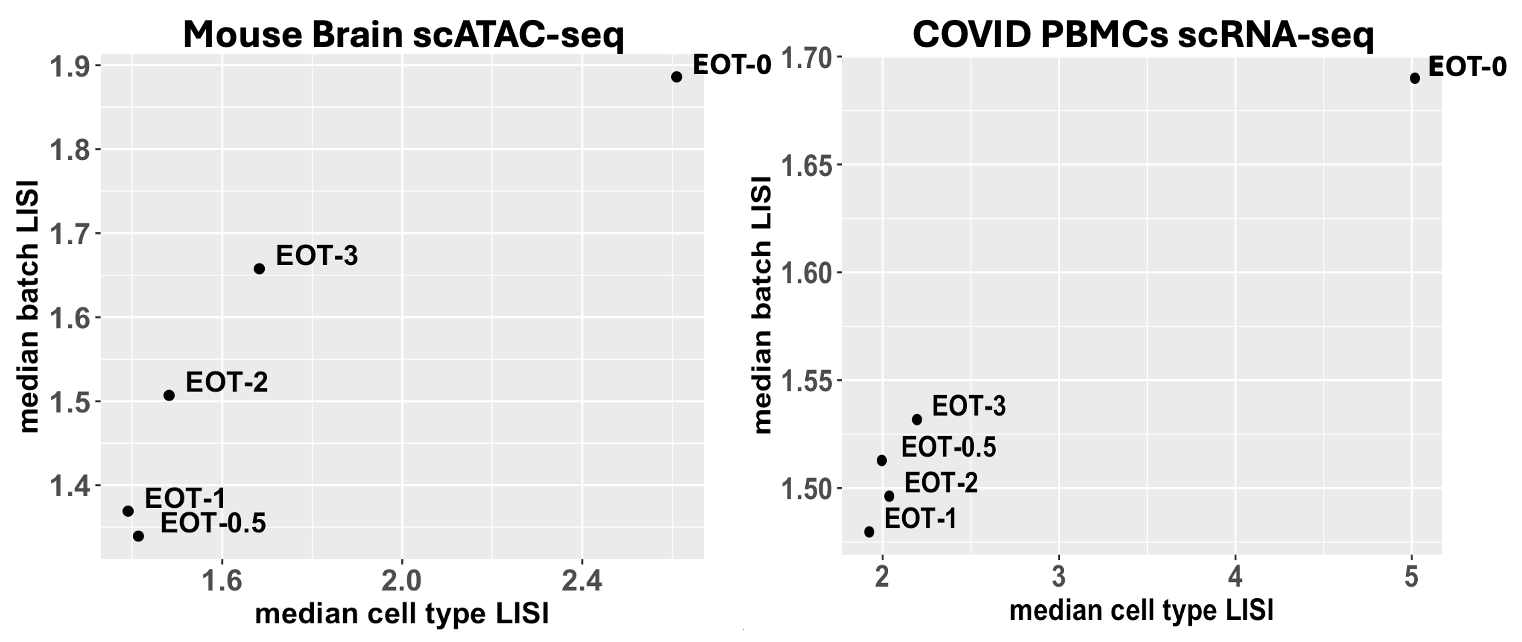}
	\caption{\footnotesize Evaluation of the performance of the proposed EOT joint embedding in the real datasets, under different choices of $t\in\{0,0.5,1,2,3\}$. 
    } 
	\label{real.t.fig}
\end{figure}

{For the diffusion step $t$, we also conducted additional experiments under the noisy manifold alignment simulation setting 2, where we  compared the performance of the proposed EOT method under each $t\in\{0,0.5, 1,2,3\}$. We found that the proposed EOT joint embedding achieved the best performance when $t=0.5$, followed by $t=0$ and $t=1$, which still outperformed the alternative methods, whereas $t=2$ and $t=3$ led to the worst performance (Figure~\ref{epsilon.fig}b). Similar comparisons were conducted on the real datasets. We applied EOT with different values of $t \in \{0, 0.5, 1, 2, 3\}$ to obtain joint embeddings for the Mouse Brain scATAC-seq datasets, and the COVID PBMCs scRNA-seq datasets. Comparing Figure~\ref{real.t.fig} with Figure~\ref{real.fig}, we found that, compared with the existing methods, EOT with $t \in \{0.5, 1, 2\}$ consistently achieved strong integration performance while preserving cell type information. In contrast, EOT with $t \in \{0, 3\}$ tended to yield inflated cell type LISI values, indicating reduced ability to maintain distinct cell types in the joint embedding space. In summary, EOT with $t=0.5$ and $t=1$ achieved the best performance in both our simulation and real data analyses.}

The embedding dimension $q$ depends on the informativeness of the leading singular values and vectors of the EOT plan $W$. Here we recommend setting $q=
	\max_{1 \leq i \leq  m-1} \left\{ i :  {s_i}/{s_{i+1}} \geq 1+\mathsf{s}  \right\}$,
where $\mathsf{s}=0.02$. For the diffusion parameter $t\ge 0$, based on our discussion after Proposition \ref{prop: diffusion distance identities}, we recommend setting $t=0$ if one is more confident about the embedding dimension $q$, and setting $t>0$ otherwise.

\begin{figure}[h!]
	\centering
\includegraphics[angle=0,width=8cm]{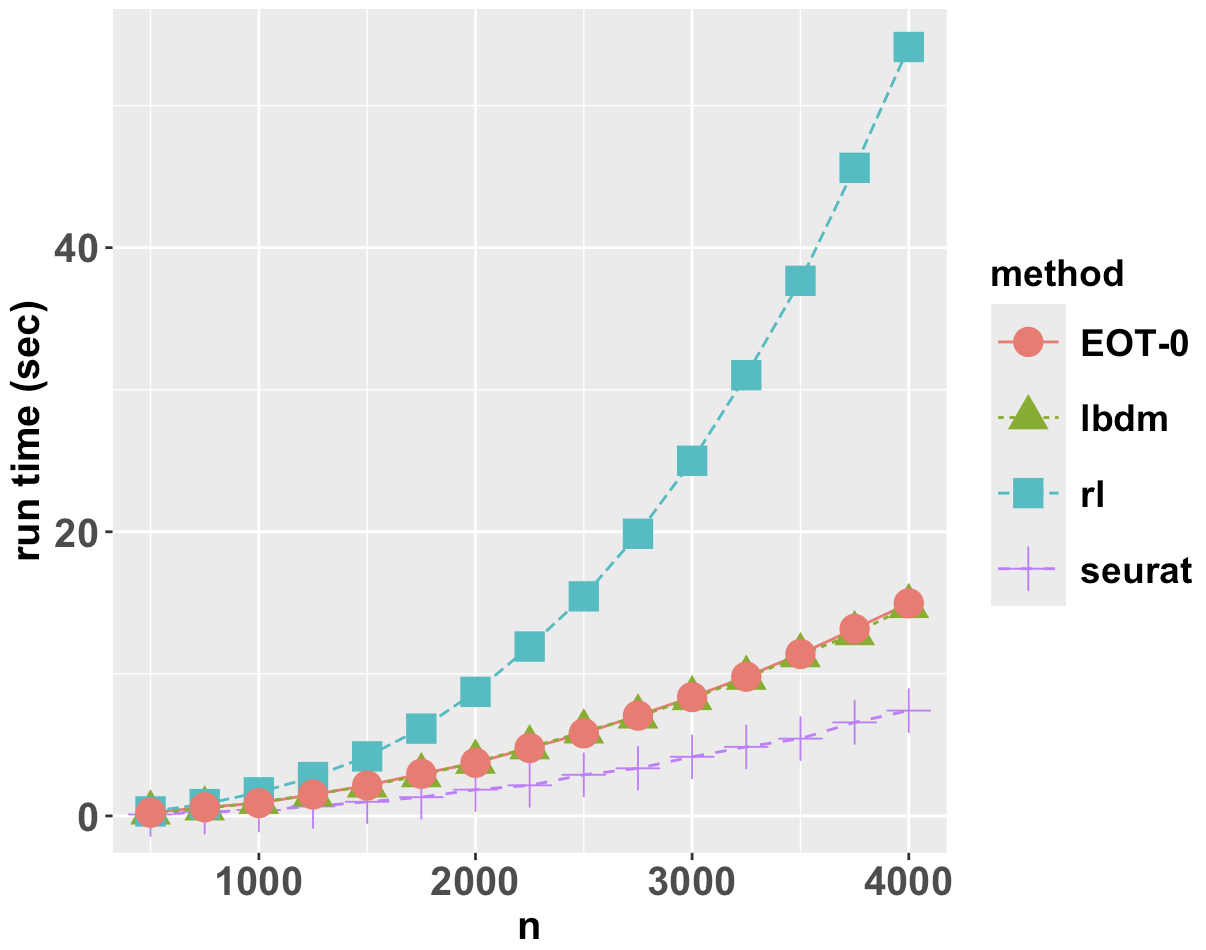}
	\caption{\footnotesize Comparison of computing time of EOT-1, lbdm, rl, and seurat over the simulated datasets, generated under the same setup as setting 2 of the noisy manifold alignment task, with fixed $\theta$. 
    } 
	\label{time.fig}
\end{figure}

{\paragraph{Computing time} In Figure~\ref{time.fig}, we compared the computing time of EOT-1 with three existing methods, ``lbdm", ``rl" and ``seurat", which achieved relatively good performance in Figure~\ref{simu.fig}, over the simulated datasets (generated under the same setup as setting 2 of the noisy manifold alignment task, with fixed $\theta$) with various sample sizes. Our results suggest that EOT required computing time comparable to ``lbdm," substantially less time than ``rl", and slightly more time than ``seurat." The longer computation time required by ``rl" is largely attributable to the additional matrix multiplications involved in its landmark-affinity-based normalization step \cite{shen2022scalability}. This indicates that the proposed algorithm is scalable to large datasets, requiring less than 20 seconds to process datasets with sample size around 4000. }

\paragraph{Inferring simultaneous accessibility-expression modules} We consider the joint analysis of simultaneous profiling of gene expression and chromatin accessibility of human prefrontal cortex (PFC) cells generated from the Single Cell Opioid Responses in the Context of HIV (SCORCH) Study \cite{ament2024single}. Specifically, we examined a dataset for a PFC sample of an HIV patient containing $n=7534$ cells, generated from the 10X Multiome technology capturing concurrent measurements of gene expression profiles via single nucleus RNA sequencing (snRNA-seq) and chromatin accessibility landscapes within individual cells. We focused on a selection of chromatin accessibility regions proximal to gene promoters. The degree of chromatin openness in these promoter regions is indicative of the corresponding gene’s activity level. For the gene expression data, we selected $p_1=2000$ most variable genes, whereas for the chromatin accessibility data, we selected $p_2=5000$ most variably accessible regions. Our goal is to obtain a joint embedding of all the $p_1+p_2$ features from both modalities into the same low-dimensional space, so as to identify associations between chromatin accessible regions and genes. %
To obtain joint embeddings of the features, we apply our proposed method by reversing the roles of cells and features. Additional implementation details are provided in Appendix~\ref{supp.simu.sec}.

\begin{figure}[t!]
	\centering
\includegraphics[angle=0,width=14cm]{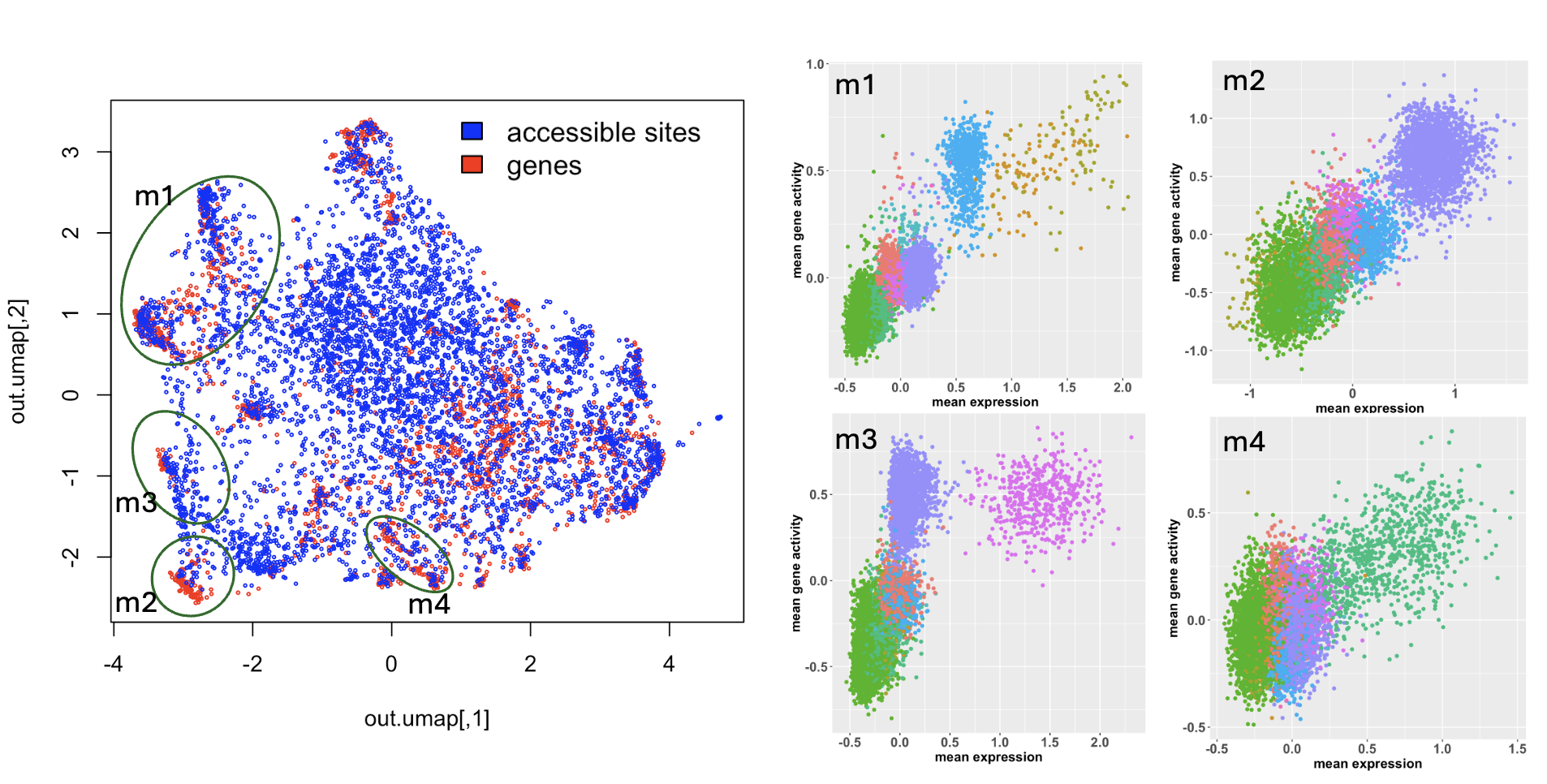}
	\caption{\footnotesize Integrative analyses of single-cell multi-omics data. Left: UMAP visualization of the joint low-dimensional embedding of genes and accessible chromatin regions, colored according to feature modalities. We select four clusters of features  (m1-m4), identified by DBSCAN  and each consisting of a regulatory module, for closer examination. Right: Scatter plots of the average expression (x-axis) of the genes  and the average level of accessibility or gene activity (y-axis) of the accessible regions contained in each module for all the cells, where the cells are colored according to their cell type annotations. 
    } 
	\label{real.figb}
\end{figure} 

Figure~\ref{real.figb} left contains a UMAP visualization~\cite{becht2019dimensionality} of the obtained joint embedding, where the embedding dimension $r=34$ is determined based on thresholding the eigengap (Appendix~\ref{supp.simu.sec}). The closeness of features in the joint embedding space, or in the UMAP visualization, indicates a higher association between their values across the cell population. The UMAP visualization reveals clusters of features, with many clusters containing both genes and accessible chromatin regions. We interpret these clusters as simultaneous accessibility-expression regulatory modules. Specifically, each module consists of a group of correlated (co-accessible) chromatin accessible regions and a group of correlated (co-expressed) genes, which are mutually associated to each other, possibly as a result of some shared regulatory pathways. To further examine the relationship between the associated accessible sites and the genes, we choose as examples four different modules (m1-m4), indicated in Figure~\ref{real.figb} left, that are identified by the DBSCAN algorithm \cite{hahsler2019dbscan}  based on our joint embeddings (Figure~\ref{supp.fig.cluster}).  Figure~\ref{real.figb} right contains the scatter plots of the average expression (x-axis) of the genes  and the average level of accessibility or gene activity (y-axis) of the accessible regions contained in each module for all the cells, where the cells are colored according to their cell type annotations. We observe that, for each module, there exists at least one cell type that demonstrates a higher expression level for the genes in the module, and/or a higher accessibility level for the chromatin regions in the same module, as compared with  other cell types.  For example, module m2 contains 133 genes and 31 accessible regions, whose average values are simultaneously higher in oligodendrocytes as compared with other cell types. %
Interestingly, in module m3, both oligodendrocytes and OPCs have higher average values for the 33 accessible regions, whereas only OPCs have higher expression level for the 133 genes in m3, indicating the cell-type-specific nature of gene regulation. These analyses are useful for %
understanding the gene regulation heterogeneity across cell types.

\paragraph{Implementation details}
For the joint embedding method ``seurat" evaluated in our experiments, we consider its core step described in the original publication \cite{stuart2019comprehensive}, where the low-dimensional embeddings for datasets $X\in\mathbb{R}^{m\times p}$ and $Y\in \mathbb{R}^{n\times p}$ are defined as the leading left and right singular vectors of the product matrix $XY^\top$.
For standard kernel methods such as ``kpca" and ``j-kpca", we again follow \cite{ding2023learning} and use the median of the respective pairwise distances as the kernel bandwidth.

To pre-process the single-cell gene expression data, we performed quality control, normalization, and scaling of the raw count data using the R functions \texttt{CreateSeuratObject}, \texttt{NormalizeData} and \texttt{ScaleData}
under default settings as incorporated in the R package \texttt{Seurat}. We also applied the R
function \texttt{FindVariableFeatures} in \texttt{Seurat} to identify $p=1000$ (three sample alignment tasks) or $p=2000$ (SCORCH multiomics analysis)
most variable genes for subsequent analysis. For the single-cell chromatin accessibility data, we used R functions \texttt{RunTFIDF} and \texttt{FindTopFeatures} from the R package \texttt{Signac} to normalize the count matrix and find the $p=5000$ most variable  accessible chromatin regions.

In our analysis of the SCORCH multiomics dataset, we selected the embedding dimension $q$ based on the following eigenvalue thresholding method. Suppose $\{s_i\}_{i=1}^m$ are the singular values of $W$. Then we choose $q=
	\max_{1 \leq i \leq  m-1} \left\{ i :  {s_i}/{s_{i+1}} \geq 1+\mathsf{s}  \right\}$,
where $\mathsf{s}=0.02$.

\section{Supplementary Figures} \label{supp.fig.sec}
\setcounter{figure}{0}
\renewcommand{\thefigure}{S\arabic{figure}}

Additional figures from our numerical simulations and real data analyses are presented below.

\begin{figure}[t!]
	\centering
\includegraphics[angle=0,width=15cm]{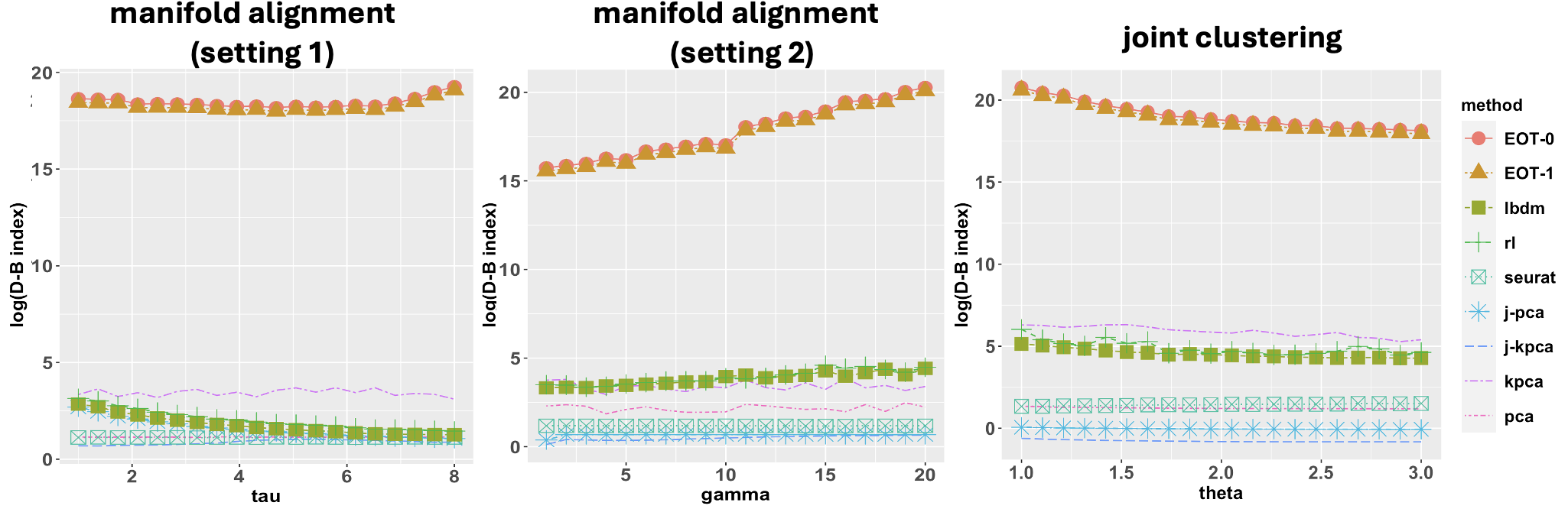}
	\caption{\footnotesize Comparison of Davies-Bouldin index of nine integration methods in various simulations. Left and Middle: simulations for noisy manifold alignment, setting 1 (left) and setting 2 (middle). Right: simulations for joint clustering. Our results indicate superior performance of the proposed methods (``EOT-0" and ``EOT-1") in aligning the latent structures.} 
	\label{supp.fig}
\end{figure}

\begin{figure}[h!]
	\centering
\includegraphics[angle=0,width=15cm]{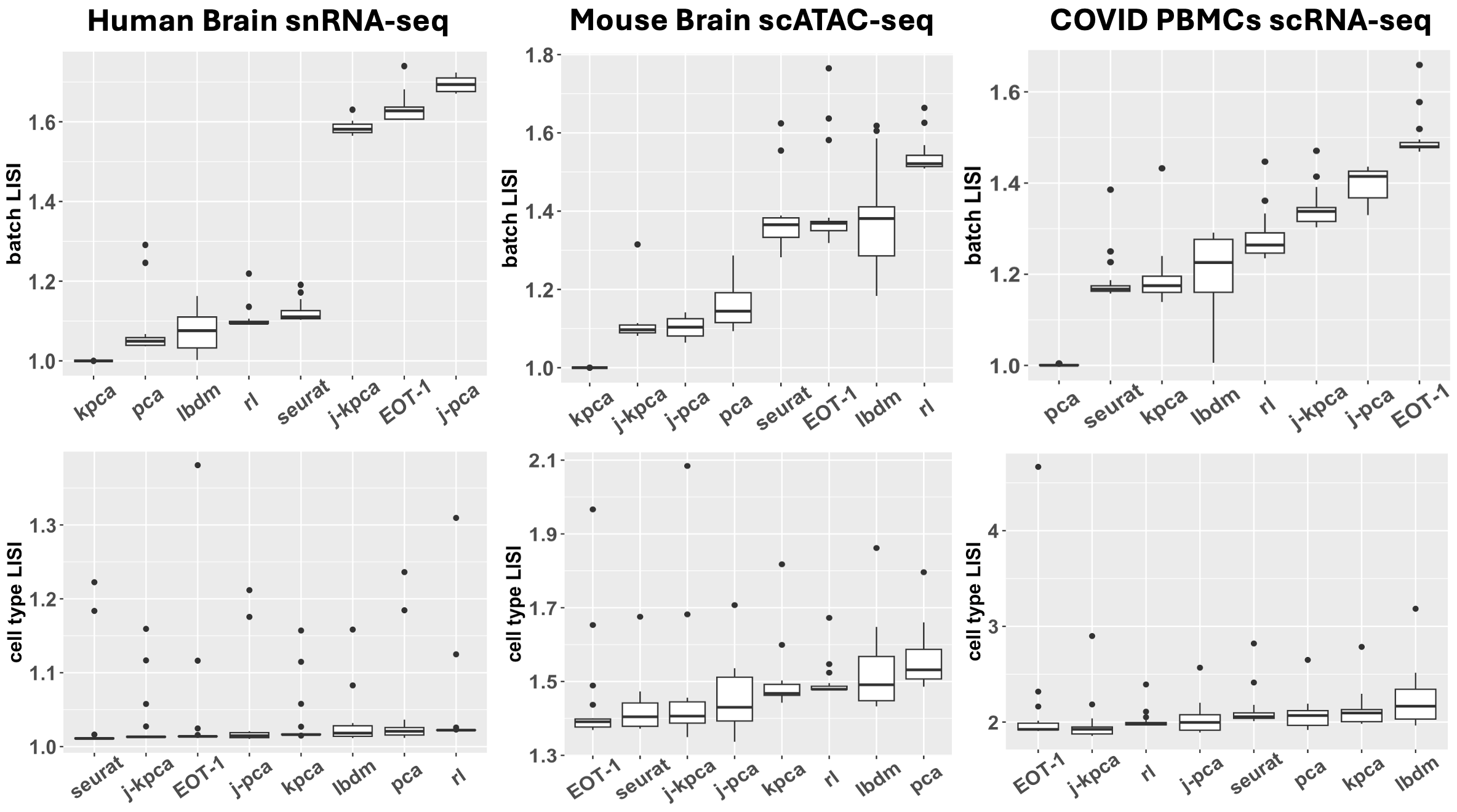}
	\caption{\footnotesize Comparison of performance metrics--batch LISI (Top row) and cell type LISI (Bottom row)--of eight integration methods in three pairs of single-cell omics data, where each boxplot contains the metrics for each method across a range of embedding dimensions $q$ (from 2 to 20). 
    } 
	\label{box.fig2}
\end{figure}

\begin{figure}[h!]
	\centering
\includegraphics[angle=0,width=8cm]{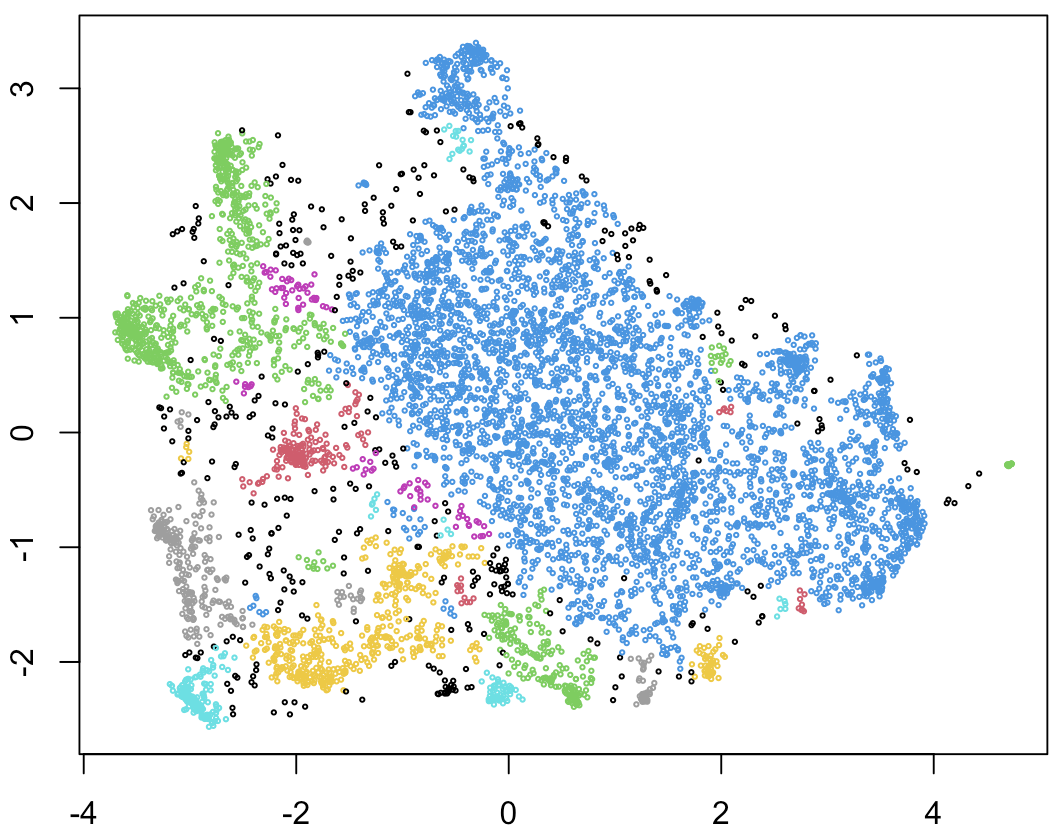}
	\caption{Joint clustering of genes and accessible chromatin regions identified by DBSCAN algorithm \cite{hahsler2019dbscan}. Each cluster of features has a distinct color, whereas the black dots correspond to the ``singletons" identified by DBSCAN.} 
	\label{supp.fig.cluster}
\end{figure}

\end{appendices}

\end{document}